\documentclass[11pt]{article}

\usepackage[preprint]{acl}

\usepackage{caption}
\usepackage{times}
\usepackage{latexsym}
\usepackage{url}
\usepackage[T1]{fontenc}

\usepackage[utf8]{inputenc}

\usepackage{microtype}

\usepackage{inconsolata}

\usepackage{graphicx}

\usepackage{booktabs}
\usepackage{multirow}
\usepackage[table]{xcolor}
\usepackage{colortbl}
\usepackage{pifont} 
\usepackage{subcaption}
\usepackage{amsmath}
\definecolor{groupgray}{RGB}{245,245,245}
\definecolor{oursgreen}{RGB}{226,242,226}
\definecolor{headergray}{RGB}{235,238,242}
\definecolor{textgray}{RGB}{90,90,90}

\newcommand{\cmark}{\ding{51}}
\newcommand{\xmark}{\ding{55}}

\usepackage{csquotes}
\MakeOuterQuote{"}
\title{The Illusion of Specialization: Unveiling the Domain-Invariant "Standing Committee" in Mixture-of-Experts Models}

\author{
 \textbf{Yan Wang\textsuperscript{1}\thanks{~These authors contributed equally as co-first authors.}\thanks{~Corresponding author}},
 \textbf{Yitao Xu\textsuperscript{2}\footnotemark[1]},
 \textbf{Nanhan Shen\textsuperscript{2}\footnotemark[1]},
 \\
 \textbf{Jinyan Su\textsuperscript{3}},
 \textbf{Jimin Huang\textsuperscript{5,1}},
 \textbf{Zining Zhu\textsuperscript{4}}
\\
\\
 \textsuperscript{1}The Fin AI
 \textsuperscript{2}Georgia Institute of Technology
 \\
 \textsuperscript{3}Cornell University
 \textsuperscript{4}Stevens Institute of Technology
 \textsuperscript{5}The University of Manchester
\\
 \small{
   \textbf{Emails:} 
   \href{mailto:wy2266336@gmail.com}{wy2266336@gmail.com},
   \href{mailto:zzhu41@stevens.edu}{zzhu41@stevens.edu}
 }
}

\begin{document}
\maketitle
\begin{abstract}
Mixture of Experts models are widely assumed to achieve domain specialization through sparse routing. In this work, we question this assumption by introducing \textsc{COMMITTEEAUDIT}, a post hoc framework that analyzes routing behavior at the level of expert groups rather than individual experts. Across three representative models and the MMLU benchmark, we uncover a domain invariant \textbf{Standing Committee}. This is a compact coalition of routed experts that consistently captures the majority of routing mass across domains, layers, and routing budgets, even when architectures already include shared experts. Qualitative analysis further shows that Standing Committees anchor reasoning structure and syntax, while peripheral experts handle domain-specific knowledge. These findings reveal a strong structural bias toward centralized computation, suggesting that specialization in Mixture of Experts models is far less pervasive than commonly believed. Crucially, this inherent bias indicates that current training objectives, such as load-balancing losses that enforce uniform expert utilization, may be working against the model's natural optimization path, thereby limiting training efficiency and performance. The code is available at GitHub\footnote{\url{https://github.com/The-FinAI/CommitteeAudit}}.
\end{abstract}

\section{Introduction}
\label{intro}
Large Language Models (LLMs) have demonstrated remarkable capabilities in complex reasoning and understanding tasks~\cite{qian2025fino1transferabilityreasoningenhancedllms,wang2025rkefino1,wang2025fintaggingbenchmarkingllmsextracting,wang2025finauditingfinancialtaxonomystructuredmultidocument}. To further scale these models without incurring proportional computational costs, the Mixture-of-Experts (MoE) architecture has emerged as a dominant approach. By activating only a sparse subset of parameters for each input token, MoE models promise to decouple model capacity from inference latency. This conditional computation paradigm is particularly appealing for general-purpose LLMs because different domains exhibit heterogeneous computational patterns that theoretically benefit from expert specialization.

\begin{figure}[t]
  \centering
  \begin{subfigure}[t]{0.87\linewidth}
    \centering
    \includegraphics[width=\linewidth]{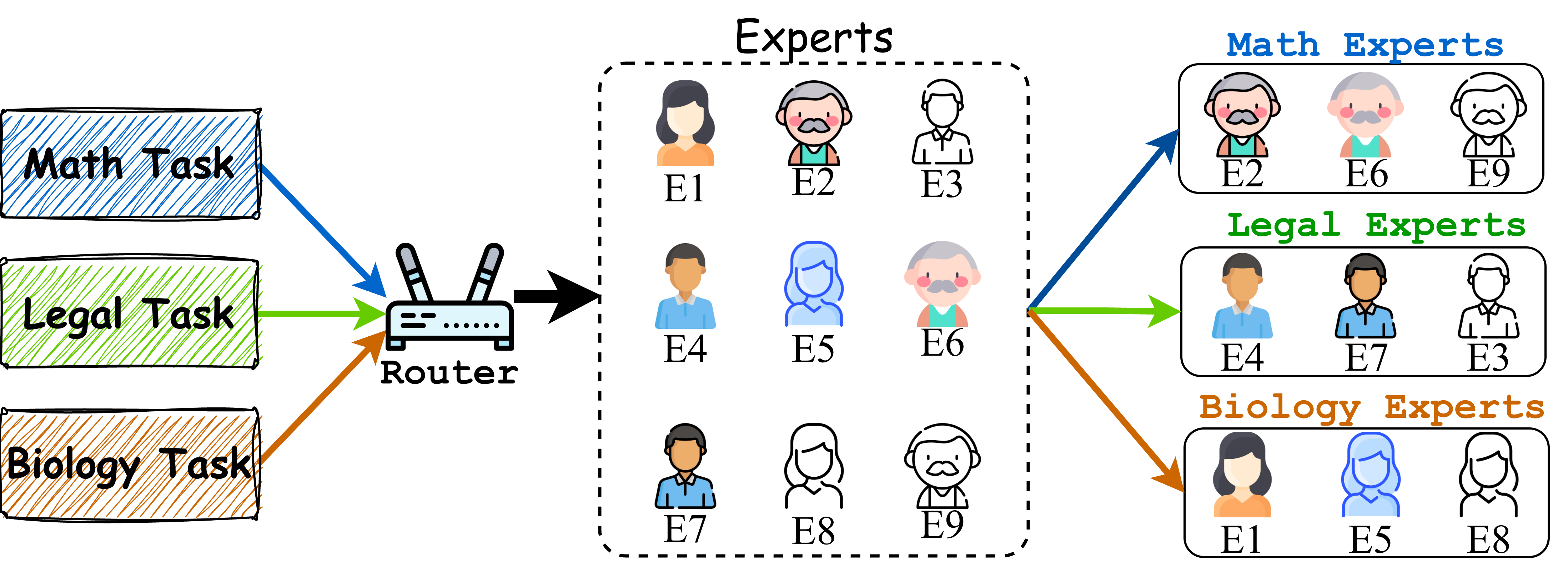}
    \caption{Conventional domain-specific intuition.}
    \label{fig:motivation:intuition}
  \end{subfigure}

  \vspace{0.8em}

  \begin{subfigure}[t]{0.87\linewidth}
    \centering
    \includegraphics[width=\linewidth]{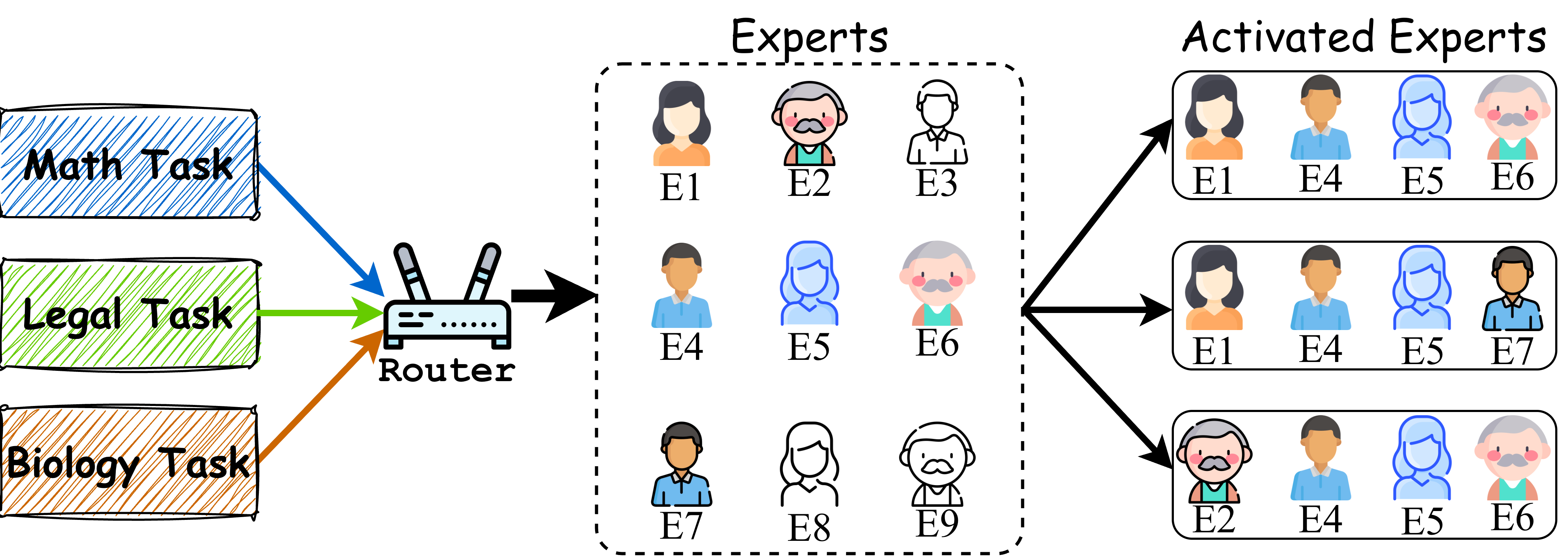}
    \caption{Empirically observed expert sharing.}
    \label{fig:motivation:observation}
  \end{subfigure}

  \caption{From domain-specific intuition to empirically observed expert sharing in Mixture-of-Experts models. 
  (a) \textbf{The Intuition:} The ideal "Divide-and-Conquer" strategy assumes disjoint sets of experts for different domains. 
  (b) \textbf{The Observation:} Empirical routing patterns reveal a \textbf{Standing Committee} (e.g., Experts E4 and E5) that is consistently activated across disparate domains (Math, Legal, Biology), acting as a generalist core hidden within the routed experts.}
  \label{fig:motivation}
\end{figure}


The design philosophy of MoE models often follows a divide-and-conquer principle where experts are expected to specialize by domain. Under this view, sparsity arises when the model routes inputs to distinct expert groups, such as a dedicated set for mathematics and another for legal tasks, as illustrated in Figure~\ref{fig:motivation:intuition}. However, the optimization dynamics of sparse routing frequently contradict this ideal separation. Prior research on \textit{Representation Collapse} \cite{chi2022representation,do2025simsmoe} warns of a pathological state where gating networks fail to optimize effectively, causing experts to become redundant or completely inactive. Recognizing that natural language relies heavily on high-frequency and domain-agnostic patterns, recent state-of-the-art (SOTA) architectures have moved to institutionalize a centralized processing unit. For instance, DeepSeek \cite{dai2024deepseekmoe,deepseekai2024deepseekv2strongeconomicalefficient,deepseekai2025deepseekv3technicalreport} introduces \textbf{Shared Experts} that are always activated to isolate common knowledge from the routed experts. The prevailing assumption is that by architecturally separating these generalists, the remaining routed experts are free to become true specialists. Yet, our empirical analysis reveals that this architectural fix does not fully suppress the drive toward centralization. As shown in Figure~\ref{fig:motivation:observation}, we observe substantial cross-domain sharing even among the experts explicitly designated for specialization. Unlike representation collapse, where experts die out due to optimization failure, these shared experts are highly active and functionally competent but simply refuse to specialize. This evidence suggests that the formation of a generalist core is not merely an architectural choice but an inevitable emergent property of sparse computation. However, this emergent structural bias poses a significant challenge to conventional MoE training: standard load-balancing auxiliary losses, which are designed to prevent expert idle-out by encouraging uniform selection, may inadvertently suppress this natural computational hierarchy. This conflict potentially leads to suboptimal convergence and wasted FLOPs on peripheral experts that lack fundamental reasoning capabilities.

\begin{table}[t]
\centering
\caption{Comparison with existing MoE interpretability works. Prior studies mainly focus on individual expert specialization, internal representations, or frequency-based importance. In contrast, our work identifies stable, domain-invariant expert committees.}
\label{tab:comparison}
\resizebox{\linewidth}{!}{
\begin{tabular}{>{\raggedright\arraybackslash}p{3.2cm} >{\raggedright\arraybackslash}p{5.3cm} c c c}
\toprule
\rowcolor{headergray}
\textbf{Research Focus} & \textbf{Representative Works} & \textbf{\begin{tabular}[c]{@{}c@{}}Unit of\\ Analysis\end{tabular}} & \textbf{\begin{tabular}[c]{@{}c@{}}Dominant\\ Assumption\end{tabular}} & \textbf{\begin{tabular}[c]{@{}c@{}}Captures\\ Co-activation\end{tabular}} \\
\midrule

\multirow{4}{*}{\textbf{\begin{tabular}[l]{@{}l@{}}Routing \&\\ Behavior\\ Analysis\end{tabular}}}
& A Closer Look into MoE~\cite{lo2025closer} & Individual & General Polysemy & \xmark \\
& Probing Semantic Routing~\cite{olson2025probing} & Individual & Domain-Specific & \xmark \\
& Context Faithfulness~\cite{bai2025understanding} & Individual & Context-Dependent & \xmark \\
& \multicolumn{4}{l}{\small\textcolor{textgray}{\textit{$\rightarrow$ Focus: Which individual expert activates for a specific token?}}} \\
\midrule

\multirow{4}{*}{\textbf{\begin{tabular}[l]{@{}l@{}}Representation\\ \& Intrinsic\\ Mechanisms\end{tabular}}}
& Secretly Embedding Model~\cite{li2024your} & Global/Layer & Latent Space & \xmark \\
& MoE-X~\cite{yang2025mixture} & Individual & Modular & \xmark \\
& Intrinsic User-Centric~\cite{swamy2024intrinsic} & Individual & User-Aligned & \xmark \\
& \multicolumn{4}{l}{\small\textcolor{textgray}{\textit{$\rightarrow$ Focus: What information is encoded within experts?}}} \\
\midrule

\multirow{2}{*}{\textbf{\begin{tabular}[l]{@{}l@{}}Criticality\\ Analysis\end{tabular}}}
& Unveiling Super Experts~\cite{su2025unveilingsuperexpertsmixtureofexperts} & Individual & Pareto Principle & \xmark \\
& \multicolumn{4}{l}{\small\textcolor{textgray}{\textit{$\rightarrow$ Focus: Which single experts are dominant or critical?}}} \\
\midrule

\rowcolor{oursgreen}
\multirow{2}{*}{\textbf{\begin{tabular}[l]{@{}l@{}}Structural\\ Organization\end{tabular}}}
& \textbf{Standing Committee Analysis (Ours)} & \textbf{Committee} & \textbf{Domain-Invariant} & \textbf{\cmark} \\
\rowcolor{oursgreen}
& \multicolumn{4}{l}{\small\textit{$\rightarrow$ Focus: How stable groups of experts dominate computation across domains?}} \\
\bottomrule
\end{tabular}
}
\end{table}


Recent advances in MoE interpretability have largely focused on the properties of individual experts. Prior work has examined semantic routing patterns \cite{olson2025probing,lo2025closer,bai2025understanding}, analyzed internal representations \cite{li2024your,yang2025mixture}, and most recently identified frequency-based “Super Experts” \cite{su2025unveilingsuperexpertsmixtureofexperts}. However, these studies predominantly treat experts as independent computational units, where importance is quantified through isolated activation statistics. Consequently, they overlook the potential for a higher-level structural organization within the routing mechanism. While the "Super Expert" phenomenon highlights the Pareto distribution of individual criticality, it fails to capture the relational stability between experts across varying contexts. This leaves a critical gap in our understanding: do experts function as isolated specialists whose prominence is a mere statistical byproduct, or do they spontaneously organize into stable, domain-invariant coalitions?

To resolve this discrepancy between domain-specific intuition and observed expert sharing, we propose \textsc{CommitteeAudit}, a post-hoc analytical framework designed to audit the group-level structural organization of pre-trained MoE models. Unlike prior works that focus on individual expert statistics, our framework quantifies the stability and intersection of expert coalitions across divergent tasks. We apply this auditing process to three representative models (\textbf{OLMoE~\cite{muennighoff2025olmoeopenmixtureofexpertslanguage}, Qwen3-30B-A3B~\cite{yang2025qwen3technicalreport}, and DeepSeek-V2-Lite~\cite{deepseekai2024deepseekv2strongeconomicalefficient}}), covering both standard routing and architectures with explicit shared experts.

Guided by this structural perspective, our study addresses three fundamental questions regarding the hidden organization of MoE computation:
\noindent \textbf{Existence:} Do routed experts naturally self-organize into stable, domain-invariant groups that dominate computation, or do they remain specialized by task?
\noindent \textbf{Dynamics:} How does this group-level organization evolve across network depths? Is the centralization of experts an inevitable emergent property of sparse routing?
\noindent \textbf{Functionality:} What functional roles do these stable groups play? Specifically, does the model rely on them for general reasoning while relegating specific knowledge to a fringe of other experts?

Our contributions answer these questions and challenge the prevailing view of MoE specialization:

(1) We provide systematic evidence of a domain-invariant \textbf{Standing Committee}, a compact expert coalition that emerges regardless of shared-expert architectures, revealing a structural bias that challenges "fairness-oriented" load balancing.

(2) We introduce a model-agnostic framework, \textsc{COMMITTEEAUDIT}, that utilizes Pareto-optimal ranking and stability diagnostics to quantify group-level expert organization beyond individual activation statistics.

(3) Through qualitative analysis, we uncover a core-periphery organization where committee members anchor logical and syntactic structures, while peripheral experts manage domain-specific knowledge.

\section{Related Works}

\noindent \textbf{Expert Specialization and Routing Analysis}
MoE models are commonly motivated by a divide-and-conquer intuition: sparsity arises when tokens are routed to domain-specialized experts~\cite{xue2024openmoe,jiang2024mixtral,zoph2022st,dai2024deepseek,fan2024empiricalunderstandingmoedesign}. Early multilingual studies supported this view, reporting experts that preferentially served specific languages~\cite{lepikhin2020gshard,zheng2024efficiently}.
However, recent analyses of general-purpose LLMs reveal a more nuanced picture. Experts often behave polysemously rather than strictly specializing~\cite{lo2025closer}, and routing only weakly aligns with human semantic domains~\cite{olson2025probing}. Other work shows that specialization is modulated by context rather than being an intrinsic property of an expert~\cite{bai2025understanding}. In parallel, studies on "super experts" highlight a small set of disproportionately active experts~\cite{su2025unveilingsuperexpertsmixtureofexperts}, shifting attention from specialization to expert criticality.
\\
\noindent \textbf{Internal Representations and Intrinsic Interpretability}
A complementary line of work examines the internal representations of MoE systems. Evidence suggests that experts contribute to a shared latent space rather than operating as isolated modules~\cite{li2024your}. To improve interpretability, architectural interventions have been proposed, including constraints that encourage interpretable expert roles~\cite{yang2025mixture} and routing mechanisms designed to align usage with higher-level semantic concepts~\cite{swamy2024intrinsic}.
\\
\noindent \textbf{From Individual Experts to Collective Structure}
Despite these advances, most prior studies analyze experts as independent computational units, focusing on activation patterns or internal states~\cite{ghandeharioun2024patchscopes}. What remains unclear is whether experts organize into stable, co-activated groups that persist across tasks. Our work addresses this gap by shifting the lens from individual experts to structured collectives, "standing committees", and shows that such committees emerge in a domain-invariant manner, challenging the conventional assumption of purely domain-specific routing.

\section{\textsc{CommitteeAudit}}
\subsection{Preliminaries}

\paragraph{Mixture-of-Experts Architecture.}
Mixture-of-Experts (MoE) models extend the Transformer by replacing the feed-forward network with a set of $E$ parallel experts $\{E_i\}_{i=1}^E$~\cite{shazeer2017outrageously,lepikhin2020gshard,fedus2022switch}. For a token $x \in \mathbf{R}^d$ at layer $\ell$, a gating network produces a routing vector $G^{(\ell)}(x)$. 
Under Top-$k$ routing, the layer output is the weighted sum of $k$ activated experts: 
\begin{equation}
    y = \sum_{i \in \text{Top-}k} G^{(\ell)}(x)_i E_i(x).
\end{equation}

While token-level routing is sparse, aggregating decisions over a corpus reveals structural regularities.

\paragraph{Expert Contribution Index (ECI).}
To quantify expert importance at the domain task level, we define the \emph{Expert Contribution Index} (ECI). Given a corpus $\mathcal{D}$ partitioned into domain tasks $\mathcal{T} = \{\tau\}$, we denote $\mathcal{D}_\tau$ as the subset for a domain task $\tau$. For expert $i$ at layer $\ell$, the ECI is the expected routing weight:


\begin{equation}
\label{eq:eci}
c^{(\ell)}_{i,\tau}
=\mathbf{E}_{x \in \mathcal{D}_\tau}
\!\left[
G^{(\ell)}(x)_i
\right]
\end{equation}

Unlike activation frequency, ECI preserves the \emph{magnitude} of router preference, providing a more informative signal for ranking. ECI serves as the building block for analyzing cross-task invariants.

\subsection{Framework Description}
\label{framework}
\begin{figure}[t]
    \centering
    \includegraphics[width=1\linewidth]{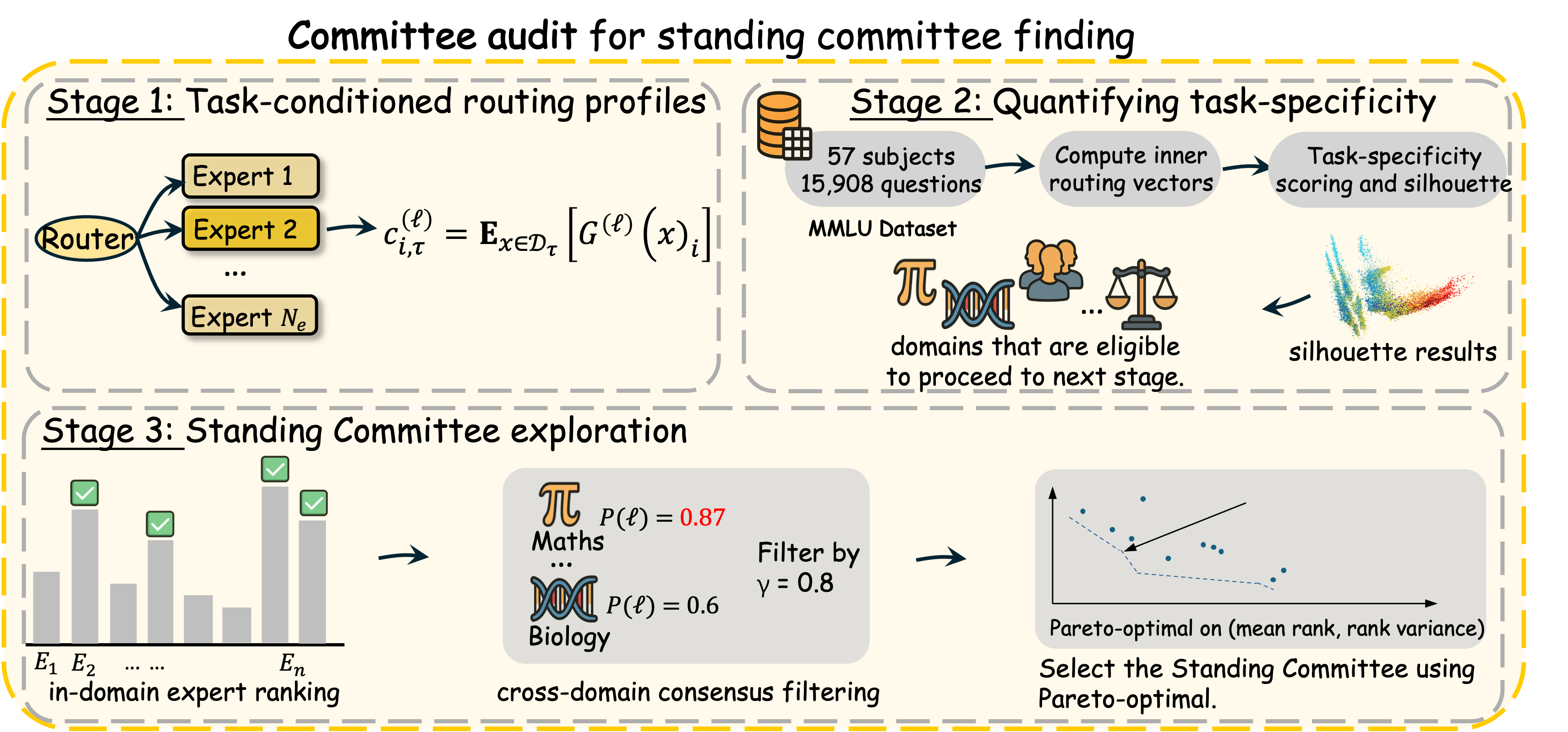}
    \caption{Overview of the COMMITTEEAUDIT framework.}
    \label{fig:framework}
\end{figure}

\textsc{CommitteeAudit}, as shown in Figure~\ref{fig:framework}, is a domain-conditioned routing analysis framework that (i) extracts domain-level routing profiles, (ii) quantifies inter-domain routing divergence, and (iii) explores Standing Committees. In high-capacity MoEs ($E \ge 64$), while single experts may occasionally dominate, activation is generally too distributed for individual-centric analysis. We hypothesize that specialization is expressed through a structured distribution over a subset of experts, referred to as a \emph{committee}.

\paragraph{Stage I: Task-conditioned routing profiles.}
Before constructing committees, we first extract routing representations from the MoE model.
For every sample $x \in \mathcal{D}_\tau$ and MoE layer $\ell$, we run the
model and record the full routing vector $G^{(\ell)}(x)$ taken at the last token unless otherwise specified:
\begin{equation}
\label{eq:vec}
    G^{(\ell)}(x)=\mathrm{softmax}(z^{(\ell)}(x)) \in \Delta^{E-1},
\end{equation}
where $\Delta^{E-1}$ denotes the probability simplex over $E$ experts, that is $\Delta^{E-1} = \{p \in \mathbf{R}^E: p_i \geq0, \Sigma_i p_i=1 \}$.

We use the full routing
distribution (rather than discrete Top-$k$ activations) because it preserves the
complete preference structure over experts.

We then aggregate routing behavior at the domain-task level. For each expert $i$, we
compute its ECI  $c^{(\ell)}_{i,\tau}$ used Eq~(\ref{eq:eci}) and collect all expert contributions into a task-conditioned profile 
\begin{equation}
    \bar{G}_{\ell,\tau} = [c^{(\ell)}_{1,\tau}, \dots, c^{(\ell)}_{i,\tau}, \dots, c^{(\ell)}_{E,\tau}]^\top,
\end{equation}

This profile serves as the building block for both task-specificity
analysis (Stage~II) and expert contribution estimation (Stage~III).

\paragraph{Stage II: Quantifying task-specificity.}

We next assess the degree of routing specialization per domain task $\tau$ using a silhouette-based score. Stage II determines whether a task's routing is sufficiently distinctive.
Let $d(\cdot,\cdot)$ denote the cosine distance between routing vectors (as defined in Eq~(\ref{eq:vec})).
For each $x_i \in \mathcal{D}_\tau$, define
\begin{equation}
\begin{aligned}
a_i &=
\frac{1}{|\mathcal{D}_\tau|-1}
\sum_{\substack{x_j\in\mathcal{D}_\tau \\ j\ne i}}
d\!\left(G^{(\ell)}(x_i),\,G^{(\ell)}(x_j)\right), \\
b_i &=
\min_{\tau' \ne \tau}
\frac{1}{|\mathcal{D}_{\tau'}|}
\sum_{x_j\in\mathcal{D}_{\tau'}}
d\!\left(G^{(\ell)}(x_i),\,G^{(\ell)}(x_j)\right),
\end{aligned}
\end{equation}
and compute the sample-level silhouette
\begin{equation}
s_i \;=\; \frac{b_i-a_i}{\max(a_i,b_i)} \in [-1,1].
\end{equation}

The task-specificity score is the mean silhouette:
\begin{equation}
S_{\ell}(\tau) \;=\; \frac{1}{|\mathcal{D}_\tau|}\sum_{x_i\in\mathcal{D}_\tau} s_i .
\end{equation}
where high $S_\ell(\tau)$ indicates that routing vectors within domain task $\tau$
form a compact and clearly separated cluster. Only tasks with sufficiently
high $S_\ell(\tau)$ proceed to Standing Committee construction in Stage~III.

\paragraph{Stage III: Standing Committee exploration.}
We finally identify the \emph{domain-invariant backbone} for well-structured tasks. Given $c^{(\ell)}_{i,\tau}$ computed in Stage I, which corresponds to the average routing weight assigned to expert $i$ when processing samples from domain task $\tau$. Experts are then ranked within each domain task, and we denote the resulting rank by
\begin{equation}
R(i,\tau)^{(\ell)},
\end{equation}
assigning a penalty rank $k\!+\!1$ to experts that do not appear in the
Top-$k$. Here, $k$ corresponds to the model's routing sparsity.

To distinguish experts that are globally useful from those that are merely task-specific, we measure how often each expert appears among the Top-$k$ across domain tasks:
\begin{equation}
P_i^{(\ell)}
=
\frac{1}{|\mathcal{T}|}
\sum_{\tau \in \mathcal{T}}
\mathbf{I}\!\left(R(i,\tau)^{(\ell)} \le k\right).
\end{equation}
Experts that occur sufficiently frequently form the set of consensus
candidates $\mathcal{E}$:
\begin{equation}
\mathcal{E}^{(\ell)}
=
\left\{\, i \;:\; P_i^{(\ell)} \ge \gamma \right\},
\end{equation}
where $\gamma$ controls the threshold for cross-domain agreement. To ensure committee members are active in nearly all observed domain tasks, in this paper, we use $\gamma > 0.8$ to set up our experiment.

For each candidate expert, we further characterize stability across domain tasks:
\begin{equation}
\begin{aligned}
\mu_i^{(\ell)} &= \mathbf{E}_{\tau}\!\left[R(i,\tau)^{(\ell)}\right], \\
\sigma_i^{(\ell)} &= \mathrm{Var}_{\tau}\!\left[R(i,\tau)^{(\ell)}\right].
\end{aligned}
\end{equation}
Here, $\mu_i^{(\ell)}$ captures how highly the expert is ranked on average,
while $\sigma_i^{(\ell)}$ quantifies how consistently it holds that position.

Finally, the \emph{Standing Committee} at layer $\ell$ is defined as the
Pareto-optimal set:
\begin{equation}
\mathcal{C}^{(\ell)}
=
\operatorname{Pareto}\!\left(
\{(\mu_i^{(\ell)},\sigma_i^{(\ell)})\}_{i\in\mathcal{E}^{(\ell)}}
\right),
\end{equation}

This selection favors experts which achieve a favorable trade-off between high average rank and low cross-domain variability.

\subsection{Experimental Setup}

\subsubsection{Dataset}

We evaluate \textsc{CommitteeAudit} on the Massive Multitask Language Understanding
(MMLU) benchmark~\cite{hendrycks2020measuring}, which contains $57$
multiple–choice subjects spanning science, humanities, social science, and
professional domains.
To study domain-conditioned routing rather than per-subject idiosyncrasies, we reorganize all subjects into nine semantically coherent domains
(Table~\ref{tab:mmlu_domains}). To provide a more comprehensive evaluation of our method, we further extend the experimental benchmark to Colossal Clean Crawled Corpus~\cite{raffel2020exploring}, with additional dataset details deferred to Appendix~\ref{C4_data}.

\begin{table}[t]
\centering
\caption{Aggregation of MMLU subjects into nine domain tasks.}
\resizebox{\linewidth}{!}{
\begin{tabular}{lp{11cm}}
\toprule
\textbf{Domain} & \textbf{Representative Subjects} \\
\midrule
STEM--Math & algebra, geometry, probability, statistics, college/high-school mathematics \\
STEM--Physics & high-school/college physics, astronomy, conceptual physics \\
STEM--Chemistry & high-school and college chemistry \\
STEM--BioMed & biology, anatomy, genetics, clinical knowledge, virology, nutrition, aging, medicine \\
CS--Eng & computer science, security, operating systems, ML, electrical engineering \\
SocSci & economics, econometrics, sociology, psychology, political science, public relations \\
Humanities & history, philosophy, ethics, religion, art history, world facts \\
Lang--Ling & English, literature, linguistics \\
Biz--Law & business, management, accounting, marketing, law \\
\bottomrule
\end{tabular}}
\label{tab:mmlu_domains}
\end{table}

Formally, each subject is mapped to a domain task $\tau \in \mathcal{T}$, yielding
domain-specific subsets $\{\mathcal{D}_\tau\}$.
All routing analyses in this paper, including task-specificity and Standing Committee extraction, are conducted at the domain level.


\subsubsection{Model}

We evaluate \textsc{CommitteeAudit} on three representative MoE language models that differ in expert-pool size and routing
configuration. All models are used in inference-only mode, and we extract routing weights from every MoE layer for
analysis.

\begin{table}[t]
\centering
\caption{MoE configurations of evaluated models.} \label{tab:models}
\resizebox{\linewidth}{!}{
\begin{tabular}{lcccc}
\toprule
\textbf{Model} & \textbf{Experts ($E$)} & \textbf{Top-$k$} & \textbf{Shared} & \textbf{Size} \\
\midrule
DeepSeek-V2-Lite & 64 & 6 & 2 & 16B\\
Qwen3-30B-A3B & 128 & 8 & 0 & 30B\\
OLMoE-1B-7B & 64 & 8 & 0 & 7B\\
\bottomrule
\end{tabular}}
\end{table}

As shown in Table~\ref{tab:models}, DeepSeek-V2-Lite~\cite{deepseekai2024deepseekv2strongeconomicalefficient} includes two shared experts that are always active, forming a centralized processing path, whereas Qwen3-30B-A3B~\cite{yang2025qwen3technicalreport} and
OLMoE-1B-7B~\cite{muennighoff2025olmoeopenmixtureofexpertslanguage} rely purely on routed experts. The variation in $(E,k)$ and shared-expert usage allows us to probe whether Standing Committees are an architectural artifact or a persistent phenomenon across MoE designs (details are in Appendix~\ref{sec:com}).


\subsection{Metrics}
\subsubsection{Jaccard Similarity (Cross-Domain Expert Sharing)}

To quantify how much different domains reuse the same experts, we compute the Jaccard similarity between domain-level Top-$k$ expert sets. For layer
$\ell$ and domains $\tau_1,\tau_2$, let $\mathcal{E}_{\ell,\tau}$ denote the
Top-$k$ experts; then
\begin{equation}
\mathrm{Jaccard}_\ell(\tau_1,\tau_2)
=
\frac{
\left|\mathcal{E}_{\ell,\tau_1} \cap \mathcal{E}_{\ell,\tau_2}\right|
}{
\left|\mathcal{E}_{\ell,\tau_1} \cup \mathcal{E}_{\ell,\tau_2}\right|
}.
\end{equation}

Values near $0$ indicate domain-specific routing, whereas larger values
reflect substantial cross-domain expert sharing.

\subsubsection{Gini Coefficient (Expert Concentration)}

To quantify the inequality of expert contributions, we compute the Gini coefficient over the distribution of the Expert Contribution Index (ECI) at each layer $\ell$. Let $\bar{\mathbf{c}}^{(\ell)} = (\bar{c}_1^{(\ell)}, \dots, \bar{c}_E^{(\ell)})$ denote the global contribution vector, where $\bar{c}_i^{(\ell)} = \mathbf{E}_{\tau \in \mathcal{T}}[c_{i,\tau}^{(\ell)}]$ represents the average ECI for expert $i$ across all tasks. The Gini coefficient is defined as:
\begin{equation}
\mathrm{Gini}(\bar{\mathbf{c}}^{(\ell)}) = \frac{\sum_{i=1}^E \sum_{j=1}^E |\bar{c}_i^{(\ell)} - \bar{c}_j^{(\ell)}|}{2E \sum_{i=1}^E \bar{c}_i^{(\ell)}}.
\end{equation}

In this context, a Gini coefficient approaching $0$ indicates a uniform utilization of experts, where each expert provides an equal contribution to the model's computation. Conversely, a high Gini coefficient (approaching $1$) signals extreme contribution inequality, where the total routing mass is monopolized by a small subset of experts, providing macroscopic evidence for the existence of a standing committee.

\section{Experiment}
In this section, we present a series of experiments to address the three questions introduced in Section~\ref{intro}. These experiments allow us to determine whether standing committees actually emerge, how they evolve across depth, and what functional role they play in model behavior.

\subsection{Existence and Stability: The "Standing Committee" Phenomenon}
\textbf{Question 1:} Do routed experts converge into stable, domain-invariant groups?

\subsubsection{Jaccard–Gini Analysis of Expert Sharing and Concentration}

Table~\ref{tab:metrics_summary} evaluates the standing-committee hypothesis from two complementary perspectives: (i) whether the same experts tend to reappear across domains, and (ii) how unevenly routing mass is distributed among them. Despite substantial differences in expert capacity ($E$) and routing sparsity ($k$), all three models display high overlap and concentration, indicating that MoE routing tends to self-organize into "standing committees" rather than task-specific specialization.

\begin{table}[t]
\centering
\caption{Cross-domain sharing (Jaccard) and expert concentration (Gini) across models.}
\resizebox{\linewidth}{!}{
\begin{tabular}{l l c c c}
\toprule
\textbf{Metric} & \textbf{Statistic} & \textbf{OLMoE} & \textbf{DeepSeek-V2-Lite} & \textbf{Qwen3-30B-A3B} \\
\midrule
\multirow{3}{*}{Jaccard Similarity}
& Max     & 1.0000 & 1.0000 & 1.0000 \\
& Min     & 0.7963 & 0.7103 & 0.5300 \\
& Overall & 0.8735 & 0.8670 & 0.8670 \\
\midrule
\multirow{3}{*}{Gini Coefficient}
& Max     & 0.9082 & 0.9360 & 0.9605 \\
& Min     & 0.8814 & 0.9092 & 0.9405 \\
& Overall & 0.8957 & 0.9207 & 0.9465 \\
\bottomrule
\end{tabular}
}
\label{tab:metrics_summary}
\end{table}

\paragraph{Membership Stability (Jaccard Similarity).}
The Jaccard index captures whether the same experts repeatedly appear among the top-$k$ routed set across domains. OLMoE ($E=64, k=8$) achieves the highest mean overlap (0.8735) and the strongest minimum stability (0.7963), suggesting that the model frequently reuses a common subset of experts. Qwen3 ($E=128,k=8$) shows greater local variability (Min: 0.5300), yet its high global average (0.8670) indicates that such deviations occur on top of a largely stable routing structure rather than replacing it entirely.

\paragraph{Contribution Concentration (Gini Coefficient).}
The Gini coefficient quantifies the inequality of ECI across the expert population. All models exhibit extreme values
($>0.88$), meaning that a small fraction of experts absorbs most of the routing mass. Interestingly, concentration correlates with expert capacity: Qwen3 ($E=128$) attains the highest overall Gini (0.9465). Rather than distributing computation more broadly, larger pools appear to amplify the dominance of a compact set of frequently selected experts.

\begin{figure*}[t]
\centering

\begin{subfigure}{0.88\linewidth}
  \centering
  \includegraphics[width=\linewidth]{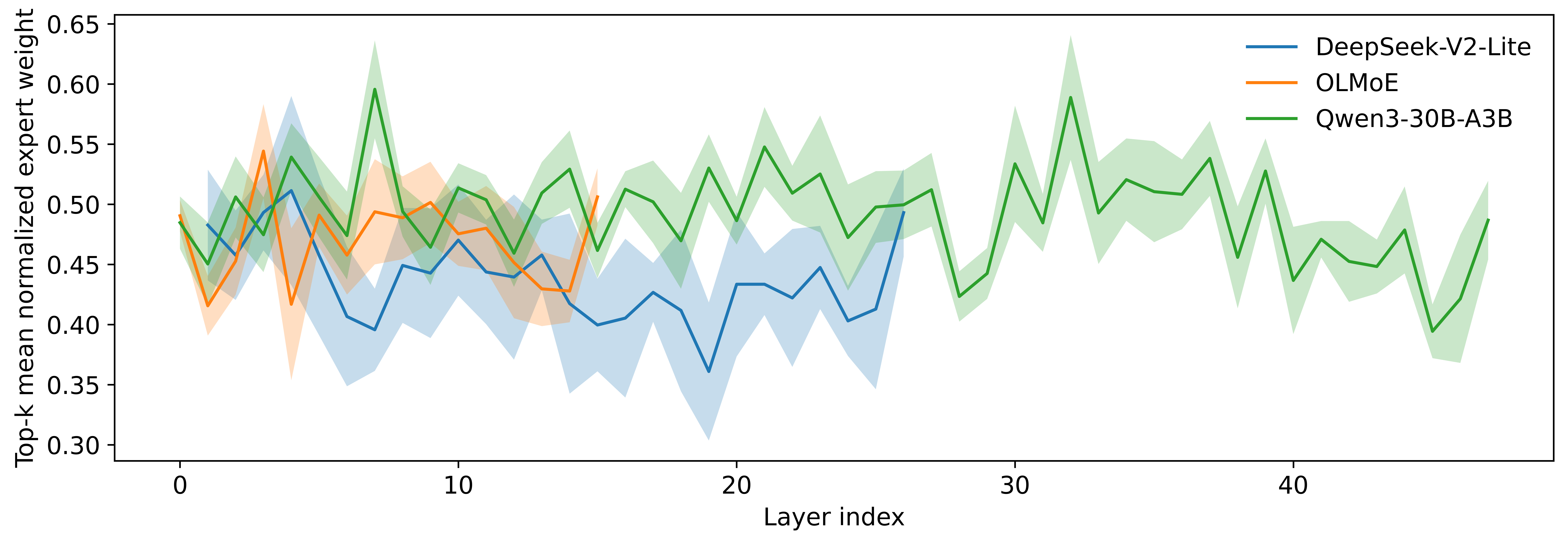}
  \caption{Standing Committee Stability Across Layers.}
\end{subfigure}

\vspace{0.35em}

\begin{subfigure}{0.28\linewidth}
  \centering
  \includegraphics[width=\linewidth]{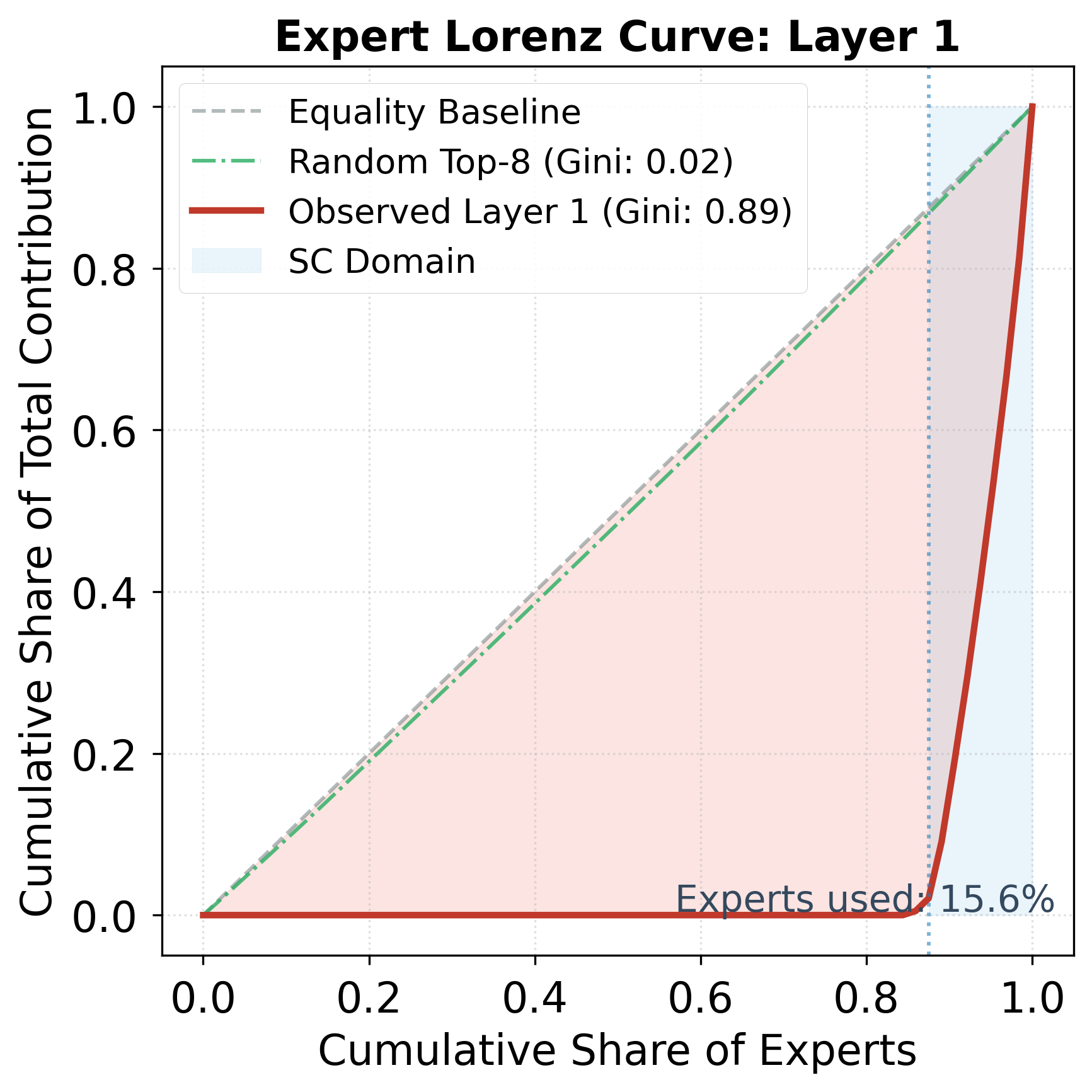}
  \caption{OLMoE expert concentration.}
\end{subfigure}
\hfill
\begin{subfigure}{0.28\linewidth}
  \centering
  \includegraphics[width=\linewidth]{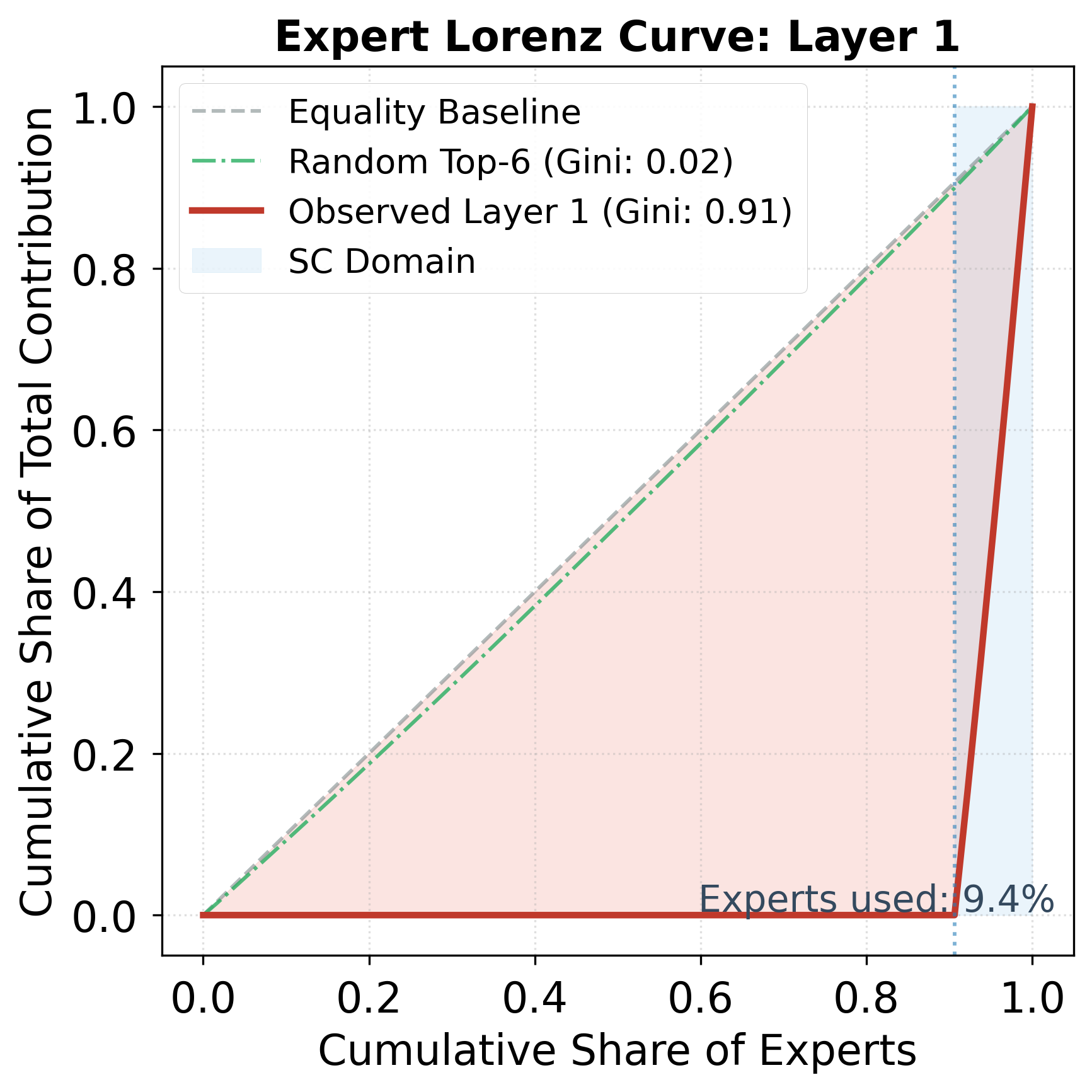}
  \caption{DeepSeek expert concentration.}
\end{subfigure}
\hfill
\begin{subfigure}{0.28\linewidth}
  \centering
  \includegraphics[width=\linewidth]{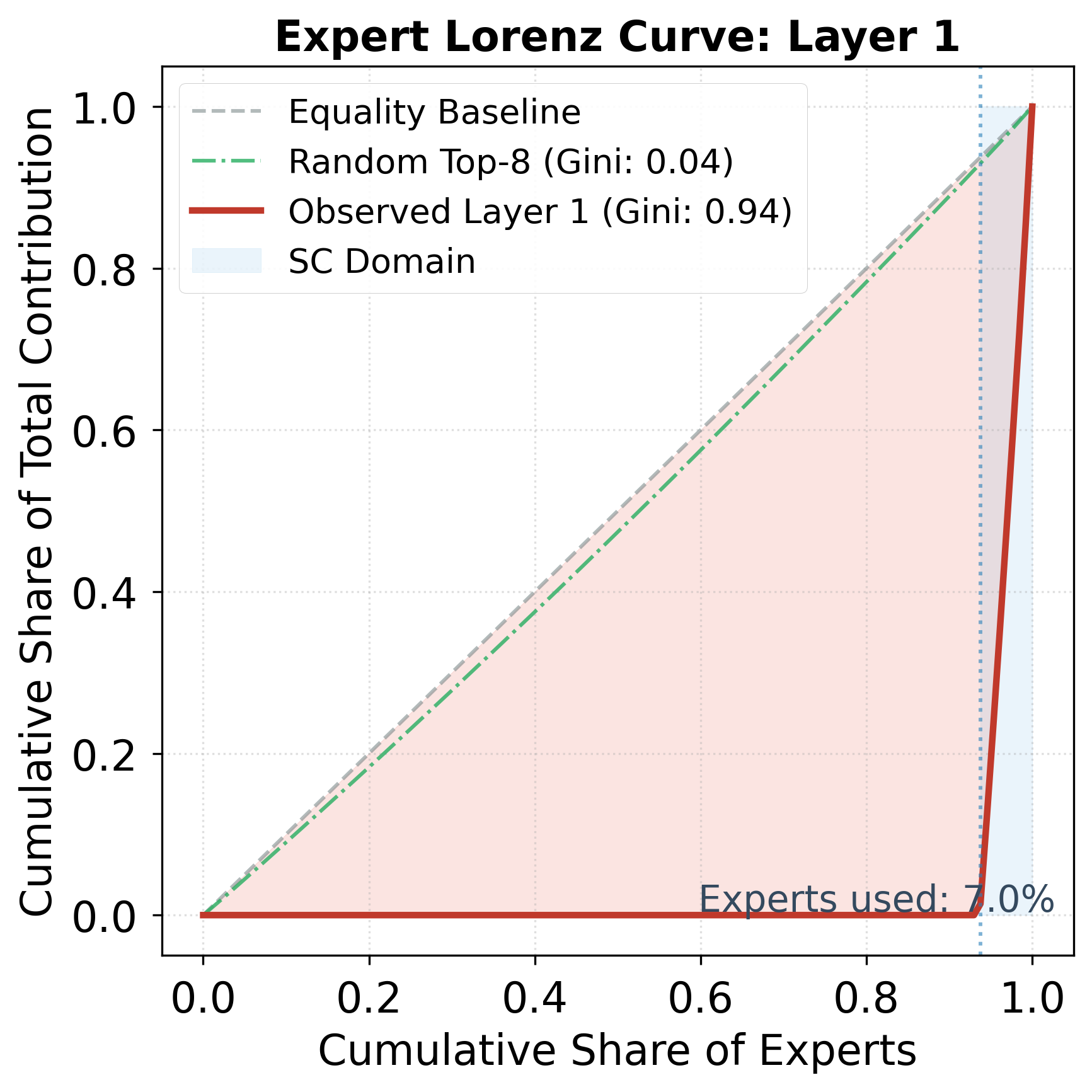}
  \caption{QWen expert concentration.}
\end{subfigure}

\caption{
Evidence of standing committees in MoE models. 
(a) Layer-wise concentration of top-k experts across tasks. For each model, the solid line shows the mean normalized weight assigned to the top-$k$ experts at each layer, and the shaded region denotes one standard deviation. High and stable values indicate a small subset of experts (``standing committees'') consistently absorbs most routing mass. 
(b–d) Lorenz curves reveal that only a small subset of experts accounts for most contributions (other Lorenz curves are shown in Appendix~\ref{Con_con_ana}), showing that these committees are highly centralized rather than uniformly shared.
}
\label{fig:committee_layout}
\end{figure*}

Figure~\ref{fig:committee_layout} links these statistics to routing behavior. In Panel (a), the mean normalized weight assigned to the routed top-$k$ experts remains both high and stable across layers. If experts were mainly task-specialized, different domains would activate different experts, and the variance bands would widen. Instead, we observe persistent dominance by the same routed subset, indicating a domain-invariant backbone.
Panels (b–d) arrive at the same conclusion from a distributional view: Lorenz curves show that only a tiny fraction of experts accounts for most
ECI, confirming a strongly centralized allocation of computation.

Together, the two views support the Standing Committee hypothesis: MoE
models concentrate routing mass onto a small, persistent core, while most
experts operate only peripherally.

\subsubsection{Analyzing the Stability and Contribution of Standing Committees}

Table~\ref{tab:standing_committees_full} summarizes representative standing
committees based on the Pareto-optimal set across depth. Committee members consistently occupy very high
routing positions (Avg.~$\mu \approx 3.1$--$3.8$) with low rank variability
($\sigma^2 \le 3.44$). For example, the middle-layer committee in OLMoE exhibits a variance of only $0.49$, indicating that these experts remain near the top of the routing hierarchy regardless of domain. Rather than transient specialists, they function as a de-facto backbone that the model repeatedly relies on. For additional transparency, Appendix~\ref{illustrative_exp} provides an illustrative expert-level table showing domain coverage and ECI statistics for representative DeepSeek-V2-Lite layers. Complete layer-wise MMLU results are provided in Appendix~\ref{sec:sc_all_layer}, Section~\ref{mmlu_re}, while matched supplementary results on the C4-based evaluation subset are reported in Appendix~\ref{sec:sc_all_layer}, Section~\ref{c4_re}.

\begin{table}[t]
\centering
\small
\caption{An audit of representative Standing Committees ($\mathcal{C}$) across network phases, showing shallow, middle, and deep layers for each model. \textbf{Avg. $\mu$} and \textbf{Avg. $\sigma^2$} represent the mean and the variance of ranks of the committee members across domains. \textbf{ECI Cov.} is the cumulative contribution, and \textbf{Ratio} indicates the influence density vs. a uniform baseline.}
\label{tab:standing_committees_full}

\resizebox{\linewidth}{!}{
\setlength{\tabcolsep}{6pt}
\begin{tabular}{llclccccc}
\toprule
\textbf{Model} & \textbf{Phase} & \textbf{Layer ($\ell$)} & \textbf{Committee Members ($\mathcal{C}$)} & \textbf{$|\mathcal{C}|$} & \textbf{Avg. $\mu$ ↓} & \textbf{Avg. $\sigma^2$ ↓} & \textbf{ECI Cov.} & \textbf{Ratio ($\times$)} \\ 
\midrule
\multirow{3}{*}{\textbf{DeepSeek-V2-Lite}} 
& Shallow & 3  & \{7, 13, 22, 42\} & 4 & 3.36 & 1.81 & 66.3\% & 29.5$\times$ \\
& Middle  & 11 & \{28, 43, 54\} & 3 & 3.15 & 1.98 & 60.7\% & 31.4$\times$ \\
& Deep    & 19 & \{4, 14, 47, 61\} & 4 & 3.11 & 0.76 & 70.5\% & 35.8$\times$ \\
\midrule
\multirow{3}{*}{\textbf{OLMoE}} 
& Shallow & 2  & \{30, 58, 63\} & 3 & 3.41 & 2.15 & 43.9\% & 15.9$\times$ \\
& Middle  & 8  & \{13, 45\} & 2 & 3.28 & 0.49 & 29.7\% & 13.1$\times$ \\
& Deep    & 16 & \{17, 52, 60\} & 3 & 3.19 & 1.52 & 44.0\% & 16.0$\times$ \\
\midrule
\multirow{3}{*}{\textbf{Qwen3-30B-A3B}} 
& Shallow & 3  & \{38, 40, 80, 93\} & 4 & 3.61 & 3.44 & 54.0\% & 36.4$\times$ \\
& Middle  & 33 & \{16, 26, 57, 116, 121\} & 5 & 3.82 & 2.16 & 67.0\% & 49.9$\times$ \\
& Deep    & 46 & \{94, 101, 107\} & 3 & 3.15 & 1.59 & 50.9\% & 43.3$\times$ \\
\bottomrule
\end{tabular}
}
\end{table}

Although $|\mathcal{C}|$ remains small (2-5) across all models, these groups capture up to $70.5\%$ of total routing mass. Importantly, the size of $\mathcal{C}$ remains stable even as capacity increases from $E=64$ to $E=128$. This suggests that committee-like behavior is not an artifact of a particular architecture, but an emergent pattern of sparse routing optimization.






\subsection{Variation and Sensitivity: Structural Dynamics}

\textbf{Question 2:} How does group structure evolve with depth, and is centralization inevitable under sparse routing?

\subsubsection{Robustness of Standing Committees}

\begin{figure}[t]
\centering
\includegraphics[width=\linewidth]{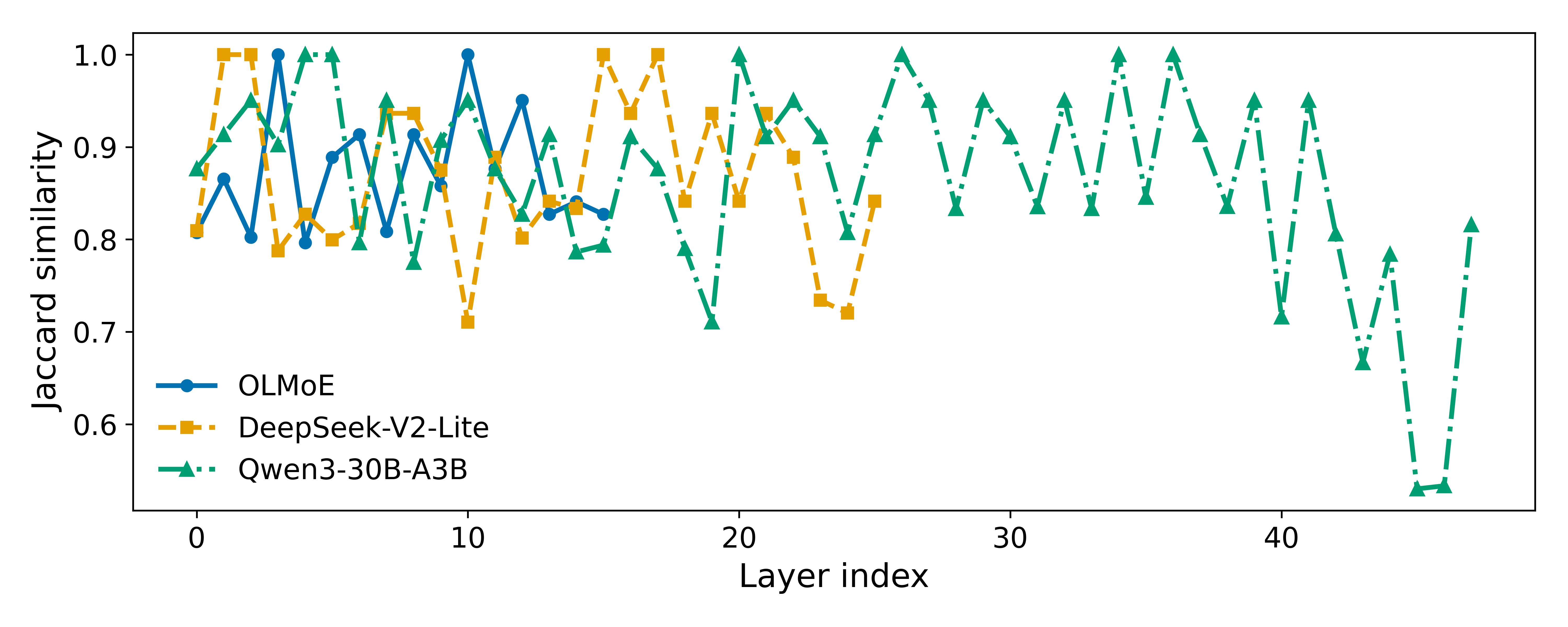}
\caption{
Cross-layer stability of routed experts across models, measured by Jaccard
similarity between top-$k$ expert sets over domains.
All three MoE models maintain high overlap ($\geq 0.8$ for most layers),
showing that the same experts are repeatedly selected despite changes in
input domain and network depth.
}
\label{fig:jaccard_models}
\end{figure}

We begin by examining whether routed experts actually form persistent groups.
Figure~\ref{fig:jaccard_models} reports the Jaccard similarity of top-$k$
expert sets across domains for each layer. Across all three MoE models, the
similarity remains consistently high (often $\ge 0.85$), showing that the
same experts are repeatedly activated across tasks and depths. Rather than
rotating specialists, the routing network converges to a shared backbone of
experts that is largely invariant to both input domain and layer position.

\begin{figure}[t]
\centering
\begin{subfigure}{.85\linewidth}
  \centering
  \includegraphics[width=\linewidth]{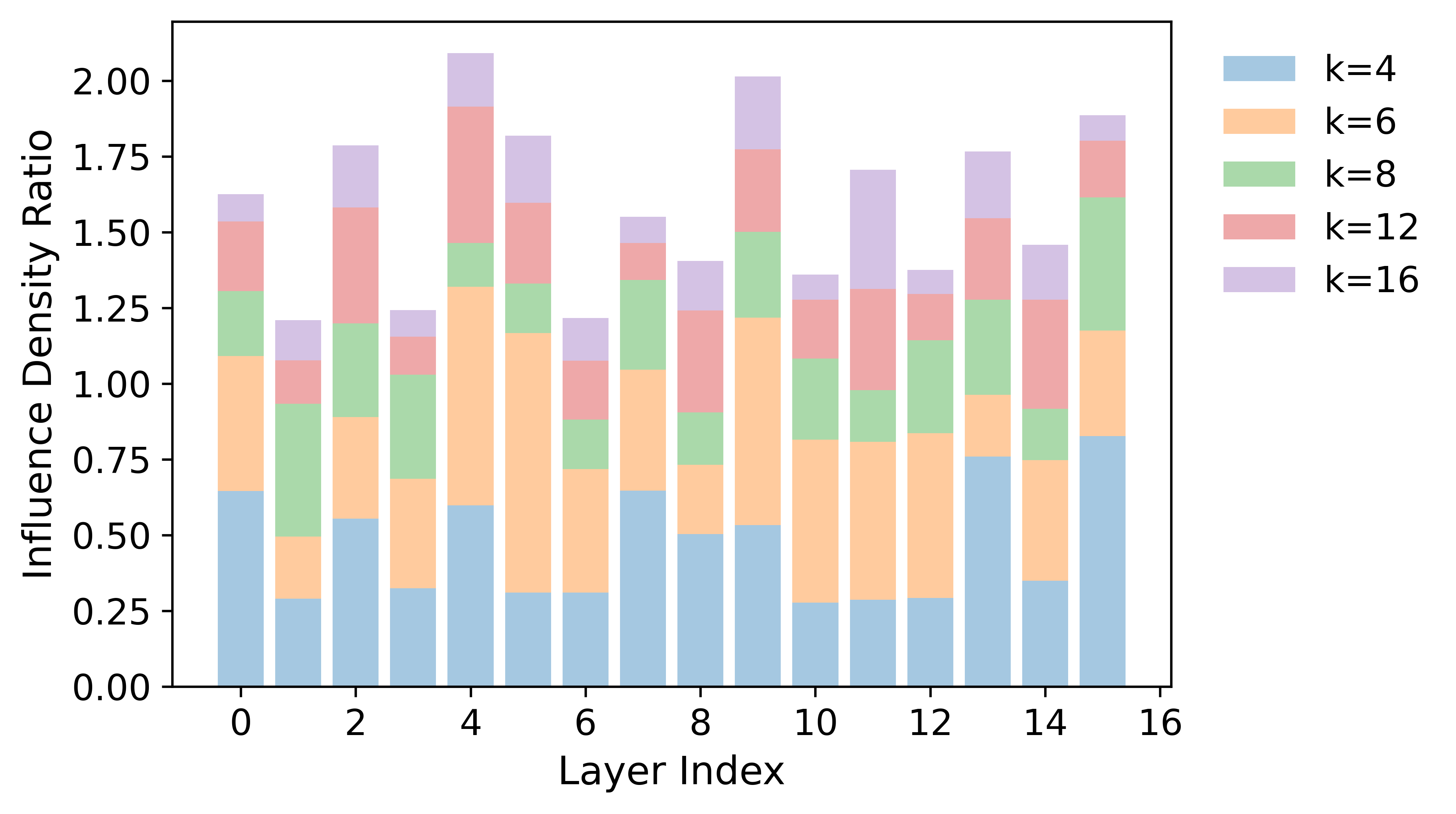}
  \caption{Influence Density Ratio across layers and routing budgets $k$.}
\end{subfigure}

\begin{subfigure}{.85\linewidth}
  \centering
  \includegraphics[width=\linewidth]{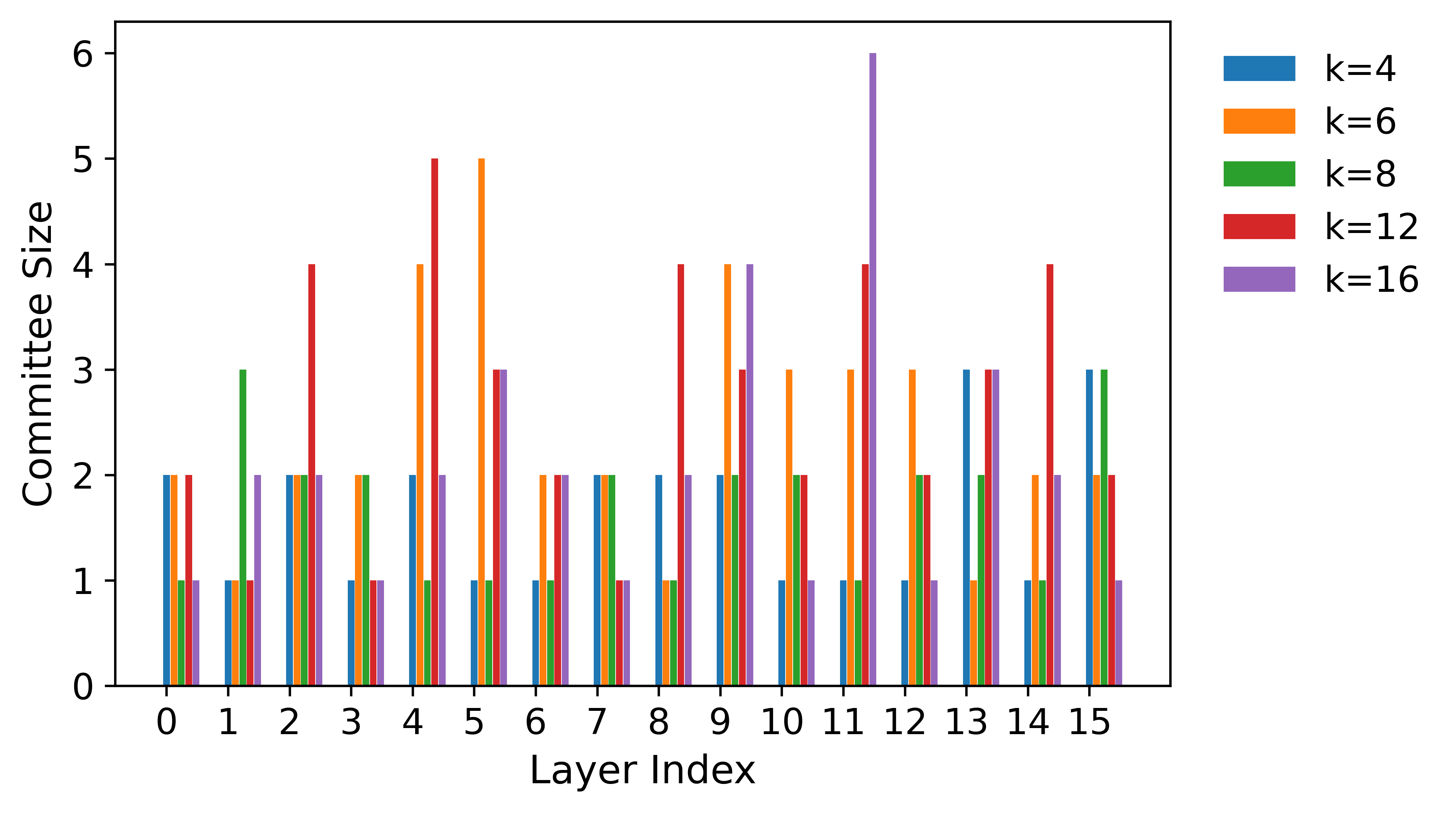}
  \caption{Committee size $|\mathcal{C}|$ across layers and routing budgets $k$.}
\end{subfigure}
\caption{
Dynamics of Standing Committees in OLMoE under different routing budgets.
We show all 16 layers.
(a) The Influence Density Ratio measures the average per-expert contribution of committee members relative to that of non-committee experts. Values greater than 1 indicate that committee members absorb disproportionately more routing mass per expert. Increasing $k$ does not eliminate this concentration.
(b) The size of the identified committees stays small and changes only mildly with depth and $k$, indicating a compact but persistent core of experts.
}
\label{fig:committee_dynamics}
\end{figure}

Having established the existence of a persistent backbone, we next ask how it evolves with depth and routing sparsity. Figure~\ref{fig:committee_dynamics} analyzes OLMoE under different routing budgets $k$.

In Figure~\ref{fig:committee_dynamics}(a), the Influence Density Ratio remains consistently above 1 across layers and routing budgets. This shows that, on a per-expert basis, committee members absorb substantially more routing mass than non-committee experts. Importantly, increasing $k$ does not flatten this imbalance. Instead, additional routing capacity mainly introduces peripheral experts with relatively low contribution density, while the committee continues to dominate the allocation of routing mass.

Figure~\ref{fig:committee_dynamics}(b) shows that the committee size itself remains small, typically around 1-4 experts, and changes only mildly across depth and routing budgets. Thus, MoE models do not substantially expand the committee as depth increases or routing capacity grows; they instead rely on a compact, persistent core whose influence remains structurally robust across different values of $k$.
Together with Figure~\ref{fig:jaccard_models}, these findings indicate that centralization is not accidental but an emergent structural property of sparse routing.

\subsubsection{Top-$k$ Sensitivity Sweep}

To probe how sensitive the standing committee is to the routing budget, we
perform a top-$k$ sensitivity sweep on OLMoE. For each setting of $k \in \{4,6,8,12,16\}$, we re-identify the standing committee based on the Pareto-optimal set and compute the retention rate with respect to the reference committee at $k{=}8$, i.e., the fraction of $k{=}8$ core members that remain in the new committee.

\begin{figure}[t]
\centering
\includegraphics[width=0.9\linewidth]{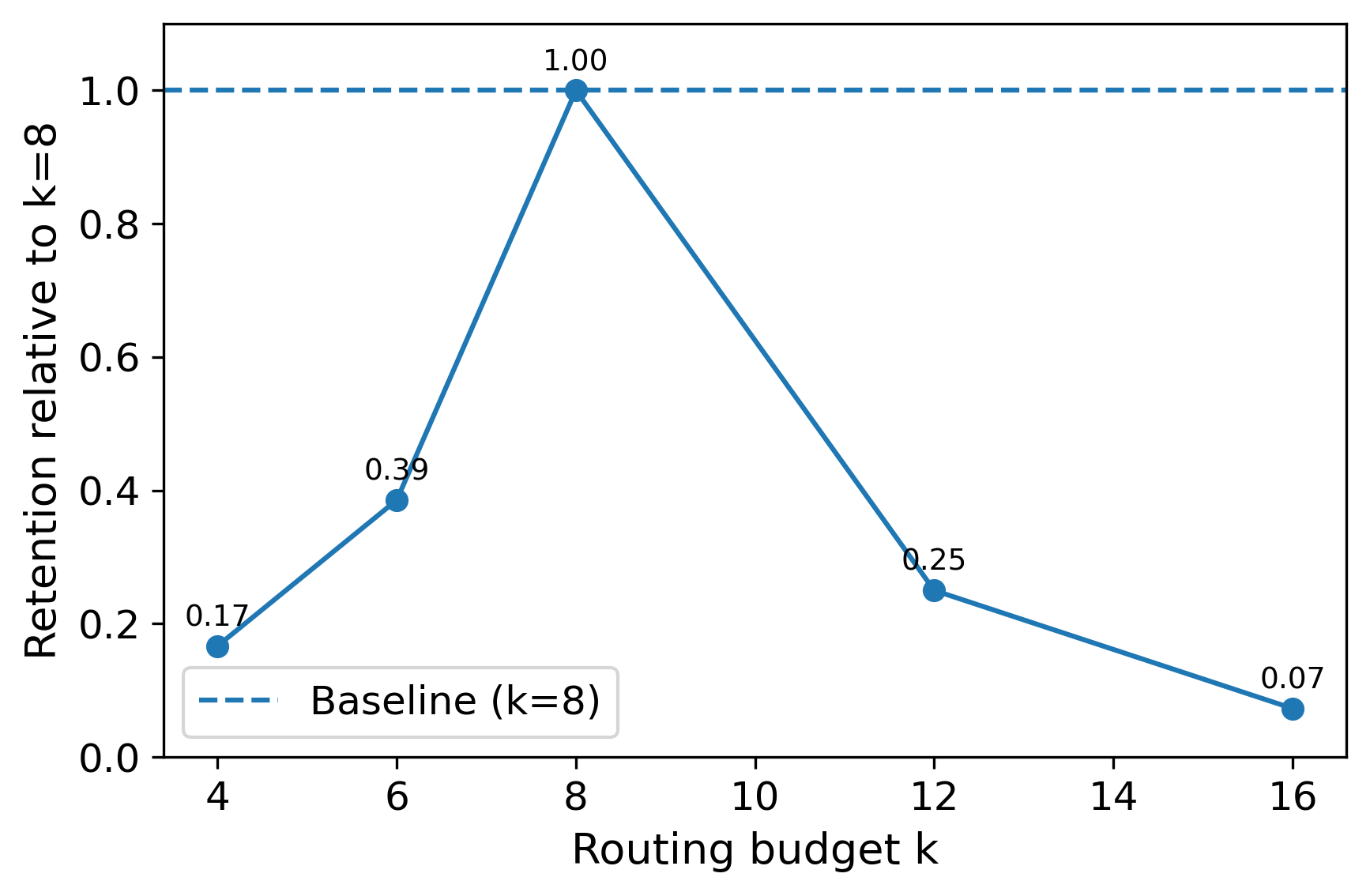}
\caption{
Top-$k$ sensitivity sweep for OLMoE.
We take the standing committee identified at $k\!=\!8$ as the reference core
and measure, for each routing budget $k$, the fraction of its members that
are still present in the new committee (averaged over layers).
The dashed line marks the $k\!=\!8$ baseline (retention $=1.0$).
}
\label{fig:olmoe_topk_sweep}
\end{figure}

Figure~\ref{fig:olmoe_topk_sweep} shows that retention peaks at the nominal
configuration $k{=}8$ and drops on both sides: it falls to $0.39$ at $k{=}6$, $0.17$ at $k{=}4$, and below $0.3$ once $k$ is expanded to $12$ or
$16$. This pattern suggests that the core experts are not an artifact of a single $k$ choice, but they are also not completely rigid. When $k$ is too small, the gate is forced to exclude part of the original core; when $k$ is too large, the gate dilutes its attention and recruits additional experts, replacing some core members.

Overall, the sweep indicates that sparse routing induces a centralized committee around $k{=}8$, but that committee can partially reorganize as the routing budget becomes substantially more aggressive or more constrained.



\subsection{Interaction and Behavior: Functional Interpretation}
\textbf{Question 3:} What roles do stable expert groups play in reasoning versus domain knowledge?

\begin{figure}[t]
\centering

\begin{subfigure}{0.9\linewidth}
    \centering
    \includegraphics[width=\linewidth]{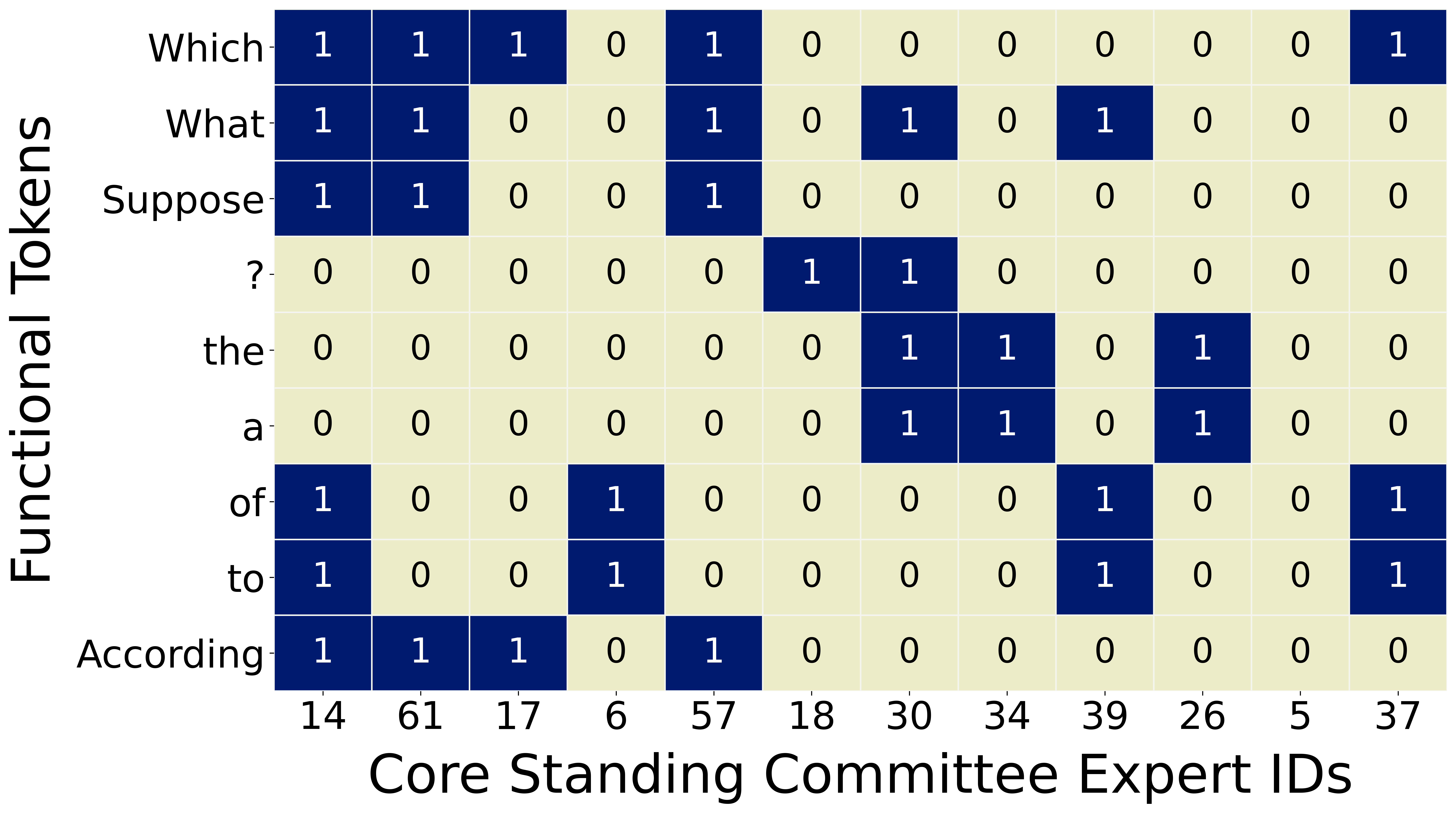}
    \caption{OLMoE: functional tokens repeatedly map to a fixed subset of committee experts.}
    \label{fig:case_olmoe}
\end{subfigure}

\begin{subfigure}{0.9\linewidth}
    \centering
    \includegraphics[width=\linewidth]{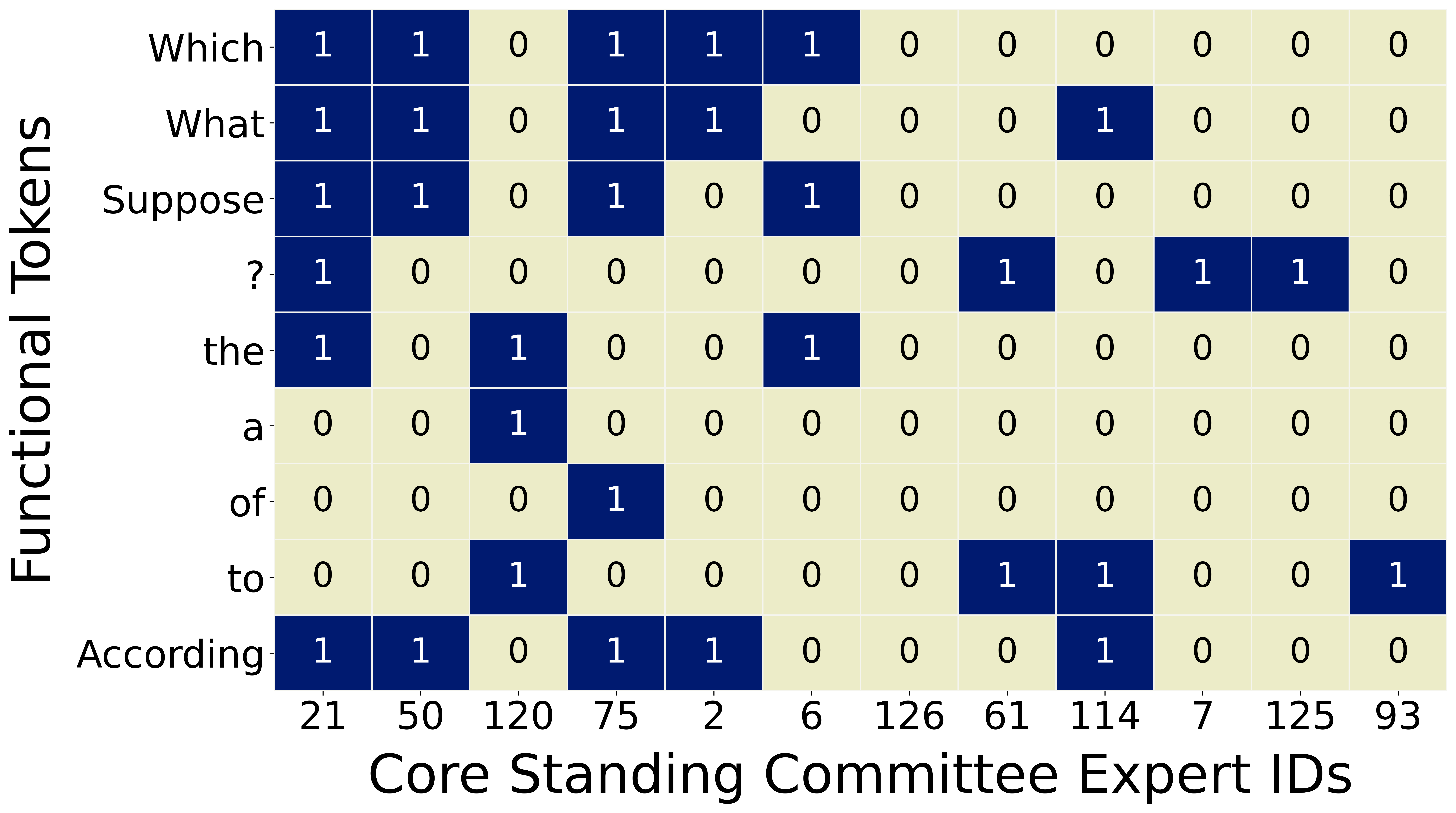}
    \caption{Qwen3-30B-A3B shows similar convergence despite larger expert capacity.}
    \label{fig:case_qwen}
\end{subfigure}

\caption{
Case study of token-level routing. Rows denote functional tokens and columns denote members of the Standing Committee. A cell is marked when a token reliably activates an expert across domains.
}
\label{fig:case_study}
\end{figure}

We illustrate the functional roles of Standing Committees using a qualitative case study. 
Figure~\ref{fig:case_study} shows an activation matrix between committee experts (columns) and functional tokens (rows). 
A cell is marked when a token repeatedly activates an expert in at least three domains. 
Two consistent behaviors emerge.

\paragraph{Anchor 1: Logical framing and reasoning control.}
Across OLMoE and Qwen3-30B-A3B, abstract reasoning triggers such as \textit{Which}, \textit{What}, \textit{Suppose}, and question marks are routed to the same subset of committee experts.
These tokens define the logical scaffolding of the prompt, suggesting that the committee acts as a reasoning controller.

\paragraph{Anchor 2: Domain-invariant syntactic backbone.}
High-frequency structural tokens, including \textit{the}, \textit{a}, and \textit{in}, also converge to overlapping committee members across domains, indicating that the committee maintains a stable syntactic layer independent of content.

\paragraph{Peripheral specialization.}
By contrast, domain-specific terminology rarely stabilizes: chemical symbols, biomedical identifiers, and financial jargon are distributed across many experts depending on context. 
This pattern supports a core-periphery organization in which the committee anchors reasoning and syntax, while peripheral experts are recruited on demand for specialized knowledge.

Taken together, Standing Committees function as a domain-invariant control layer, coordinating logical structure and grammar while delegating domain knowledge to peripheral experts.

\subsection{Masking-Based Evidence of Functional Importance}

To complement our observational analysis, we conduct a masking-based intervention on DeepSeek-V2-Lite to test whether identified Standing Committee experts are functionally important for answer generation. For each selected layer, we suppress the routing weights of the Standing Committee experts, re-normalize the remaining routing distribution, and then evaluate answer outcomes on MMLU. Table~\ref{tab:committee_masking_main} reports representative shallow, middle, and deep layers, while the full results are provided in Appendix~\ref{masking_exp}.

Masking committee experts consistently degrades performance. The unmasked baseline achieves a correct rate of $0.39$, which drops to $0.12$, $0.09$, and $0.03$ when masking representative shallow, middle, and deep committee layers, respectively. At the same time, the proportion of \textit{No Answer} responses rises sharply from $0.03$ to $0.36$--$0.38$. These results provide intervention-based evidence that Standing Committee experts are not merely statistically frequent routing destinations, but functionally important components of the model's computational backbone.

\begin{table}[t]
\centering
\small
\caption{Representative masking-based intervention results on DeepSeek-V2-Lite. For each selected layer, we suppress the routing weights of the identified Standing Committee experts, re-normalize the remaining routing distribution, and evaluate answer outcomes on MMLU.}
\label{tab:committee_masking_main}

\resizebox{\linewidth}{!}{
\begin{tabular}{l l c c c}
\toprule
\textbf{Phase} & \textbf{Mask Layer} & \textbf{Correct} & \textbf{Wrong} & \textbf{No Answer} \\
\midrule
Baseline & None & 0.39 & 0.58 & 0.03 \\
Shallow  & 2    & 0.12 & 0.52 & 0.36 \\
Middle   & 10   & 0.09 & 0.55 & 0.36 \\
Deep     & 26   & 0.03 & 0.59 & 0.38 \\
\bottomrule
\end{tabular}
}
\end{table}

\section{Conclusion}
This work introduces \textsc{COMMITTEEAUDIT} and shows that MoE models rely on a domain-invariant \textbf{Standing Committee} that anchors reasoning and syntax, while peripheral experts handle domain knowledge. Our cross-model analysis suggests that this centralized structure arises from sparse routing itself rather than from specific architectural choices. This structural bias also reveals a tension with load-balancing objectives that encourage uniform expert usage, motivating function-aware routing and architectures that explicitly support a core-periphery organization of expertise.

\section*{Limitations}
\label{limit}

This work introduces \textsc{COMMITTEEAUDIT} and reports evidence for a domain-invariant Standing Committee in Mixture-of-Experts models. However, several limitations remain.

First, our analysis covers only a small set of representative MoE architectures and settings, and does not span hybrid, hierarchical, or dynamically adaptive routing designs. Whether similar organizational patterns persist in broader systems remains an open question.

Second, although we supplement our observational analysis with masking-based interventions on representative layers, our study still falls short of a fully controlled causal analysis. In particular, we do not systematically compare committee masking against alternative controls such as random or frequency-matched non-committee masking across models and tasks.

Third, our evaluation is centered primarily on domain-level analyses over MMLU, with supplementary evidence from C4 and intervention-based experiments. While this reduces subject-level noise and broadens the empirical picture, it may not fully capture behavior in conversational, multi-step reasoning, coding, or tool-augmented scenarios.

Finally, \textsc{COMMITTEEAUDIT} focuses on routing statistics rather than training dynamics. Understanding when and how Standing Committees emerge during optimization remains an important direction for future work.

\section*{Potential Risks}
\label{sec:sec_potential}

\paragraph{Potential Positive Impacts.}
By revealing group-level routing structure, this work may inform more transparent and efficient MoE design, support diagnosis of routing failures, and encourage principled interpretability research for sparse models.

\paragraph{Potential Negative Impacts.}
However, several risks remain. First, insights into centralized computation could be misinterpreted as evidence of inherent safety or robustness, leading to overreliance in deployment. Second, identifying persistent expert coalitions may enable adversarial targeting of critical routing pathways. Third, benchmarks may become over-optimized toward interpretability metrics without improving real-world safety.

These findings should therefore be used as analytical tools rather than deployment guarantees.

\section*{Ethical Considerations and Usage Disclaimer}
\label{sec:sec_ethical}

All experiments use publicly available models and datasets without personal or sensitive information. This work is intended for academic and educational purposes only and does not constitute guidance for production deployment.

The framework exposes structural properties of MoE systems, but does not certify fairness, safety, robustness, or regulatory compliance. The authors make no warranties regarding completeness or suitability for downstream use, and any application to high-stakes settings should involve domain experts, risk assessment, and human oversight.

We acknowledge that AI-assisted tools were used during writing and editing (e.g., grammar checking, phrasing refinement, and formatting suggestions). These tools were not used to generate research ideas, experimental results, model outputs, or claims, and all technical content, analyses, and conclusions were designed, verified, and interpreted by the authors.

\section*{Licenses and Terms of Use}
All datasets and pre-trained models used in this work are publicly available and redistributed under their respective licenses. We respect the original terms of use for each artifact. MMLU is used under its public research license, and all evaluated MoE models (OLMoE, Qwen3-30B-A3B, and DeepSeek-V2-Lite) are accessed and used in compliance with their published licenses. We do not redistribute any third-party artifacts.

\section*{Acknowledgments}
The authors acknowledge \textit{The Fin AI} community for its research support, feedback, and collaborative environment that contributed to this work.


\bibliography{custom}

@inproceedings{lo2025closer,
  title={A closer look into mixture-of-experts in large language models},
  author={Lo, Ka Man and Huang, Zeyu and Qiu, Zihan and Wang, Zili and Fu, Jie},
  booktitle={Findings of the Association for Computational Linguistics: NAACL 2025},
  pages={4427--4447},
  year={2025}
}

@article{olson2025probing,
  title={Probing Semantic Routing in Large Mixture-of-Expert Models},
  author={Olson, Matthew Lyle and Ratzlaff, Neale and Hinck, Musashi and Luo, Man and Yu, Sungduk and Xue, Chendi and Lal, Vasudev},
  journal={arXiv preprint arXiv:2502.10928},
  year={2025}
}

@article{yang2025mixture,
  title={Mixture of experts made intrinsically interpretable},
  author={Yang, Xingyi and Venhoff, Constantin and Khakzar, Ashkan and de Witt, Christian Schroeder and Dokania, Puneet K and Bibi, Adel and Torr, Philip},
  journal={arXiv preprint arXiv:2503.07639},
  year={2025}
}

@article{li2024your,
  title={Your mixture-of-experts llm is secretly an embedding model for free},
  author={Li, Ziyue and Zhou, Tianyi},
  journal={arXiv preprint arXiv:2410.10814},
  year={2024}
}

@article{swamy2024intrinsic,
  title={Intrinsic user-centric interpretability through global mixture of experts},
  author={Swamy, Vinitra and Montariol, Syrielle and Blackwell, Julian and Frej, Jibril and Jaggi, Martin and K{\"a}ser, Tanja},
  journal={arXiv preprint arXiv:2402.02933},
  year={2024}
}

@inproceedings{bai2025understanding,
  title={Understanding and leveraging the expert specialization of context faithfulness in mixture-of-experts llms},
  author={Bai, Jun and Tong, Minghao and Liu, Yang and Jia, Zixia and Zheng, Zilong},
  booktitle={Proceedings of the 2025 Conference on Empirical Methods in Natural Language Processing},
  pages={21938--21953},
  year={2025}
}

@article{xue2024openmoe,
  title={Openmoe: An early effort on open mixture-of-experts language models},
  author={Xue, Fuzhao and Zheng, Zian and Fu, Yao and Ni, Jinjie and Zheng, Zangwei and Zhou, Wangchunshu and You, Yang},
  journal={arXiv preprint arXiv:2402.01739},
  year={2024}
}

@article{jiang2024mixtral,
  title={Mixtral of experts},
  author={Jiang, Albert Q and Sablayrolles, Alexandre and Roux, Antoine and Mensch, Arthur and Savary, Blanche and Bamford, Chris and Chaplot, Devendra Singh and Casas, Diego de las and Hanna, Emma Bou and Bressand, Florian and others},
  journal={arXiv preprint arXiv:2401.04088},
  year={2024}
}

@article{zoph2022st,
  title={St-moe: Designing stable and transferable sparse expert models},
  author={Zoph, Barret and Bello, Irwan and Kumar, Sameer and Du, Nan and Huang, Yanping and Dean, Jeff and Shazeer, Noam and Fedus, William},
  journal={arXiv preprint arXiv:2202.08906},
  year={2022}
}

@misc{fan2024empiricalunderstandingmoedesign,
      title={Towards an empirical understanding of MoE design choices}, 
      author={Dongyang Fan and Bettina Messmer and Martin Jaggi},
      year={2024},
      eprint={2402.13089},
      archivePrefix={arXiv},
      primaryClass={cs.LG},
      url={https://arxiv.org/abs/2402.13089}, 
}

@article{zheng2024efficiently,
  title={Efficiently democratizing medical llms for 50 languages via a mixture of language family experts},
  author={Zheng, Guorui and Wang, Xidong and Liang, Juhao and Chen, Nuo and Zheng, Yuping and Wang, Benyou},
  journal={arXiv preprint arXiv:2410.10626},
  year={2024}
}

@article{ghandeharioun2024patchscopes,
  title={Patchscopes: A unifying framework for inspecting hidden representations of language models},
  author={Ghandeharioun, Asma and Caciularu, Avi and Pearce, Adam and Dixon, Lucas and Geva, Mor},
  journal={arXiv preprint arXiv:2401.06102},
  year={2024}
}

@article{chi2022representation,
  title={On the representation collapse of sparse mixture of experts},
  author={Chi, Zewen and Dong, Li and Huang, Shaohan and Dai, Damai and Ma, Shuming and Patra, Barun and Singhal, Saksham and Bajaj, Payal and Song, Xia and Mao, Xian-Ling and others},
  journal={Advances in Neural Information Processing Systems},
  volume={35},
  pages={34600--34613},
  year={2022}
}

@inproceedings{do2025simsmoe,
  title={SimSMoE: Toward Efficient Training Mixture of Experts via Solving Representational Collapse},
  author={Do, Giang and Le, Hung and Tran, Truyen},
  booktitle={Findings of the Association for Computational Linguistics: NAACL 2025},
  pages={2012--2025},
  year={2025}
}

@article{dai2024deepseekmoe,
  title={Deepseekmoe: Towards ultimate expert specialization in mixture-of-experts language models},
  author={Dai, Damai and Deng, Chengqi and Zhao, Chenggang and Xu, RX and Gao, Huazuo and Chen, Deli and Li, Jiashi and Zeng, Wangding and Yu, Xingkai and Wu, Yu and others},
  journal={arXiv preprint arXiv:2401.06066},
  year={2024}
}

@misc{su2025unveilingsuperexpertsmixtureofexperts,
      title={Unveiling Super Experts in Mixture-of-Experts Large Language Models}, 
      author={Zunhai Su and Qingyuan Li and Hao Zhang and Weihao Ye and Qibo Xue and YuLei Qian and Yuchen Xie and Ngai Wong and Kehong Yuan},
      year={2025},
      eprint={2507.23279},
      archivePrefix={arXiv},
      primaryClass={cs.CL},
      url={https://arxiv.org/abs/2507.23279}, 
}

@misc{deepseekai2025deepseekv3technicalreport,
      title={DeepSeek-V3 Technical Report}, 
      author={DeepSeek-AI and Aixin Liu and Bei Feng and Bing Xue and Bingxuan Wang and Bochao Wu and Chengda Lu and Chenggang Zhao and Chengqi Deng and Chenyu Zhang and Chong Ruan and Damai Dai and Daya Guo and Dejian Yang and Deli Chen and Dongjie Ji and Erhang Li and Fangyun Lin and Fucong Dai and Fuli Luo and Guangbo Hao and Guanting Chen and Guowei Li and H. Zhang and Han Bao and Hanwei Xu and Haocheng Wang and Haowei Zhang and Honghui Ding and Huajian Xin and Huazuo Gao and Hui Li and Hui Qu and J. L. Cai and Jian Liang and Jianzhong Guo and Jiaqi Ni and Jiashi Li and Jiawei Wang and Jin Chen and Jingchang Chen and Jingyang Yuan and Junjie Qiu and Junlong Li and Junxiao Song and Kai Dong and Kai Hu and Kaige Gao and Kang Guan and Kexin Huang and Kuai Yu and Lean Wang and Lecong Zhang and Lei Xu and Leyi Xia and Liang Zhao and Litong Wang and Liyue Zhang and Meng Li and Miaojun Wang and Mingchuan Zhang and Minghua Zhang and Minghui Tang and Mingming Li and Ning Tian and Panpan Huang and Peiyi Wang and Peng Zhang and Qiancheng Wang and Qihao Zhu and Qinyu Chen and Qiushi Du and R. J. Chen and R. L. Jin and Ruiqi Ge and Ruisong Zhang and Ruizhe Pan and Runji Wang and Runxin Xu and Ruoyu Zhang and Ruyi Chen and S. S. Li and Shanghao Lu and Shangyan Zhou and Shanhuang Chen and Shaoqing Wu and Shengfeng Ye and Shengfeng Ye and Shirong Ma and Shiyu Wang and Shuang Zhou and Shuiping Yu and Shunfeng Zhou and Shuting Pan and T. Wang and Tao Yun and Tian Pei and Tianyu Sun and W. L. Xiao and Wangding Zeng and Wanjia Zhao and Wei An and Wen Liu and Wenfeng Liang and Wenjun Gao and Wenqin Yu and Wentao Zhang and X. Q. Li and Xiangyue Jin and Xianzu Wang and Xiao Bi and Xiaodong Liu and Xiaohan Wang and Xiaojin Shen and Xiaokang Chen and Xiaokang Zhang and Xiaosha Chen and Xiaotao Nie and Xiaowen Sun and Xiaoxiang Wang and Xin Cheng and Xin Liu and Xin Xie and Xingchao Liu and Xingkai Yu and Xinnan Song and Xinxia Shan and Xinyi Zhou and Xinyu Yang and Xinyuan Li and Xuecheng Su and Xuheng Lin and Y. K. Li and Y. Q. Wang and Y. X. Wei and Y. X. Zhu and Yang Zhang and Yanhong Xu and Yanhong Xu and Yanping Huang and Yao Li and Yao Zhao and Yaofeng Sun and Yaohui Li and Yaohui Wang and Yi Yu and Yi Zheng and Yichao Zhang and Yifan Shi and Yiliang Xiong and Ying He and Ying Tang and Yishi Piao and Yisong Wang and Yixuan Tan and Yiyang Ma and Yiyuan Liu and Yongqiang Guo and Yu Wu and Yuan Ou and Yuchen Zhu and Yuduan Wang and Yue Gong and Yuheng Zou and Yujia He and Yukun Zha and Yunfan Xiong and Yunxian Ma and Yuting Yan and Yuxiang Luo and Yuxiang You and Yuxuan Liu and Yuyang Zhou and Z. F. Wu and Z. Z. Ren and Zehui Ren and Zhangli Sha and Zhe Fu and Zhean Xu and Zhen Huang and Zhen Zhang and Zhenda Xie and Zhengyan Zhang and Zhewen Hao and Zhibin Gou and Zhicheng Ma and Zhigang Yan and Zhihong Shao and Zhipeng Xu and Zhiyu Wu and Zhongyu Zhang and Zhuoshu Li and Zihui Gu and Zijia Zhu and Zijun Liu and Zilin Li and Ziwei Xie and Ziyang Song and Ziyi Gao and Zizheng Pan},
      year={2025},
      eprint={2412.19437},
      archivePrefix={arXiv},
      primaryClass={cs.CL},
      url={https://arxiv.org/abs/2412.19437}, 
}

@misc{deepseekai2024deepseekv2strongeconomicalefficient,
      title={DeepSeek-V2: A Strong, Economical, and Efficient Mixture-of-Experts Language Model}, 
      author={DeepSeek-AI and Aixin Liu and Bei Feng and Bin Wang and Bingxuan Wang and Bo Liu and Chenggang Zhao and Chengqi Dengr and Chong Ruan and Damai Dai and Daya Guo and Dejian Yang and Deli Chen and Dongjie Ji and Erhang Li and Fangyun Lin and Fuli Luo and Guangbo Hao and Guanting Chen and Guowei Li and H. Zhang and Hanwei Xu and Hao Yang and Haowei Zhang and Honghui Ding and Huajian Xin and Huazuo Gao and Hui Li and Hui Qu and J. L. Cai and Jian Liang and Jianzhong Guo and Jiaqi Ni and Jiashi Li and Jin Chen and Jingyang Yuan and Junjie Qiu and Junxiao Song and Kai Dong and Kaige Gao and Kang Guan and Lean Wang and Lecong Zhang and Lei Xu and Leyi Xia and Liang Zhao and Liyue Zhang and Meng Li and Miaojun Wang and Mingchuan Zhang and Minghua Zhang and Minghui Tang and Mingming Li and Ning Tian and Panpan Huang and Peiyi Wang and Peng Zhang and Qihao Zhu and Qinyu Chen and Qiushi Du and R. J. Chen and R. L. Jin and Ruiqi Ge and Ruizhe Pan and Runxin Xu and Ruyi Chen and S. S. Li and Shanghao Lu and Shangyan Zhou and Shanhuang Chen and Shaoqing Wu and Shengfeng Ye and Shirong Ma and Shiyu Wang and Shuang Zhou and Shuiping Yu and Shunfeng Zhou and Size Zheng and T. Wang and Tian Pei and Tian Yuan and Tianyu Sun and W. L. Xiao and Wangding Zeng and Wei An and Wen Liu and Wenfeng Liang and Wenjun Gao and Wentao Zhang and X. Q. Li and Xiangyue Jin and Xianzu Wang and Xiao Bi and Xiaodong Liu and Xiaohan Wang and Xiaojin Shen and Xiaokang Chen and Xiaosha Chen and Xiaotao Nie and Xiaowen Sun and Xiaoxiang Wang and Xin Liu and Xin Xie and Xingkai Yu and Xinnan Song and Xinyi Zhou and Xinyu Yang and Xuan Lu and Xuecheng Su and Y. Wu and Y. K. Li and Y. X. Wei and Y. X. Zhu and Yanhong Xu and Yanping Huang and Yao Li and Yao Zhao and Yaofeng Sun and Yaohui Li and Yaohui Wang and Yi Zheng and Yichao Zhang and Yiliang Xiong and Yilong Zhao and Ying He and Ying Tang and Yishi Piao and Yixin Dong and Yixuan Tan and Yiyuan Liu and Yongji Wang and Yongqiang Guo and Yuchen Zhu and Yuduan Wang and Yuheng Zou and Yukun Zha and Yunxian Ma and Yuting Yan and Yuxiang You and Yuxuan Liu and Z. Z. Ren and Zehui Ren and Zhangli Sha and Zhe Fu and Zhen Huang and Zhen Zhang and Zhenda Xie and Zhewen Hao and Zhihong Shao and Zhiniu Wen and Zhipeng Xu and Zhongyu Zhang and Zhuoshu Li and Zihan Wang and Zihui Gu and Zilin Li and Ziwei Xie},
      year={2024},
      eprint={2405.04434},
      archivePrefix={arXiv},
      primaryClass={cs.CL},
      url={https://arxiv.org/abs/2405.04434}, 
}

@misc{yang2025qwen3technicalreport,
      title={Qwen3 Technical Report}, 
      author={An Yang and Anfeng Li and Baosong Yang and Beichen Zhang and Binyuan Hui and Bo Zheng and Bowen Yu and Chang Gao and Chengen Huang and Chenxu Lv and Chujie Zheng and Dayiheng Liu and Fan Zhou and Fei Huang and Feng Hu and Hao Ge and Haoran Wei and Huan Lin and Jialong Tang and Jian Yang and Jianhong Tu and Jianwei Zhang and Jianxin Yang and Jiaxi Yang and Jing Zhou and Jingren Zhou and Junyang Lin and Kai Dang and Keqin Bao and Kexin Yang and Le Yu and Lianghao Deng and Mei Li and Mingfeng Xue and Mingze Li and Pei Zhang and Peng Wang and Qin Zhu and Rui Men and Ruize Gao and Shixuan Liu and Shuang Luo and Tianhao Li and Tianyi Tang and Wenbiao Yin and Xingzhang Ren and Xinyu Wang and Xinyu Zhang and Xuancheng Ren and Yang Fan and Yang Su and Yichang Zhang and Yinger Zhang and Yu Wan and Yuqiong Liu and Zekun Wang and Zeyu Cui and Zhenru Zhang and Zhipeng Zhou and Zihan Qiu},
      year={2025},
      eprint={2505.09388},
      archivePrefix={arXiv},
      primaryClass={cs.CL},
      url={https://arxiv.org/abs/2505.09388}, 
}

@misc{muennighoff2025olmoeopenmixtureofexpertslanguage,
      title={OLMoE: Open Mixture-of-Experts Language Models}, 
      author={Niklas Muennighoff and Luca Soldaini and Dirk Groeneveld and Kyle Lo and Jacob Morrison and Sewon Min and Weijia Shi and Pete Walsh and Oyvind Tafjord and Nathan Lambert and Yuling Gu and Shane Arora and Akshita Bhagia and Dustin Schwenk and David Wadden and Alexander Wettig and Binyuan Hui and Tim Dettmers and Douwe Kiela and Ali Farhadi and Noah A. Smith and Pang Wei Koh and Amanpreet Singh and Hannaneh Hajishirzi},
      year={2025},
      eprint={2409.02060},
      archivePrefix={arXiv},
      primaryClass={cs.CL},
      url={https://arxiv.org/abs/2409.02060}, 
}

@article{shazeer2017outrageously,
  title={Outrageously large neural networks: The sparsely-gated mixture-of-experts layer},
  author={Shazeer, Noam and Mirhoseini, Azalia and Maziarz, Krzysztof and Davis, Andy and Le, Quoc and Hinton, Geoffrey and Dean, Jeff},
  journal={arXiv preprint arXiv:1701.06538},
  year={2017}
}

@article{lepikhin2020gshard,
  title={Gshard: Scaling giant models with conditional computation and automatic sharding},
  author={Lepikhin, Dmitry and Lee, HyoukJoong and Xu, Yuanzhong and Chen, Dehao and Firat, Orhan and Huang, Yanping and Krikun, Maxim and Shazeer, Noam and Chen, Zhifeng},
  journal={arXiv preprint arXiv:2006.16668},
  year={2020}
}

@article{fedus2022switch,
  title={Switch transformers: Scaling to trillion parameter models with simple and efficient sparsity},
  author={Fedus, William and Zoph, Barret and Shazeer, Noam},
  journal={Journal of Machine Learning Research},
  volume={23},
  number={120},
  pages={1--39},
  year={2022}
}

@misc{qian2025fino1transferabilityreasoningenhancedllms,
      title={Fino1: On the Transferability of Reasoning-Enhanced LLMs and Reinforcement Learning to Finance}, 
      author={Lingfei Qian and Weipeng Zhou and Yan Wang and Xueqing Peng and Han Yi and Yilun Zhao and Jimin Huang and Qianqian Xie and Jian-yun Nie},
      year={2025},
      eprint={2502.08127},
      archivePrefix={arXiv},
      primaryClass={cs.CL},
      url={https://arxiv.org/abs/2502.08127}, 
}

@inproceedings{wang2025rkefino1,
  title={RKEFino1: A Regulation Knowledge-Enhanced Large Language Model},
  author={Wang, Yan and He, Yueru and Xiang, Ruoyu and Zhao, Jeff},
  booktitle={2025 IEEE 11th International Conference on Intelligent Data and Security (IDS)},
  pages={49--51},
  year={2025},
  organization={IEEE}
}

@misc{wang2025fintaggingbenchmarkingllmsextracting,
      title={FinTagging: Benchmarking LLMs for Extracting and Structuring Financial Information}, 
      author={Yan Wang and Yang Ren and Lingfei Qian and Xueqing Peng and Keyi Wang and Yi Han and Dongji Feng and Fengran Mo and Shengyuan Lin and Qinchuan Zhang and Kaiwen He and Chenri Luo and Jianxing Chen and Junwei Wu and Jimin Huang and Guojun Xiong and Xiao-Yang Liu and Qianqian Xie and Jian-Yun Nie},
      year={2025},
      eprint={2505.20650},
      archivePrefix={arXiv},
      primaryClass={cs.CL},
      url={https://arxiv.org/abs/2505.20650}, 
}

@misc{wang2025finauditingfinancialtaxonomystructuredmultidocument,
      title={FinAuditing: A Financial Taxonomy-Structured Multi-Document Benchmark for Evaluating LLMs}, 
      author={Yan Wang and Keyi Wang and Shanshan Yang and Jaisal Patel and Jeff Zhao and Fengran Mo and Xueqing Peng and Lingfei Qian and Jimin Huang and Guojun Xiong and Xiao-Yang Liu and Jian-Yun Nie},
      year={2025},
      eprint={2510.08886},
      archivePrefix={arXiv},
      primaryClass={cs.CL},
      url={https://arxiv.org/abs/2510.08886}, 
}

@article{raffel2020exploring,
  title={Exploring the limits of transfer learning with a unified text-to-text transformer},
  author={Raffel, Colin and Shazeer, Noam and Roberts, Adam and Lee, Katherine and Narang, Sharan and Matena, Michael and Zhou, Yanqi and Li, Wei and Liu, Peter J},
  journal={Journal of machine learning research},
  volume={21},
  number={140},
  pages={1--67},
  year={2020}
}

@article{hendrycks2020measuring,
  title={Measuring massive multitask language understanding},
  author={Hendrycks, Dan and Burns, Collin and Basart, Steven and Zou, Andy and Mazeika, Mantas and Song, Dawn and Steinhardt, Jacob},
  journal={arXiv preprint arXiv:2009.03300},
  year={2020}
}

\appendix

\section{Computational Budget and Infrastructure}
\label{sec:com}
All experiments are implemented in Python using PyTorch and the HuggingFace Transformers library, and are conducted in inference-only mode. We evaluate each model on the full MMLU benchmark using two NVIDIA A100 (80 GB) GPUs. OLMoE-1B-7B, DeepSeek-V2-Lite, and Qwen3-30B-A3B contain approximately 7B, 16B, and 30B parameters, respectively. The total computational cost of the routing analyses is about 40 GPU-hours, including forward passes and the collection of routing statistics.

\section{C4 Dataset}
\label{C4_data}
To complement the domain-level analysis on MMLU, we additionally construct a structured evaluation subset from C4 (Colossal Clean Crawled Corpus)~\cite{raffel2020exploring}for multilingual routing analysis. While MMLU provides subject annotations that can be aggregated into higher-level semantic domains, C4 does not contain analogous task labels. We therefore organize the selected C4 subsets by language-oriented groups, yielding four higher-level domains: \textbf{European} (de, es, fr, pt), \textbf{Asian} (zh, hi, ja), \textbf{Middle Eastern} (ar), and \textbf{Other} (en, ru).

This C4-based evaluation set contains \textbf{1,000 samples} in total, drawn from \textbf{10 language-specific sources}, with \textbf{100 samples per language}. The resulting domain distribution is moderately imbalanced by construction, including 400 European samples, 300 Asian samples, 100 Middle Eastern samples, and 200 Other samples. This aggregation mirrors the design principle of Table~\ref{tab:mmlu_domains}: rather than analyzing routing behavior at the level of individual sources, we group related sources into broader domains in order to examine domain-conditioned routing patterns at a more stable level of abstraction.

We also observe non-trivial variation in input length across these groups. The average sequence length ranges from 348.8 in the Other group to 467.9 in the Middle Eastern group, while all groups contain samples reaching the truncation limit of 512 tokens. These differences make the C4 subset a useful complement to MMLU, allowing us to test whether Standing Committee behavior persists beyond supervised benchmark subjects and extends to a broader multilingual web corpus.

\begin{table}[t]
\centering
\small
\resizebox{\linewidth}{!}{
\begin{tabular}{l l c c}
\toprule
Domain Group & Representative Languages & \# Samples & Avg. Seq. Len. \\
\midrule
European & German, Spanish, French, Portuguese & 400 & 402.7 \\
Asian & Chinese, Hindi, Japanese & 300 & 402.9 \\
Middle Eastern & Arabic & 100 & 467.9 \\
Other & English, Russian & 200 & 348.8 \\
\bottomrule
\end{tabular}
}
\caption{Aggregation of the C4-based evaluation subset into four language-oriented domain groups. The subset contains 1,000 samples in total, drawn from 10 language-specific sources with 100 samples per language.}
\label{tab:c4_domains}
\end{table}

\section{Illustrative Expert-Level Statistics for Committee Selection}
\label{illustrative_exp}

To make the committee selection process more transparent, we provide an illustrative expert-level view for representative DeepSeek-V2-Lite layers, as shown in Table~\ref{tab:committee_selection_illustration}. For each activated expert, we report its domain coverage, mean ECI, standard deviation of ECI, and an auxiliary stability score defined as the product of domain coverage and mean ECI. This score is introduced only for interpretive presentation, to provide a more tangible view of how highly covered and high-contribution experts differ from weaker candidates. It is \textbf{not} part of the formal committee selection criterion. The formal method remains exactly as defined in Section~\ref{framework}: experts must first satisfy the cross-domain agreement threshold $\gamma > 0.8$, and committee members are then determined based on the Pareto-optimal set of rank statistics.

The table also clarifies an important point: domain coverage alone is not sufficient for committee membership. Some experts are activated in all or nearly all domains, but their mean ECI remains substantially lower than that of the final committee members. As a result, they do not belong to the committee, even though they are broadly active. This provides a more direct illustration that Standing Committees are not defined by activation frequency alone, but by stable and substantial contribution across domains.

\begin{table*}[t]
\centering
\small
\caption{Illustrative expert-level statistics for committee selection in representative DeepSeek-V2-Lite layers. Domain Coverage is measured out of the nine aggregated MMLU domains. The Stability Score is an auxiliary presentation-only quantity defined as Domain Coverage $\times$ Mean ECI, and is included only to improve interpretability; it is not part of the formal committee selection rule.}
\label{tab:committee_selection_illustration}

\resizebox{\textwidth}{!}{
\begin{tabular}{c c c c c c c c}
\toprule
\textbf{Layer} & \textbf{Phase} & \textbf{Activated Expert} & \textbf{Domain Coverage} & \textbf{Mean ECI} & \textbf{Std ECI} & \textbf{Stability Score} & \textbf{Committee} \\
\midrule
3  & Shallow & 7  & 9 & 0.8832 & 0.03 & 7.9488 & Y \\
3  & Shallow & 13 & 9 & 0.8253 & 0.02 & 7.4277 & Y \\
3  & Shallow & 22 & 9 & 0.8043 & 0.03 & 7.2387 & Y \\
3  & Shallow & 42 & 9 & 0.7825 & 0.02 & 7.0425 & Y \\
3  & Shallow & 1  & 9 & 0.5310 & 0.01 & 4.7790 & N \\
3  & Shallow & 9  & 9 & 0.4928 & 0.02 & 4.4352 & N \\
\midrule
11 & Middle  & 28 & 9 & 0.9259 & 0.03 & 8.3331 & Y \\
11 & Middle  & 43 & 9 & 0.9215 & 0.01 & 8.2935 & Y \\
11 & Middle  & 54 & 9 & 0.9005 & 0.02 & 8.1045 & Y \\
11 & Middle  & 47 & 7 & 0.8642 & 0.05 & 6.0494 & N \\
11 & Middle  & 16 & 6 & 0.6329 & 0.09 & 3.7974 & N \\
11 & Middle  & 62 & 5 & 0.5903 & 0.04 & 2.9515 & N \\
11 & Middle  & 35 & 3 & 0.3178 & 0.04 & 0.9534 & N \\
11 & Middle  & 8  & 3 & 0.2721 & 0.02 & 0.8163 & N \\
11 & Middle  & 1  & 1 & 0.1919 & 0.00 & 0.1919 & N \\
\midrule
19 & Deep    & 4  & 9 & 0.9386 & 0.03 & 8.4474 & Y \\
19 & Deep    & 14 & 9 & 0.9298 & 0.03 & 8.3682 & Y \\
19 & Deep    & 17 & 9 & 0.9055 & 0.04 & 8.1495 & Y \\
19 & Deep    & 61 & 9 & 0.9013 & 0.02 & 8.1117 & Y \\
19 & Deep    & 53 & 9 & 0.7524 & 0.02 & 6.7716 & N \\
19 & Deep    & 12 & 7 & 0.3111 & 0.04 & 2.1777 & N \\
19 & Deep    & 32 & 3 & 0.1767 & 0.04 & 0.5301 & N \\
\bottomrule
\end{tabular}
}
\end{table*}

\section{Standing Committees across all layers}
\label{sec:sc_all_layer}

\subsection{MMLU Dataset Results}
\label{mmlu_re}
\paragraph{Full-layer analysis of standing committees.}
To move beyond representative snapshots, we perform a layer-by-layer audit of all identified standing committees in OLMoE, DeepSeek-V2-Lite, and Qwen3-30B-A3B, as shown in Table~\ref{tab:olmoe_committees},~\ref{tab:deepseek_committees}, and~\ref{tab:qwen_committees}. The resulting tables reveal a consistent organizational pattern that is obscured when only a few layers are examined. Across models, standing committees emerge early, consolidate in the middle layers, and persist into the deepest layers, while their composition changes only gradually.

\paragraph{Small, persistent coalitions.}
Despite large expert pools, the size of each committee remains compact: typically $|\mathcal{C}| \in [1,4]$ for OLMoE and DeepSeek-V2-Lite, and occasionally up to five members in Qwen. Increasing expert capacity does not diversify routing. Instead, optimization repeatedly converges onto a small coalition of experts that are selected across domains and prompts. This suggests that sparse routing does not primarily allocate experts by domain; rather, it reinforces a stable computational core.

\paragraph{Early centralization.}
A striking finding is that centralization appears already in shallow layers. Even in the first few layers, standing committees capture a non-trivial proportion of routing mass (often $20$--$40\%$). This indicates that MoE models commit to shared processing pathways almost immediately, likely encoding high-frequency patterns such as token normalization, shallow syntactic cues, and generic lexical regularities. Contrary to the intuition that specialization gradually emerges with depth, the router begins consolidating computation from the outset.

\paragraph{Middle-layer consolidation.}
The middle layers display the clearest standing-committee behavior. Committees often grow slightly larger while their rank variance decreases, and their cumulative contribution increases sharply (frequently exceeding $50$--$65\%$ in DeepSeek-V2-Lite and Qwen). These layers appear to implement domain-agnostic abstractions, reasoning templates, discourse structure, and general semantic scaffolding, that are shared across inputs. The router does not allocate different domains to distinct experts; instead, it repeatedly routes through the same committee.

\paragraph{Deep-layer bottlenecks.}
In deep layers, both DeepSeek-V2-Lite and Qwen exhibit strong bottleneck effects: a small committee controls $50$--$70\%$ of routing mass, often with high influence density. Rather than distributing final computation across diverse experts, the network funnels decision-making through a narrow coalition. This pattern challenges the traditional ``divide-and-conquer'' view of MoE systems, suggesting that final reasoning is centralized rather than decomposed.

\paragraph{Architectural variability, consistent behavior.}
Although OLMoE shows weaker committees than Qwen and DeepSeek-V2-Lite, the qualitative trend is remarkably stable across architectures. Even with different routing designs and training recipes, all three models converge toward small, domain-invariant committees that repeatedly dominate computation. Taken together, these findings indicate that standing committees are not an artifact of any particular implementation. Instead, they appear to be an emergent consequence of sparse routing optimization, reflecting a strong inductive bias toward centralization in modern MoE language models.

\begin{table}[t]
\centering
\small
\caption{Comprehensive audit of Standing Committees for OLMoE.}
\label{tab:olmoe_committees}

\resizebox{\linewidth}{!}{
\begin{tabular}{lcccccc}
\toprule
\textbf{Layer ($\ell$)} & \textbf{Committee Members ($\mathcal{C}$)} & \textbf{$|\mathcal{C}|$} & \textbf{Avg. $\mu$ ↓} & \textbf{Avg. $\sigma^2$ ↓} & \textbf{ECI Cov.} & \textbf{Ratio ($\times$)} \\
\midrule
1 & {49} & 1 & 1.00 & 0.00 & 21.5\% & 17.25\\
2 & {58, 63, 30} & 3 & 3.41 & 2.15 & 43.9\% & 15.94\\
3 & {60, 14} & 2 & 3.17 & 1.17 & 30.9\% & 13.84\\
4 & {9, 56} & 2 & 1.67 & 0.33 & 34.4\% & 16.24\\
5 & {27} & 1 & 3.11 & 0.99 & 14.5\% & 10.68\\
6 & {53} & 1 & 1.78 & 1.06 & 16.3\% & 12.27\\
7 & {1} & 1 & 1.89 & 0.54 & 16.4\% & 12.38\\
8 & {13, 45} & 2 & 3.28 & 0.49 & 29.7\% & 13.10\\
9 & {27} & 1 & 1.33 & 0.22 & 17.3\% & 13.15\\
10 & {8, 12} & 2 & 3.78 & 0.49 & 28.3\% & 12.27\\
11 & {30, 6} & 2 & 4.28 & 1.05 & 26.8\% & 11.35\\
12 & {33} & 1 & 2.00 & 2.00 & 17.1\% & 12.96\\
13 & {30, 55} & 2 & 2.50 & 1.95 & 30.6\% & 13.69\\
14 & {46, 4} & 2 & 2.22 & 0.73 & 31.3\% & 14.14\\
15 & {1} & 1 & 1.78 & 0.62 & 16.9\% & 12.78\\
16 & {60, 52, 17} & 3 & 3.19 & 1.52 & 44.0\% & 15.99\\
\bottomrule
\end{tabular}
}
\end{table}

\begin{table}[t]
\centering
\small
\caption{Comprehensive audit of Standing Committees for DeepSeek-V2-Lite.}
\label{tab:deepseek_committees}

\resizebox{\linewidth}{!}{
\begin{tabular}{lcccccc}
\toprule
\textbf{Layer ($\ell$)} & \textbf{Committee Members ($\mathcal{C}$)} & \textbf{$|\mathcal{C}|$} & \textbf{Avg. $\mu$ ↓} & \textbf{Avg. $\sigma^2$ ↓} & \textbf{ECI Cov.} & \textbf{Ratio ($\times$)} \\
\midrule
1 & {25, 57} & 2 & 3.33 & 0.58 & 34.8\% & 16.58\\
2 & {19, 51, 46} & 3 & 3.00 & 1.98 & 52.0\% & 22.02\\
3 & {42, 7, 13, 22} & 4 & 3.36 & 1.81 & 66.3\% & 29.46\\
4 & {25, 59, 13} & 3 & 2.44 & 1.72 & 57.3\% & 27.25\\
5 & {38} & 1 & 1.22 & 0.17 & 22.0\% & 17.74\\
6 & {35, 46} & 2 & 3.83 & 1.00 & 31.5\% & 14.25\\
7 & {50, 17} & 2 & 3.33 & 0.33 & 34.5\% & 16.36\\
8 & {45, 46} & 2 & 2.33 & 1.51 & 37.9\% & 18.88\\
9 & {38, 46, 41} & 3 & 2.70 & 1.43 & 56.7\% & 26.66\\
10 & {60} & 1 & 1.22 & 0.40 & 23.6\% & 19.41\\
11 & {54, 43, 28} & 3 & 2.59 & 0.35 & 60.7\% & 31.36\\
12 & {30} & 1 & 1.33 & 0.44 & 21.9\% & 17.63\\
13 & {29, 6} & 2 & 2.39 & 0.46 & 40.6\% & 21.17\\
14 & {5, 28, 33} & 3 & 2.37 & 1.43 & 58.7\% & 28.93\\
15 & {8, 6, 0} & 3 & 3.04 & 1.22 & 56.1\% & 26.00\\
16 & {24} & 1 & 1.33 & 0.22 & 20.4\% & 16.18\\
17 & {40, 25, 31} & 3 & 3.89 & 0.91 & 44.3\% & 16.18\\
18 & {51, 46, 53, 0} & 4 & 3.67 & 0.65 & 65.5\% & 28.43\\
19 & {61, 14, 47, 4} & 4 & 3.11 & 0.76 & 70.5\% & 35.78\\
20 & {44, 7} & 2 & 3.33 & 0.47 & 34.3\% & 16.17\\
21 & {48} & 1 & 1.67 & 0.67 & 22.1\% & 17.85\\
22 & {40, 21} & 2 & 3.33 & 1.32 & 35.4\% & 17.00\\
23 & {23, 6, 38} & 3 & 3.30 & 0.58 & 52.1\% & 22.14\\
24 & {60, 61} & 2 & 2.00 & 1.06 & 42.7\% & 23.08\\
25 & {44, 1} & 2 & 1.83 & 0.32 & 46.3\% & 26.74\\
26 & {36, 56} & 2 & 3.17 & 0.56 & 34.6\% & 16.41\\
\bottomrule
\end{tabular}
}
\end{table}

\begin{table}[t]
\centering
\small
\caption{Comprehensive audit of Standing Committees for Qwen3-30B-A3B.}
\label{tab:qwen_committees}

\resizebox{\linewidth}{!}{
\begin{tabular}{lcccccc}
\toprule
\textbf{Layer ($\ell$)} & \textbf{Committee Members ($\mathcal{C}$)} & \textbf{$|\mathcal{C}|$} & \textbf{Avg. $\mu$ ↓} & \textbf{Avg. $\sigma^2$ ↓} & \textbf{ECI Cov.} & \textbf{Ratio ($\times$)} \\
\midrule
1 & {114} & 1 & 2.56 & 0.91 & 14.9\% & 22.26\\
2 & {119} & 1 & 2.78 & 2.17 & 14.4\% & 21.40\\
3 & {40, 93, 80, 38} & 4 & 3.61 & 3.44 & 54.0\% & 36.43\\
4 & {34, 120, 84} & 3 & 3.63 & 3.78 & 40.2\% & 28.02\\
5 & {104, 63, 81} & 3 & 3.22 & 2.49 & 43.6\% & 32.16\\
6 & {68, 1, 37, 66} & 4 & 4.25 & 2.95 & 52.6\% & 34.39\\
7 & {56, 71, 78} & 3 & 3.37 & 4.05 & 41.9\% & 30.03\\
8 & {112, 101} & 2 & 3.67 & 0.77 & 27.1\% & 23.36\\
9 & {26} & 1 & 1.33 & 0.44 & 17.6\% & 27.20\\
10 & {114, 84} & 2 & 4.39 & 0.68 & 26.7\% & 22.96\\
11 & {98, 53, 112} & 3 & 4.19 & 1.28 & 40.1\% & 27.89\\
12 & {60, 125, 7} & 3 & 3.15 & 1.98 & 43.3\% & 31.83\\
13 & {65, 78, 56} & 3 & 3.59 & 1.67 & 43.2\% & 31.64\\
14 & {122} & 1 & 1.78 & 0.62 & 17.1\% & 26.25\\
15 & {20, 90, 17} & 3 & 4.04 & 3.21 & 38.9\% & 26.47\\
16 & {116} & 1 & 3.33 & 1.33 & 20.6\% & 24.64\\
17 & {70, 34, 83} & 3 & 3.48 & 1.74 & 39.4\% & 27.05\\
18 & {61, 31, 77} & 3 & 3.59 & 2.44 & 46.7\% & 32.06\\
19 & {6, 105, 21} & 3 & 3.04 & 1.54 & 41.2\% & 30.50\\
20 & {121, 92, 17} & 3 & 3.89 & 1.83 & 42.1\% & 28.01\\
21 & {93, 106, 63} & 3 & 3.59 & 2.65 & 41.9\% & 27.91\\
22 & {99, 106} & 2 & 2.39 & 1.21 & 27.7\% & 25.04\\
23 & {37, 44, 65} & 3 & 3.78 & 2.71 & 39.2\% & 27.69\\
24 & {121, 86, 36} & 3 & 3.63 & 2.46 & 42.4\% & 29.03\\
25 & {52, 35, 24} & 3 & 3.22 & 2.27 & 39.7\% & 27.28\\
26 & {113, 109, 33} & 3 & 3.81 & 2.13 & 44.5\% & 31.03\\
27 & {31, 123} & 2 & 2.78 & 1.38 & 28.4\% & 24.77\\
28 & {78, 73, 62} & 3 & 3.78 & 2.61 & 41.3\% & 28.77\\
29 & {17, 47, 49} & 3 & 3.74 & 1.87 & 39.4\% & 27.54\\
30 & {116, 65, 40} & 3 & 3.81 & 2.24 & 41.7\% & 28.96\\
31 & {57, 24, 92} & 3 & 3.56 & 2.32 & 41.1\% & 29.18\\
32 & {83, 9, 32} & 3 & 3.63 & 2.49 & 43.6\% & 30.14\\
33 & {57, 121, 16, 26, 116} & 5 & 3.82 & 2.16 & 67.0\% & 49.88\\
34 & {9, 96, 110, 64} & 4 & 3.86 & 2.83 & 54.0\% & 36.40\\
35 & {105, 56} & 2 & 3.11 & 1.53 & 29.0\% & 25.75\\
36 & {63, 23} & 2 & 3.00 & 2.99 & 28.7\% & 25.35\\
37 & {96} & 1 & 2.33 & 0.89 & 14.7\% & 21.84\\
38 & {0} & 1 & 1.11 & 0.10 & 17.2\% & 26.40\\
39 & {20, 48, 86} & 3 & 4.15 & 1.96 & 40.0\% & 27.80\\
40 & {85, 49} & 2 & 2.33 & 1.32 & 30.9\% & 28.16\\
41 & {81, 17, 87} & 3 & 4.30 & 1.07 & 38.9\% & 26.51\\
42 & {55, 21} & 2 & 2.17 & 2.14 & 30.6\% & 27.80\\
43 & {31, 6, 56} & 3 & 3.74 & 2.46 & 41.8\% & 29.91\\
44 & {71, 31} & 2 & 3.33 & 0.84 & 29.6\% & 26.53\\
45 & {90} & 1 & 2.11 & 0.54 & 16.1\% & 24.40\\
46 & {107, 94, 101} & 3 & 3.15 & 1.59 & 50.9\% & 43.26\\
47 & {38, 34} & 2 & 2.39 & 1.21 & 36.8\% & 36.65\\
48 & {101} & 1 & 2.89 & 0.99 & 14.2\% & 20.99\\
\bottomrule
\end{tabular}
}
\end{table}

\subsection{C4 Dataset Results}
\label{c4_re}
\paragraph{Matched layer-wise analysis on free-form web text.}
Based on the C4-based domain construction described in Appendix~\ref{C4_data}, we further conduct a matched supplementary analysis to test whether the Standing Committee phenomenon depends on the multiple-choice structure of MMLU. Specifically, we apply the same committee-identification procedure to the C4-based evaluation subset using the identical DeepSeek-V2-Lite configuration as in the main experiments. We keep the routing sparsity fixed at $k=6$ and exclude shared experts from committee construction, so that the analysis isolates the effect of input distribution while holding the routing protocol constant. Due to space constraints, we report representative shallow, middle, and deep layers rather than the full layer-wise C4 results.

\paragraph{Compact committees persist beyond MMLU.}
The resulting pattern remains consistent with our main findings on MMLU. Across the reported representative layers, the identified committees remain compact, typically containing only $1$--$4$ experts. This indicates that the emergence of a small dominant expert coalition is not specific to subject-structured benchmark inputs, but also appears under free-form web-text inputs drawn from C4.

\paragraph{Substantial routing coverage under free-form inputs.}
Although the committees are small, they still capture a substantial fraction of routing mass. In the reported layers, committee coverage ranges from $12.43\%$ to $66.32\%$, with several middle and deep layers exceeding $30\%$ and the strongest layers surpassing $50\%$. In particular, Layer~0 reaches $66.32\%$ coverage, while Layer~24 reaches $62.81\%$, showing that centralized routing persists even when the input format differs substantially from MMLU.

\paragraph{High influence density and persistent centralization.}
We also observe consistently high influence density across layers. The ratio between the average contribution of committee members and that of non-members ranges from $13.48\times$ to $45.22\times$. This means that even when more experts remain nominally available, the effective routing mass is still disproportionately absorbed by a very small committee. The strongest concentration appears in deeper layers, where the ratio exceeds $40\times$, further supporting the view that sparse routing naturally converges toward a centralized computational core.

\paragraph{Implications for robustness across datasets.}
Taken together, these results suggest that the Standing Committee phenomenon is not an artifact of MMLU's multiple-choice format. Instead, it persists under a qualitatively different input distribution based on free-form web text. This cross-dataset consistency strengthens our central claim that Standing Committees reflect a general structural tendency of sparse MoE routing, rather than a benchmark-specific effect.

\begin{table}[t]
\centering
\small
\caption{Representative Standing Committees on the C4-based evaluation subset using DeepSeek-V2-Lite under the same routing configuration as in the main experiments. We report representative shallow, middle, and deep layers rather than the full layer-wise results. The committees remain compact while capturing substantial routing mass, showing that the Standing Committee phenomenon persists beyond MMLU's multiple-choice format.}
\label{tab:c4_committees}

\resizebox{\linewidth}{!}{
\begin{tabular}{lcccccc}
\toprule
\textbf{Layer ($\ell$)} & \textbf{Committee Members ($\mathcal{C}$)} & \textbf{$|\mathcal{C}|$} & \textbf{Avg. $\mu$ ↓} & \textbf{Avg. $\sigma^2$ ↓} & \textbf{ECI Cov.} & \textbf{Ratio ($\times$)} \\
\midrule
1  & {4, 49, 50, 25} & 4 & 2.41 & 0.55 & 66.32\% & 13.78 \\
2  & {0}             & 1 & 1.25 & 0.86 & 12.43\% & 18.94 \\
8  & {32, 6}         & 2 & 2.75 & 0.62 & 35.59\% & 27.13 \\
9  & {45, 41}        & 2 & 2.50 & 1.19 & 33.78\% & 25.81 \\
10  & {60}            & 1 & 2.25 & 1.66 & 17.62\% & 13.48 \\
11 & {16}            & 1 & 2.00 & 0.00 & 16.74\% & 22.67 \\
19 & {22, 6, 25, 4}  & 4 & 3.31 & 1.64 & 52.92\% & 27.36 \\
20 & {6, 44}         & 2 & 4.62 & 1.34 & 21.50\% & 18.49 \\
21 & {56}            & 1 & 2.50 & 2.23 & 22.00\% & 18.59 \\
22 & {11, 2}         & 2 & 3.12 & 5.22 & 30.56\% & 43.64 \\
23 & {22, 38}        & 2 & 3.88 & 4.72 & 27.13\% & 31.54 \\
24 & {48, 61}        & 2 & 2.12 & 2.72 & 38.05\% & 32.09 \\
25 & {39, 25, 5}     & 3 & 3.75 & 5.06 & 62.81\% & 45.22 \\
26 & {56}            & 1 & 1.25 & 0.19 & 19.92\% & 15.67 \\
\bottomrule
\end{tabular}
}
\end{table}

\section{Contribution Concentration Analysis}
\label{Con_con_ana}
\subsection{Lorenz Curve for OLMoE Model}

\begin{figure*}[t]
\centering

\begin{subcaptionblock}{0.19\linewidth}
\includegraphics[width=\linewidth]{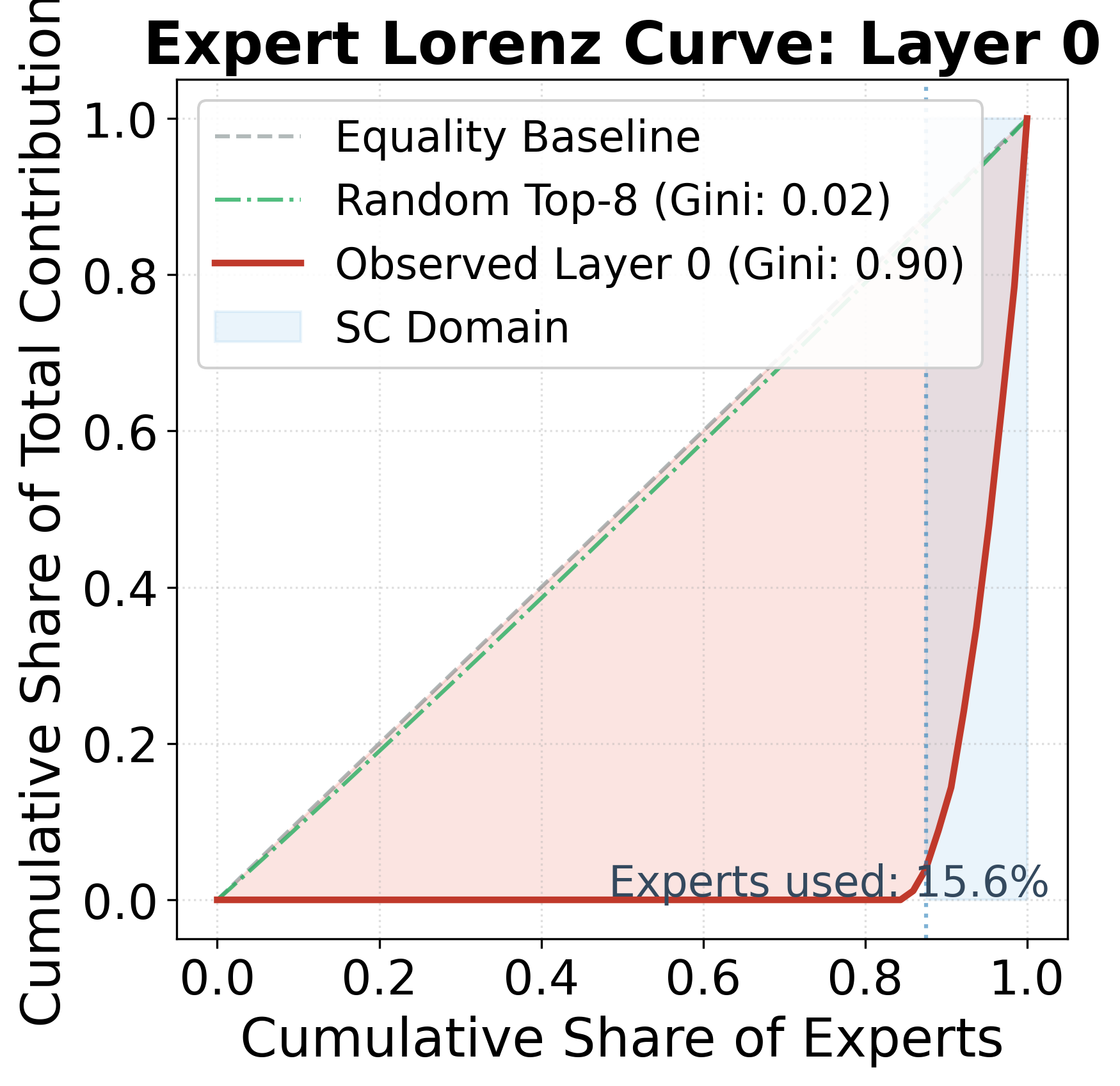}
\caption{Layer 0}
\end{subcaptionblock}
\begin{subcaptionblock}{0.19\linewidth}
\includegraphics[width=\linewidth]{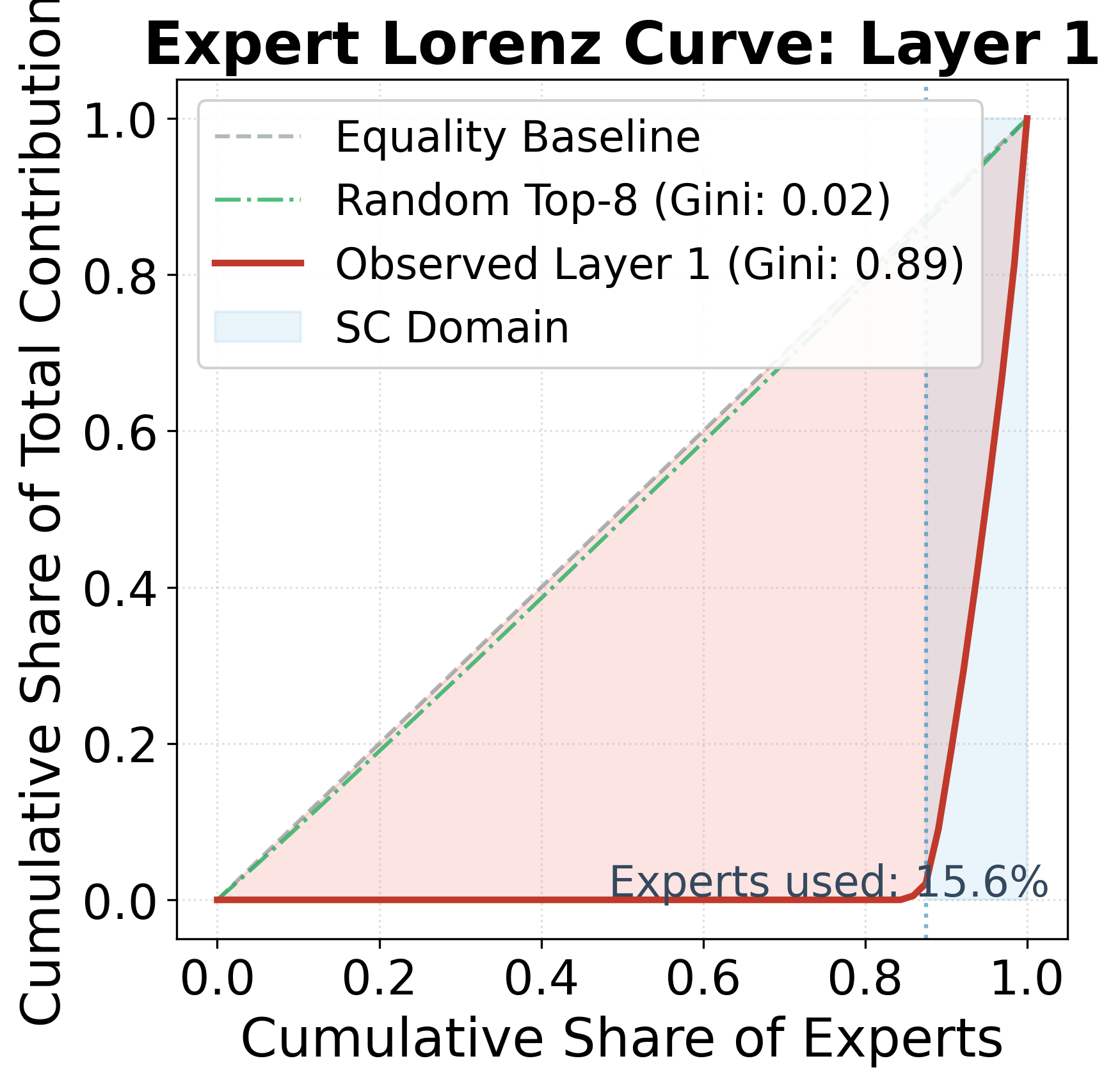}
\caption{Layer 1}
\end{subcaptionblock}
\begin{subcaptionblock}{0.19\linewidth}
\includegraphics[width=\linewidth]{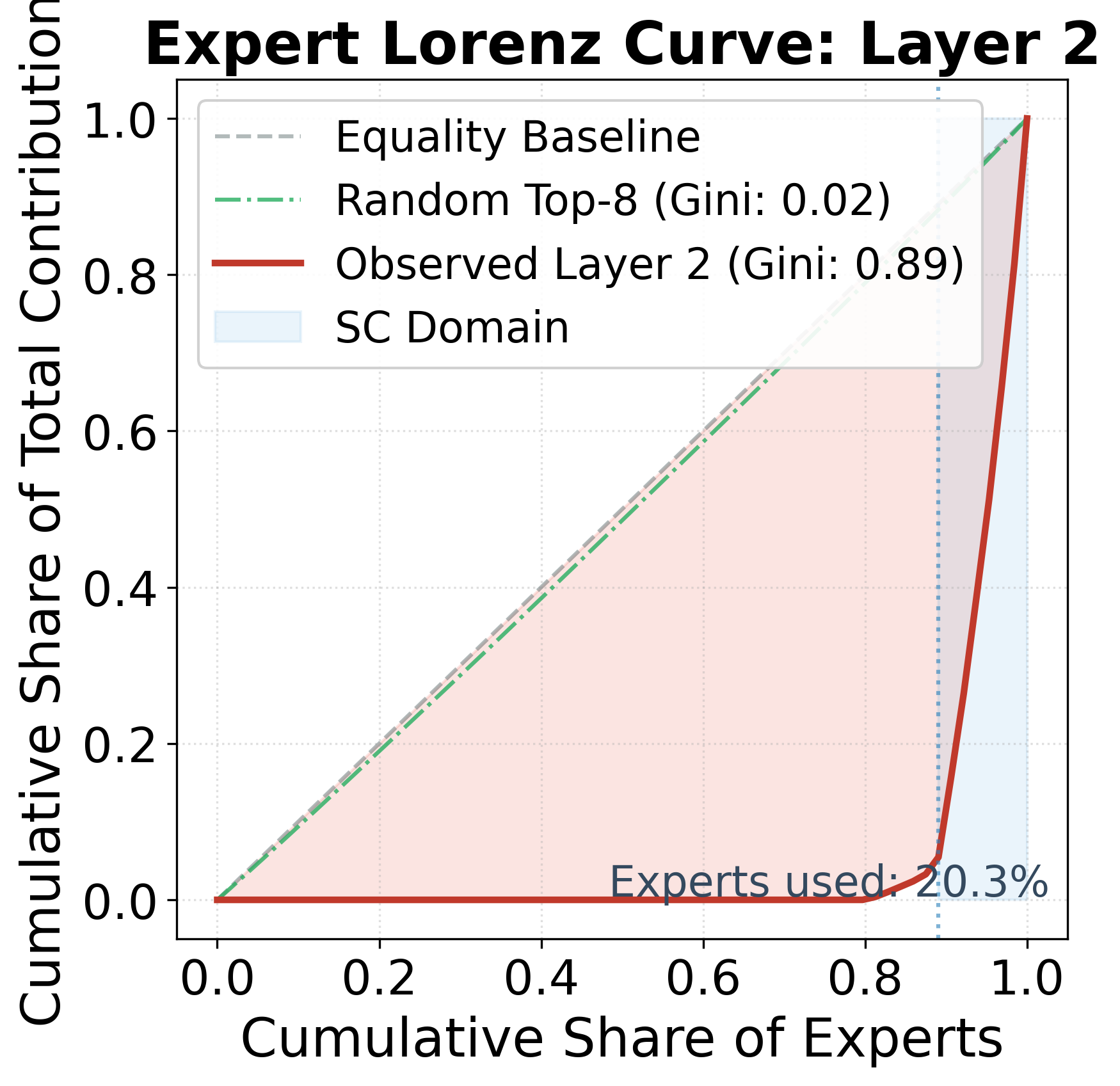}
\caption{Layer 2}
\end{subcaptionblock}

\medskip

\begin{subcaptionblock}{0.19\linewidth}
\includegraphics[width=\linewidth]{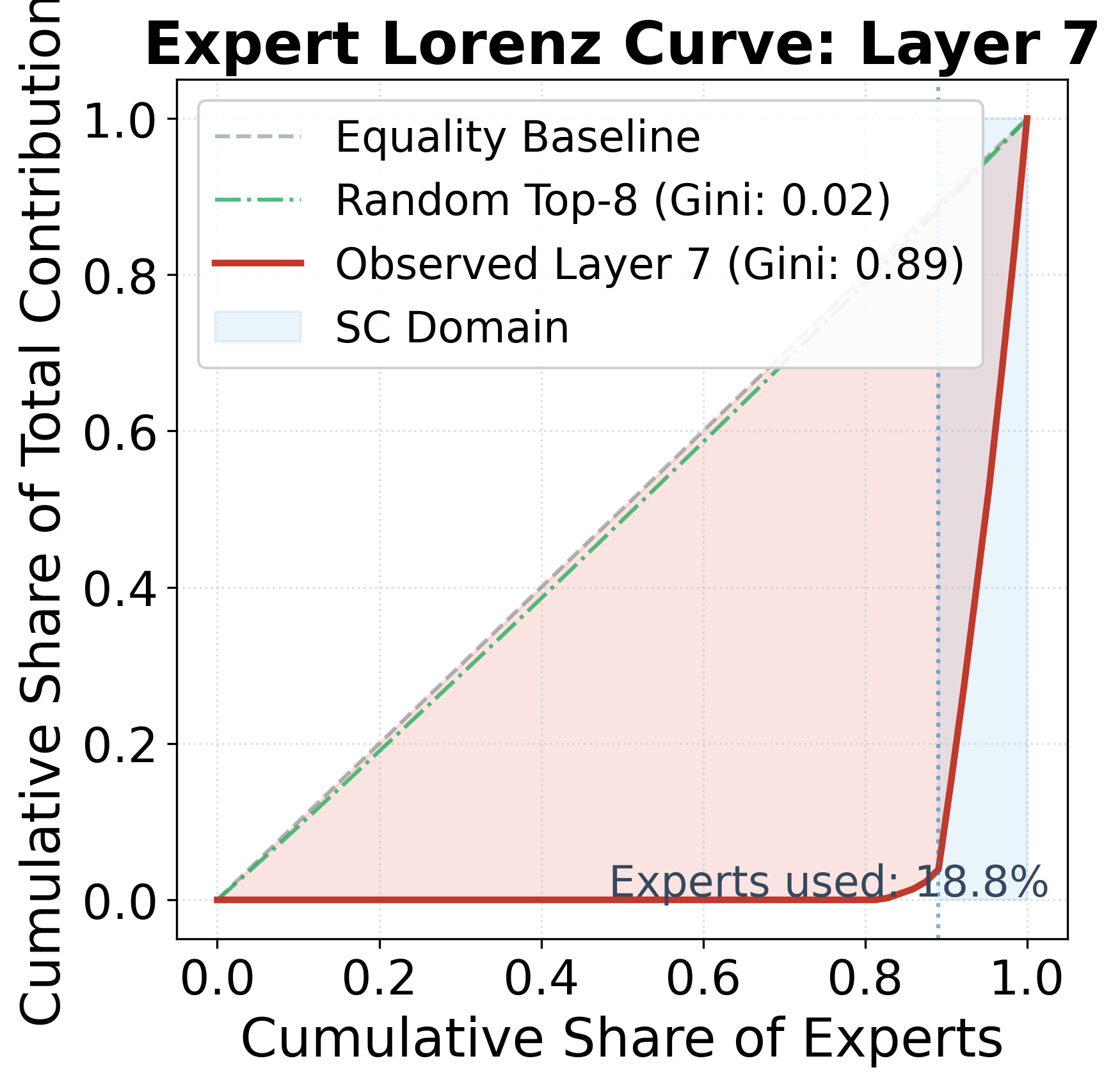}
\caption{Layer 7}
\end{subcaptionblock}
\begin{subcaptionblock}{0.19\linewidth}
\includegraphics[width=\linewidth]{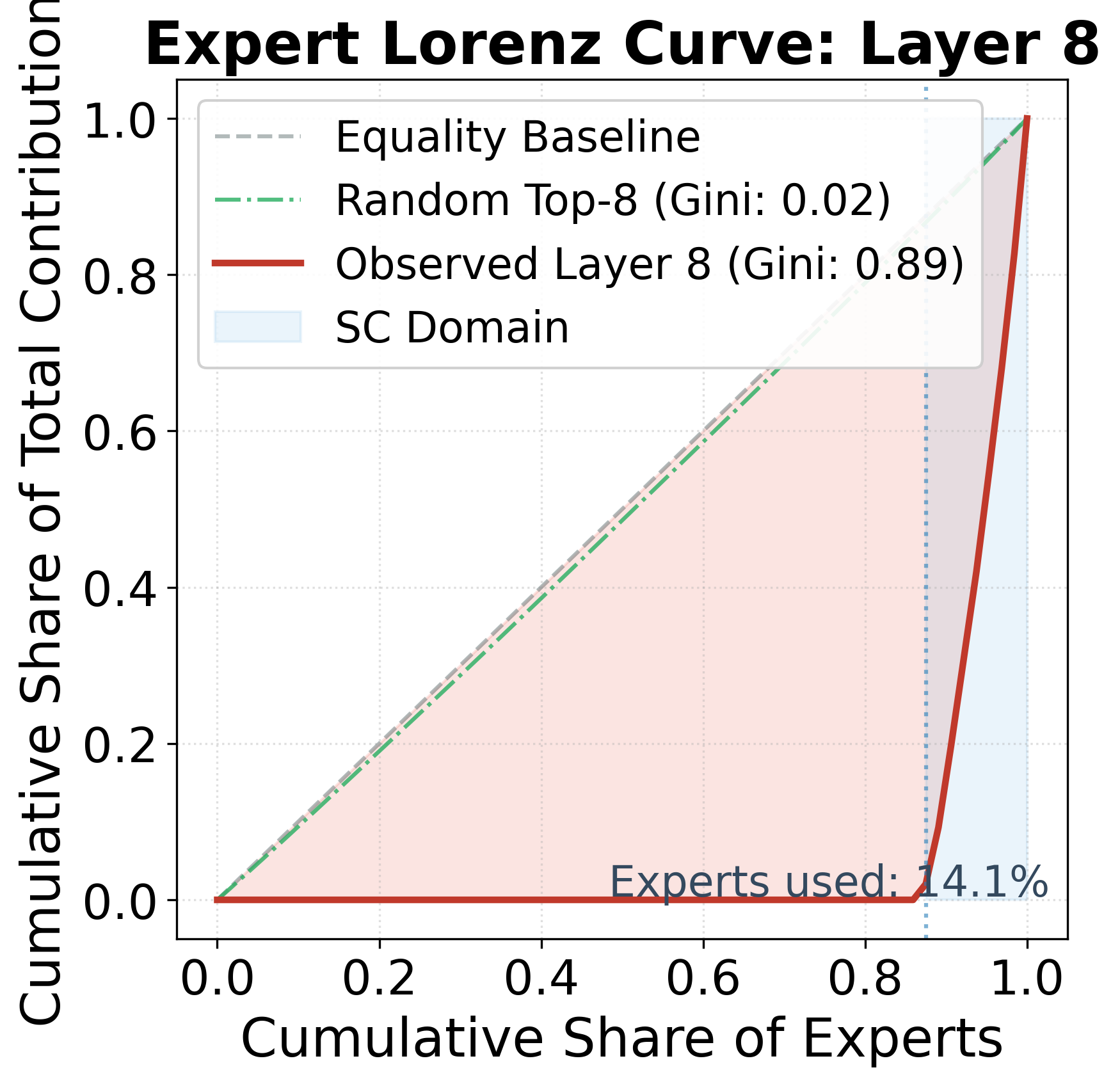}
\caption{Layer 8}
\end{subcaptionblock}
\begin{subcaptionblock}{0.19\linewidth}
\includegraphics[width=\linewidth]{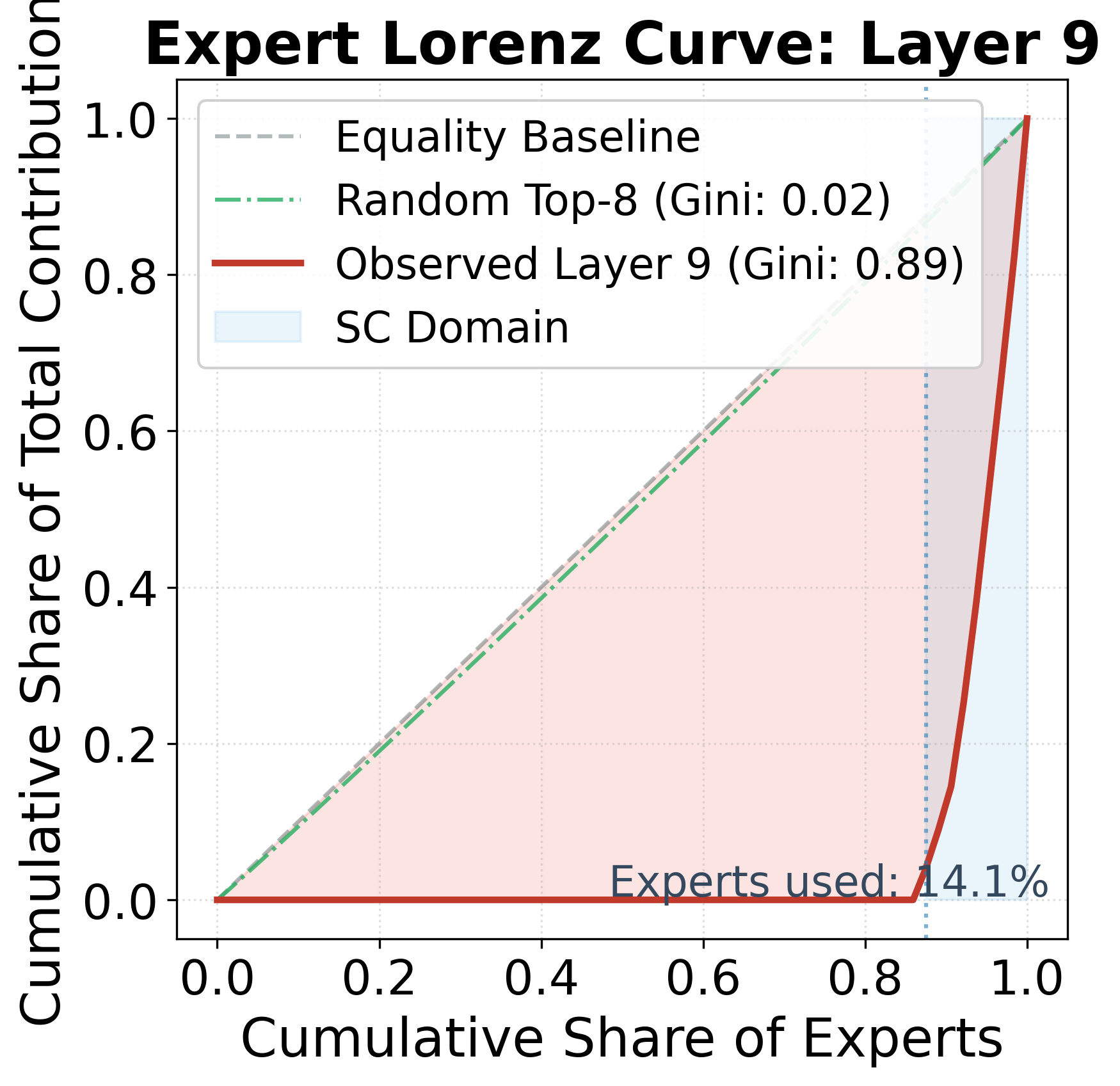}
\caption{Layer 9}
\end{subcaptionblock}

\medskip

\begin{subcaptionblock}{0.19\linewidth}
\includegraphics[width=\linewidth]{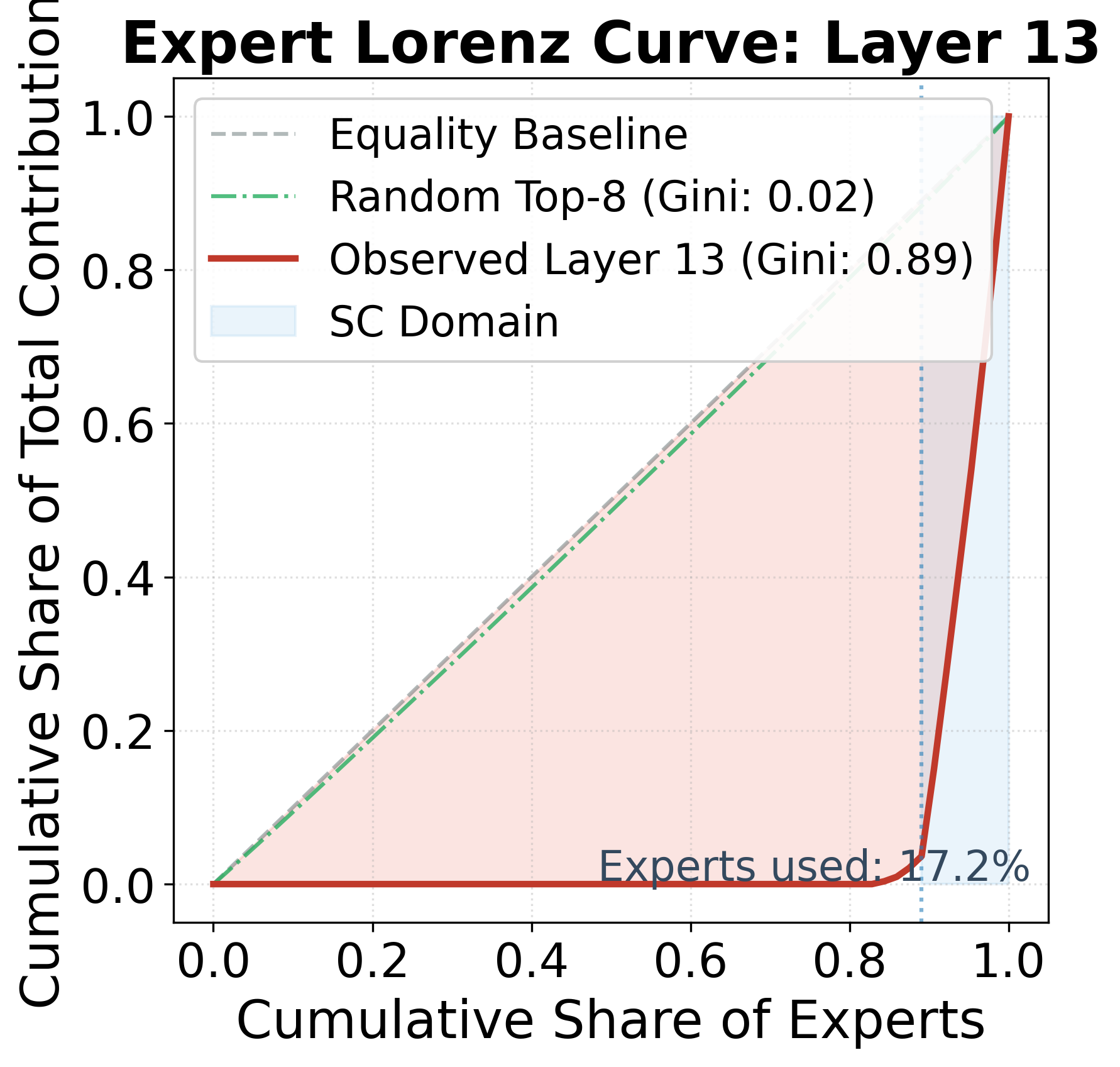}
\caption{Layer 13}
\end{subcaptionblock}
\begin{subcaptionblock}{0.19\linewidth}
\includegraphics[width=\linewidth]{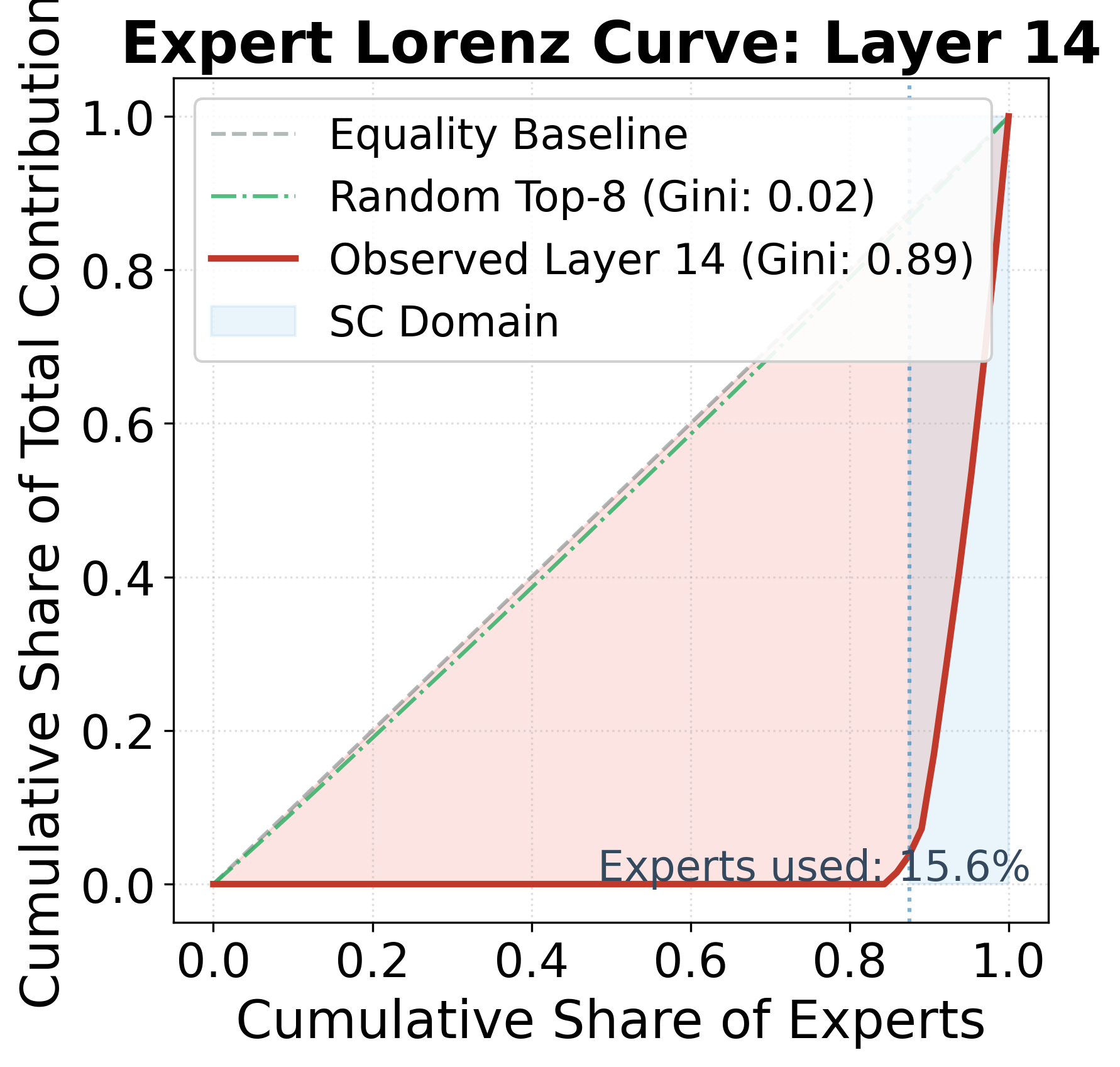}
\caption{Layer 14}
\end{subcaptionblock}
\begin{subcaptionblock}{0.19\linewidth}
\includegraphics[width=\linewidth]{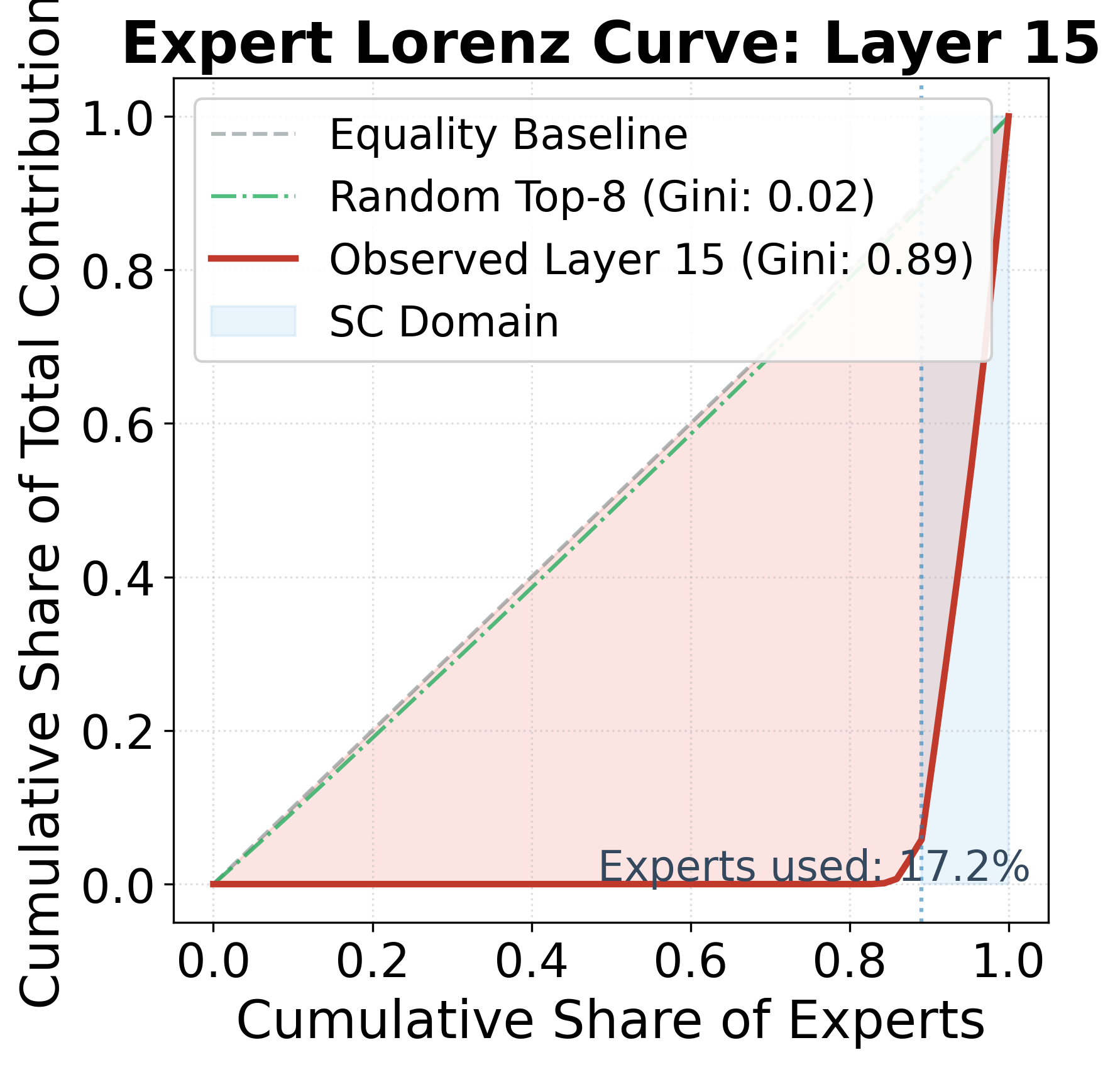}
\caption{Layer 15}
\end{subcaptionblock}

\caption{Expert Lorenz Curves across layers for OLMoE model.}
\label{fig:lorenz_all_olmoe}
\end{figure*}

The Lorenz curves for OLMoE, as shown in Figure~\ref{fig:lorenz_all_olmoe}, reveal a similarly concentrated contribution pattern, despite its smaller scale and more conservative routing design. Across layers, the curves bend sharply away from the equality baseline, with Gini coefficients consistently around $0.88$–$0.90$. This indicates that only a small subset of experts receives the majority of effective routing mass. Even when the nominal expert pool is relatively modest, the allocation of computation remains far from uniform.

Layer-wise inspection shows that this concentration is remarkably stable. Early layers, middle layers, and deep layers all display nearly identical Lorenz profiles, suggesting that specialization does not gradually diversify as representations become more abstract. Instead, OLMoE repeatedly falls back on the same compact subset of experts, while the remaining experts contribute minimally.

Interestingly, the fraction of “used” experts typically lies between $12\%$ and $20\%$, even though the router is free to assign mass more broadly. This implies that sparsity is driven less by necessity and more by the optimization dynamics of the gating network. Rather than distributing computation in a balanced way, the router converges toward a persistent core of high-traffic experts that dominate inference across inputs.

Taken together, the Lorenz curves demonstrate that contribution inequality is not merely a byproduct of model scale. Even in OLMoE, contribution is highly centralized, reinforcing the broader Standing Committee pattern: most experts exist on the periphery, while a small, repeatedly selected core absorbs the majority of computational responsibility.

\subsection{Lorenz Curve for DeepSeek-V2-Lite Model}

\begin{figure*}[t]
\centering

\begin{subcaptionblock}{0.19\linewidth}
\includegraphics[width=\linewidth]{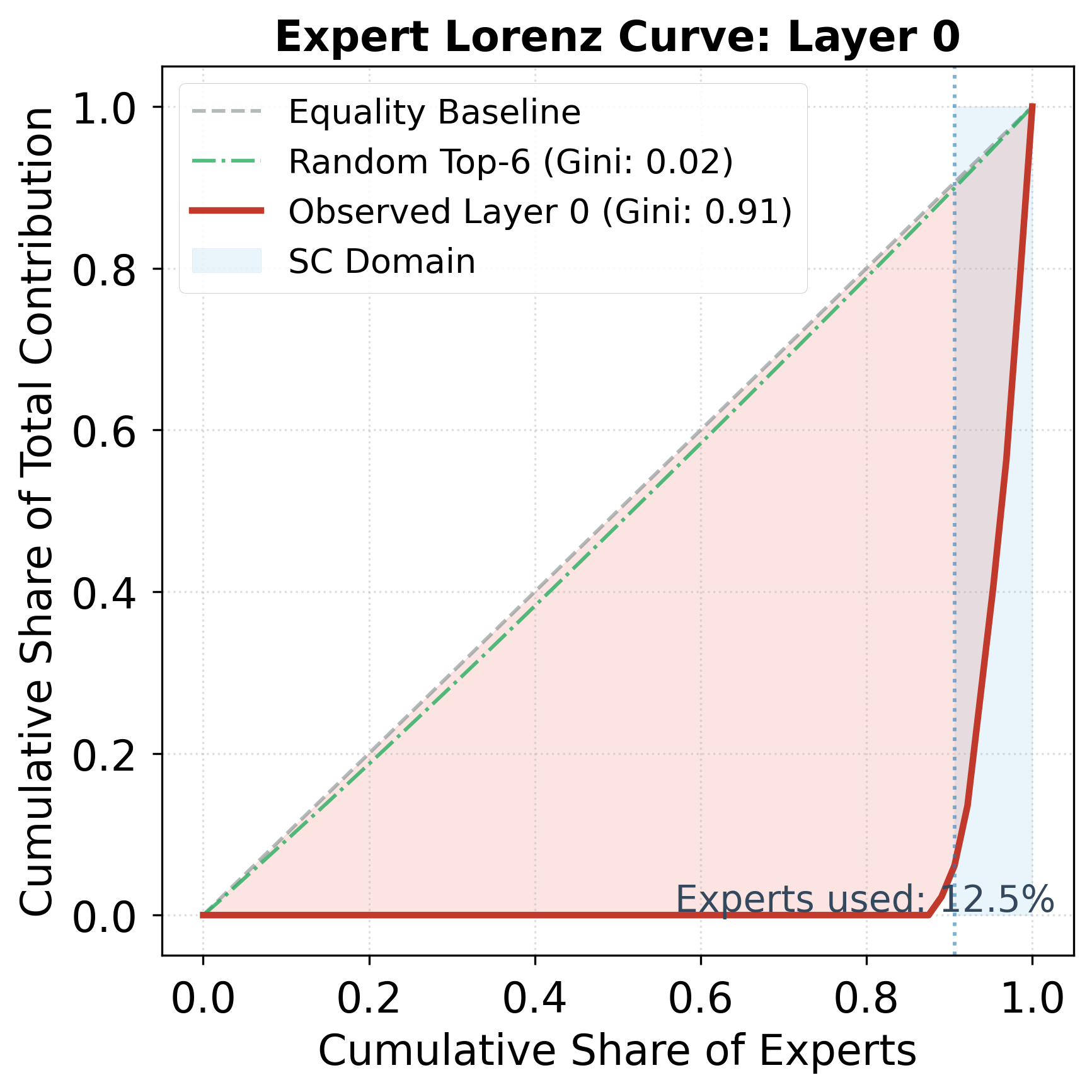}
\caption{Layer 0}
\end{subcaptionblock}
\begin{subcaptionblock}{0.19\linewidth}
\includegraphics[width=\linewidth]{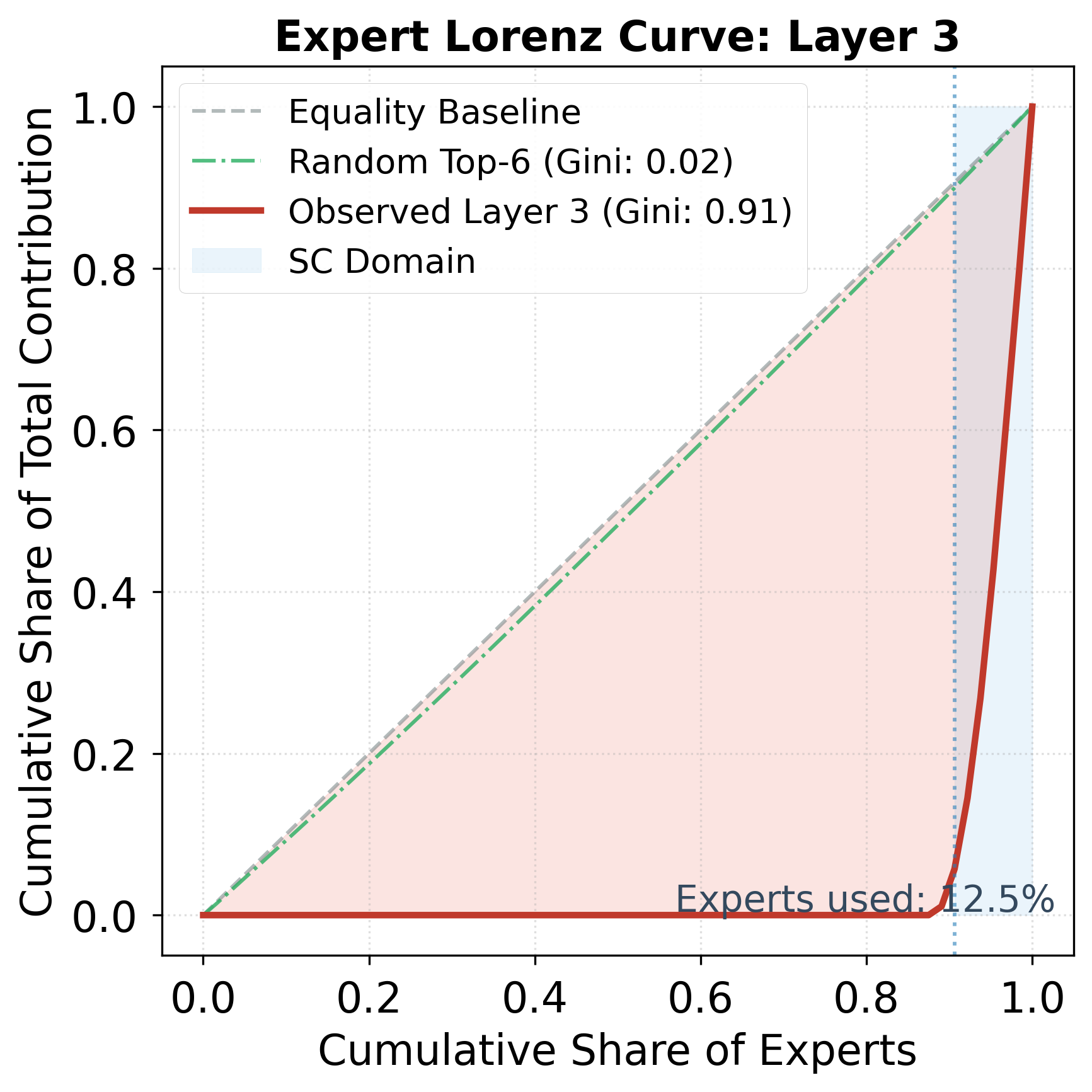}
\caption{Layer 3}
\end{subcaptionblock}
\begin{subcaptionblock}{0.19\linewidth}
\includegraphics[width=\linewidth]{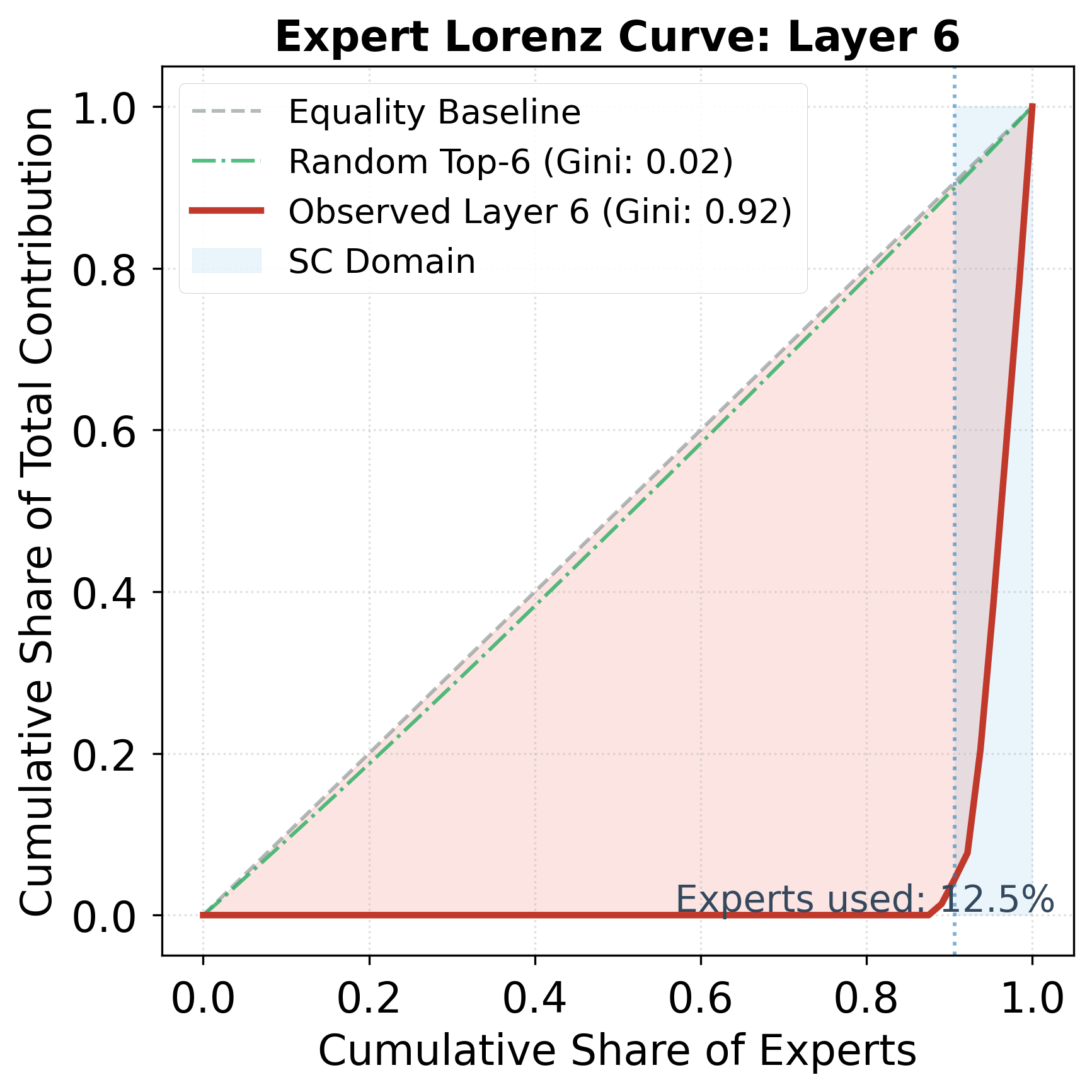}
\caption{Layer 6}
\end{subcaptionblock}

\medskip

\begin{subcaptionblock}{0.19\linewidth}
\includegraphics[width=\linewidth]{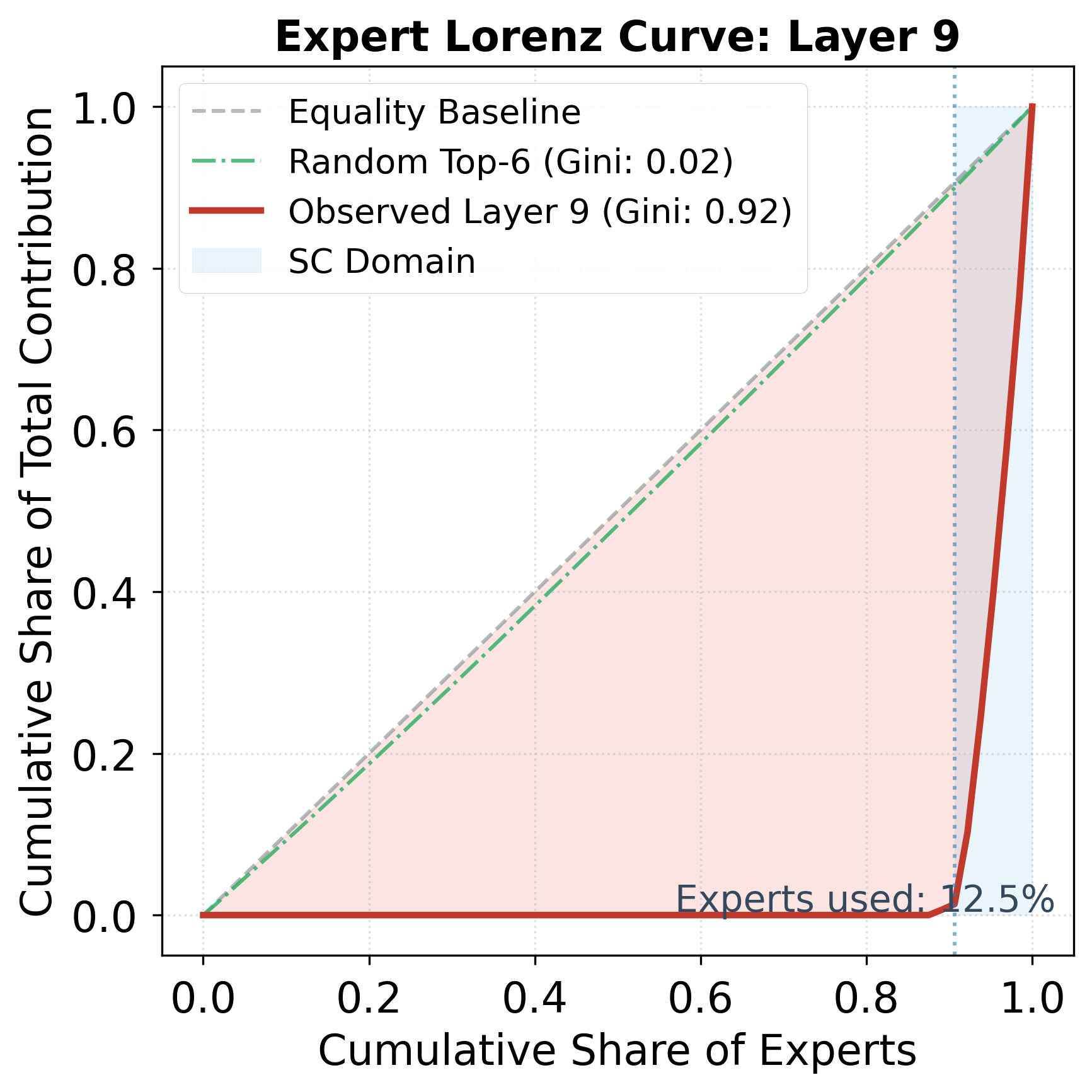}
\caption{Layer 9}
\end{subcaptionblock}
\begin{subcaptionblock}{0.19\linewidth}
\includegraphics[width=\linewidth]{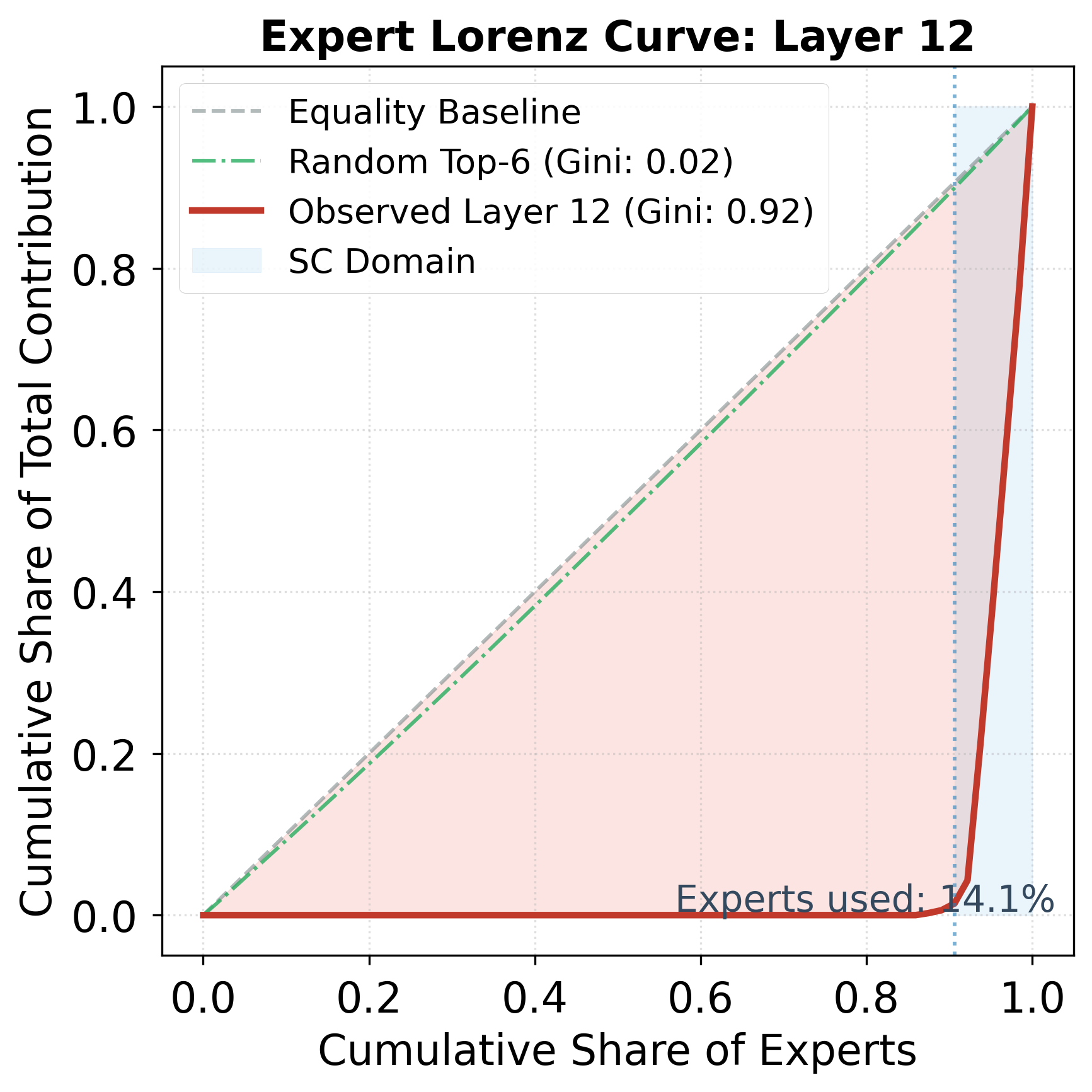}
\caption{Layer 12}
\end{subcaptionblock}
\begin{subcaptionblock}{0.19\linewidth}
\includegraphics[width=\linewidth]{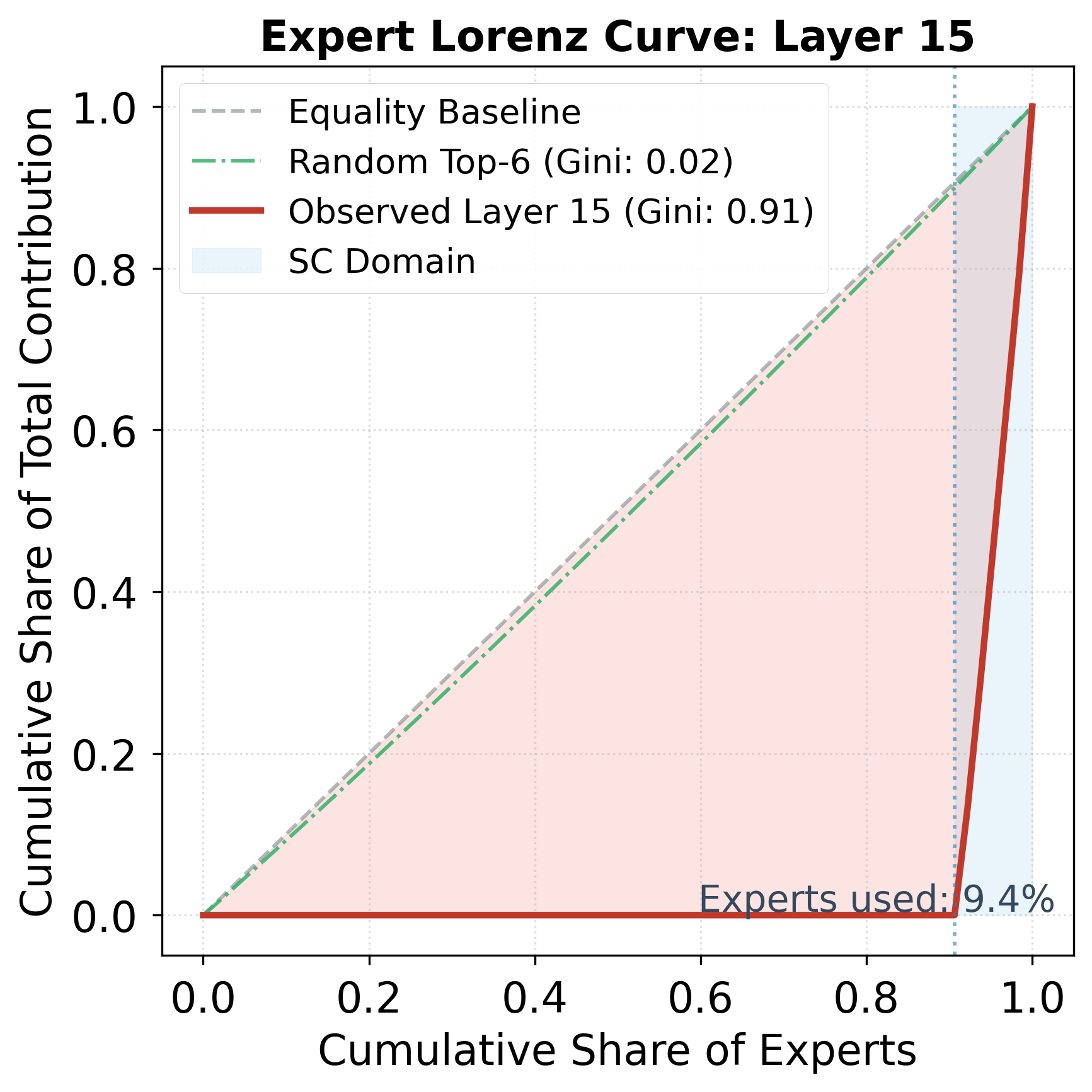}
\caption{Layer 15}
\end{subcaptionblock}

\medskip

\begin{subcaptionblock}{0.19\linewidth}
\includegraphics[width=\linewidth]{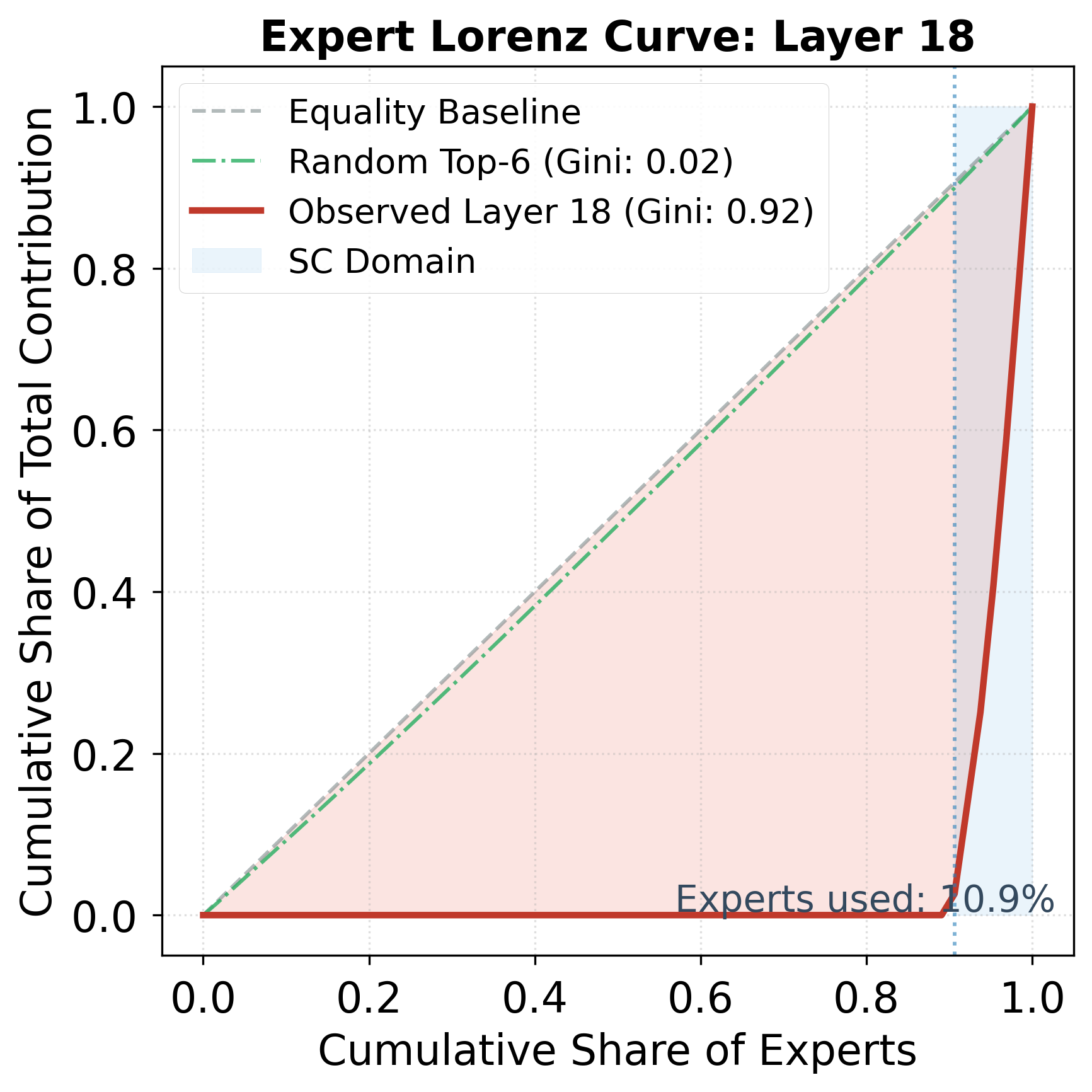}
\caption{Layer 18}
\end{subcaptionblock}
\begin{subcaptionblock}{0.19\linewidth}
\includegraphics[width=\linewidth]{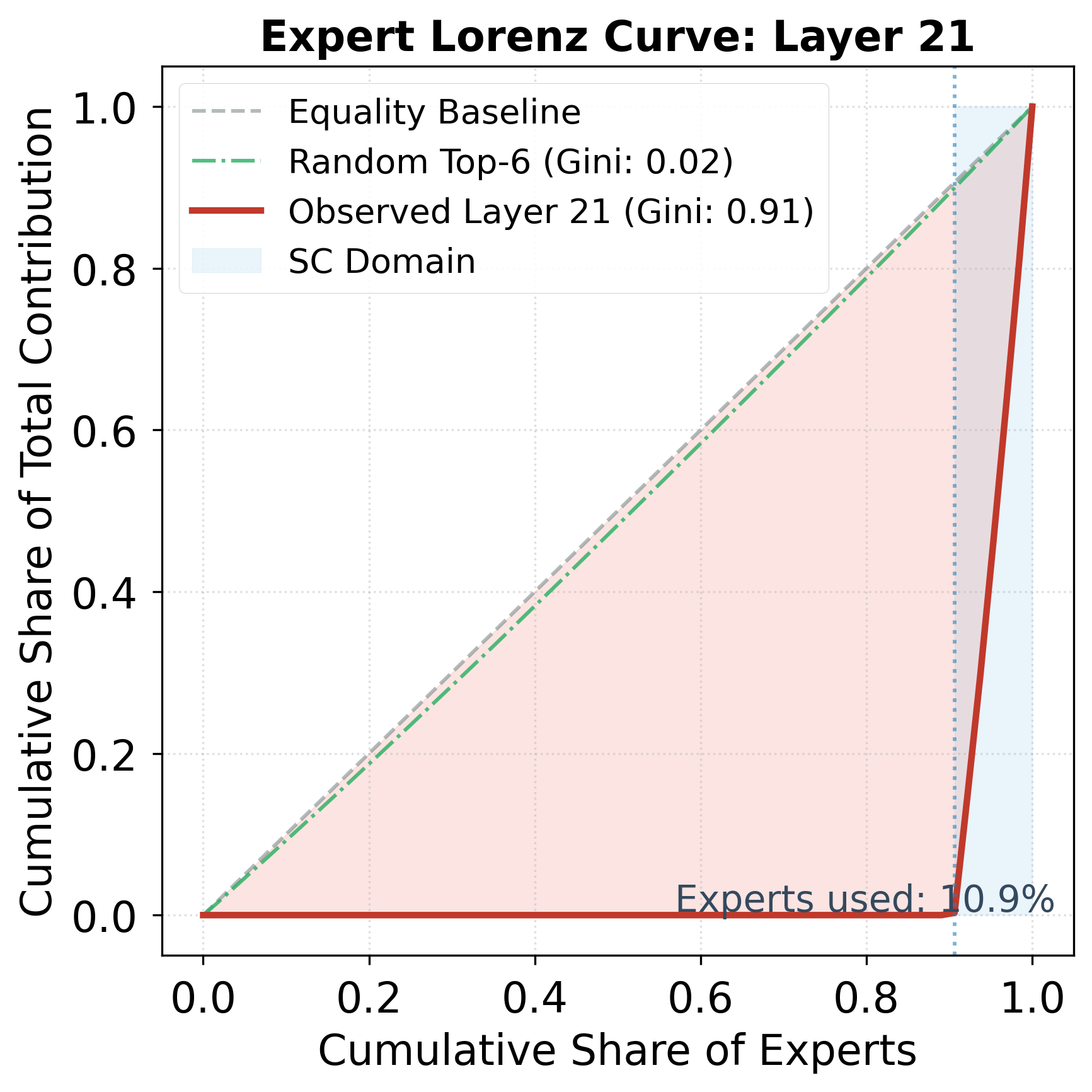}
\caption{Layer 21}
\end{subcaptionblock}
\begin{subcaptionblock}{0.19\linewidth}
\includegraphics[width=\linewidth]{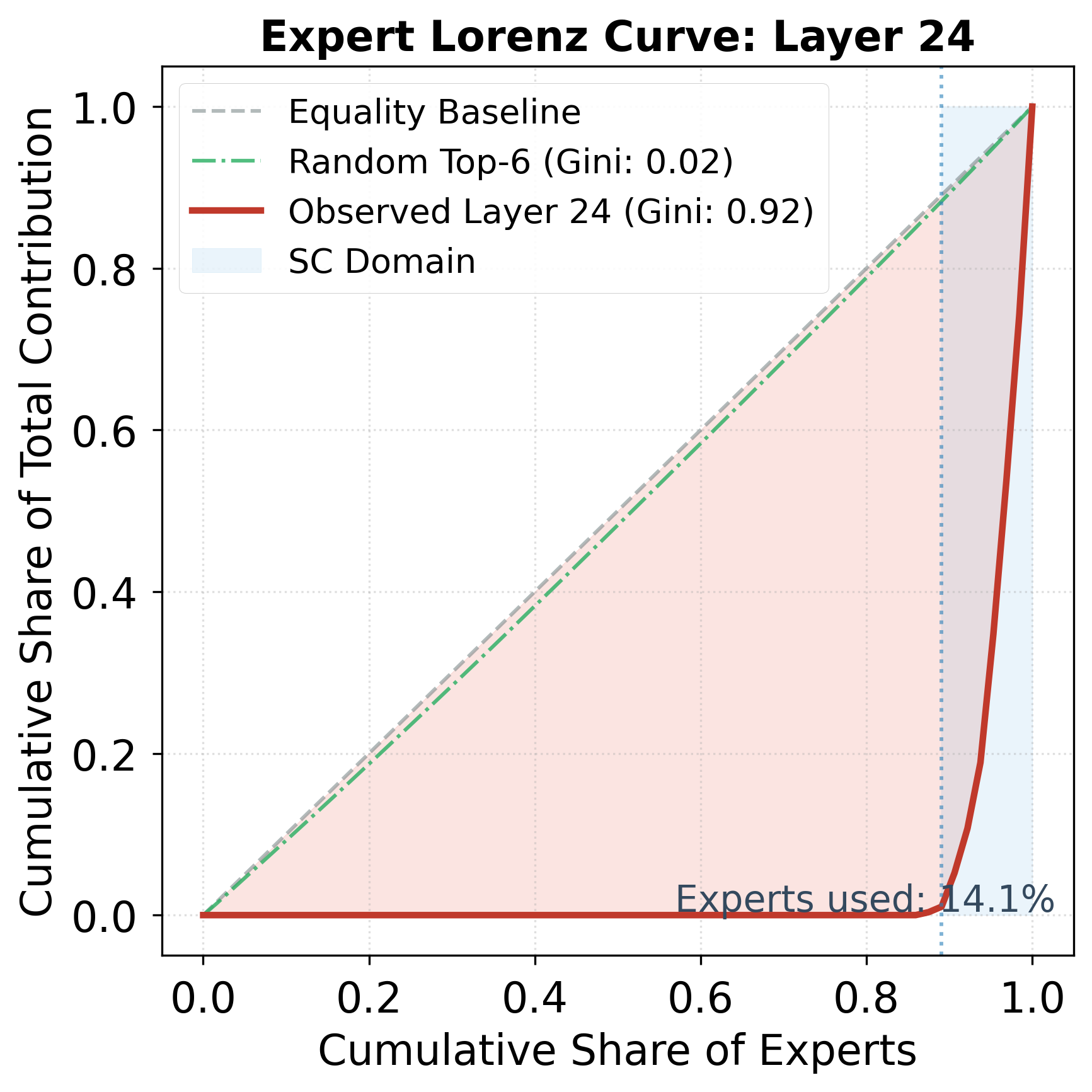}
\caption{Layer 24}
\end{subcaptionblock}

\caption{Expert Lorenz Curves across layers for DeekSeek-V2-Lite model.}
\label{fig:lorenz_all_deepseek}
\end{figure*}

DeepSeek-V2-Lite exhibits an even sharper form of contribution concentration, as shown in Figure~\ref{fig:lorenz_all_deepseek}. Across layers, the Lorenz curves bend aggressively toward the lower-right corner, with Gini coefficients consistently around $0.91$–$0.92$. This indicates that routing mass is dominated by a very small subset of experts. In several layers, fewer than $15\%$ of experts account for nearly all effective contributions, while the remaining experts receive negligible traffic.

Unlike what one might expect from a lightweight architecture optimized for efficiency, the inequality pattern does not relax as depth increases. Early, middle, and late layers display almost indistinguishable Lorenz shapes. The router repeatedly converges to the same compact set of high-traffic experts, rather than distributing load adaptively as representations evolve.

A notable pattern is the oscillation in the proportion of “used” experts. Some layers activate roughly $20\%$ of the pool, while others rely on as little as $9\%$. However, even in layers with broader activation, the cumulative share of contribution remains steeply skewed. This suggests that the increase in participation does not meaningfully change who dominates, but merely introduces additional peripheral experts who play marginal roles.

Taken together, these curves reinforce the central observation: contribution concentration is not mitigated by architectural simplification. DeepSeek-V2-Lite still organizes computation around a small, persistent Standing Committee, while most experts remain structurally available but functionally underutilized.

\subsection{Lorenz Curve for Qwen3 Model}

\begin{figure*}[t]
\centering

\begin{subcaptionblock}{0.19\linewidth}
\includegraphics[width=\linewidth]{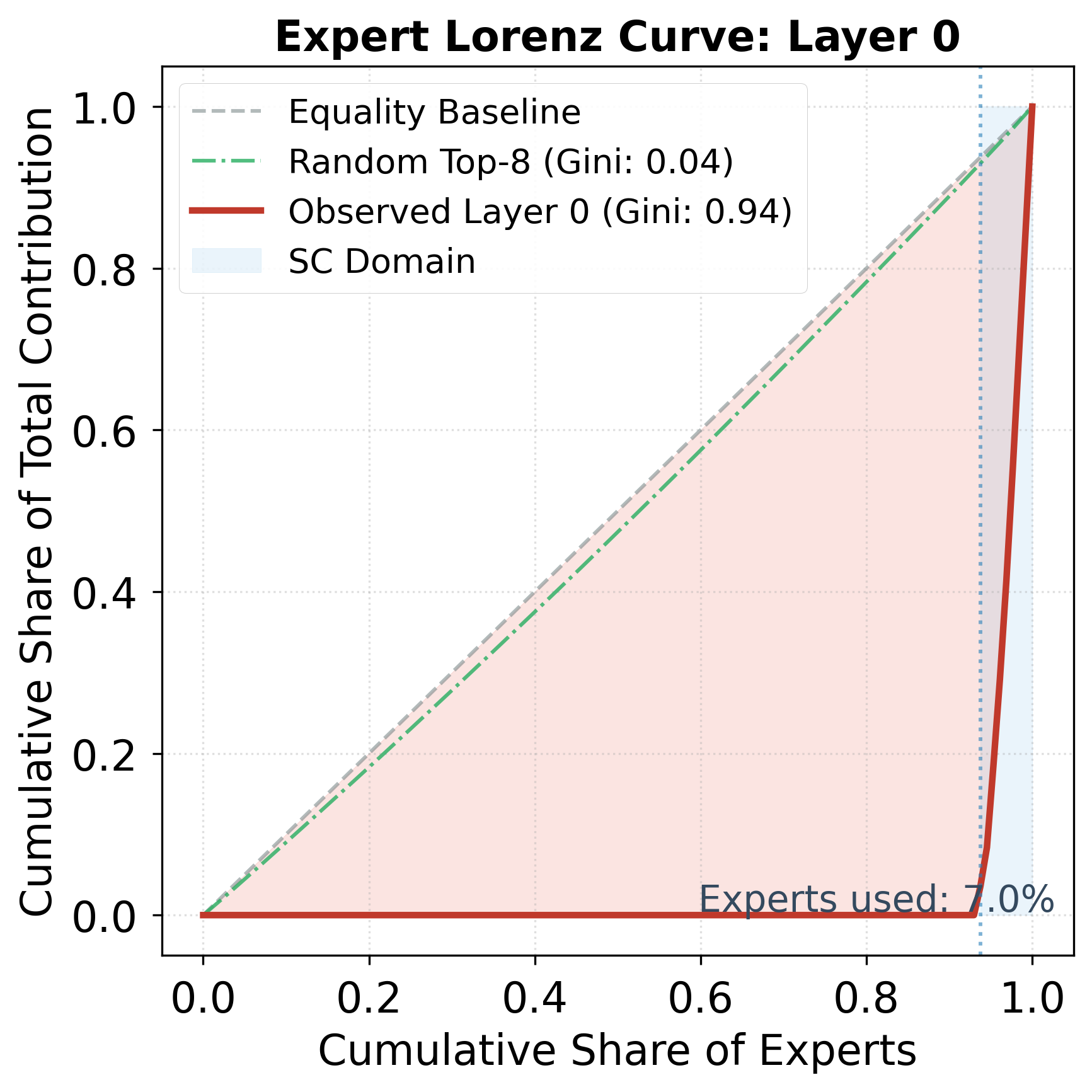}
\caption{Layer 0}
\end{subcaptionblock}
\begin{subcaptionblock}{0.19\linewidth}
\includegraphics[width=\linewidth]{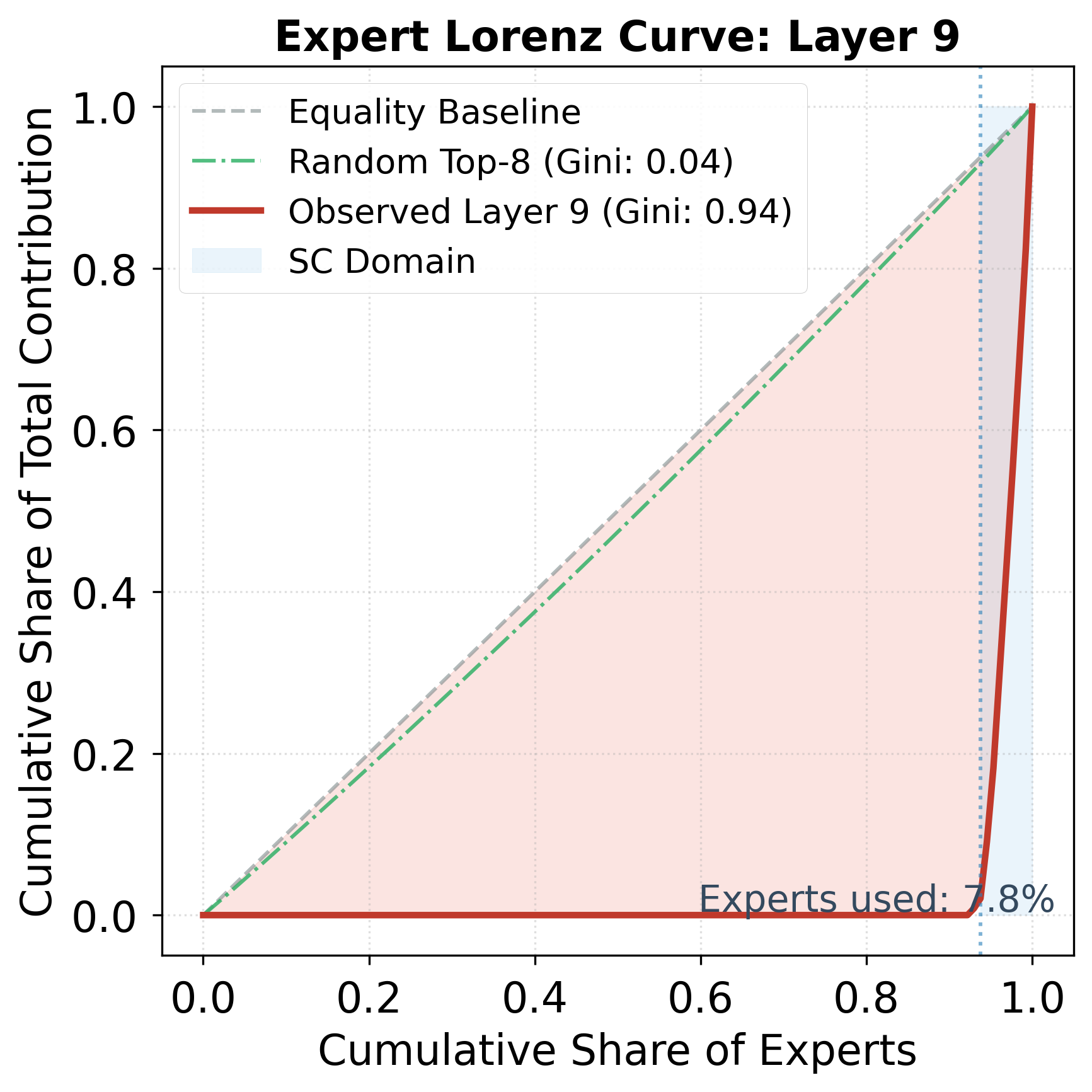}
\caption{Layer 9}
\end{subcaptionblock}
\begin{subcaptionblock}{0.19\linewidth}
\includegraphics[width=\linewidth]{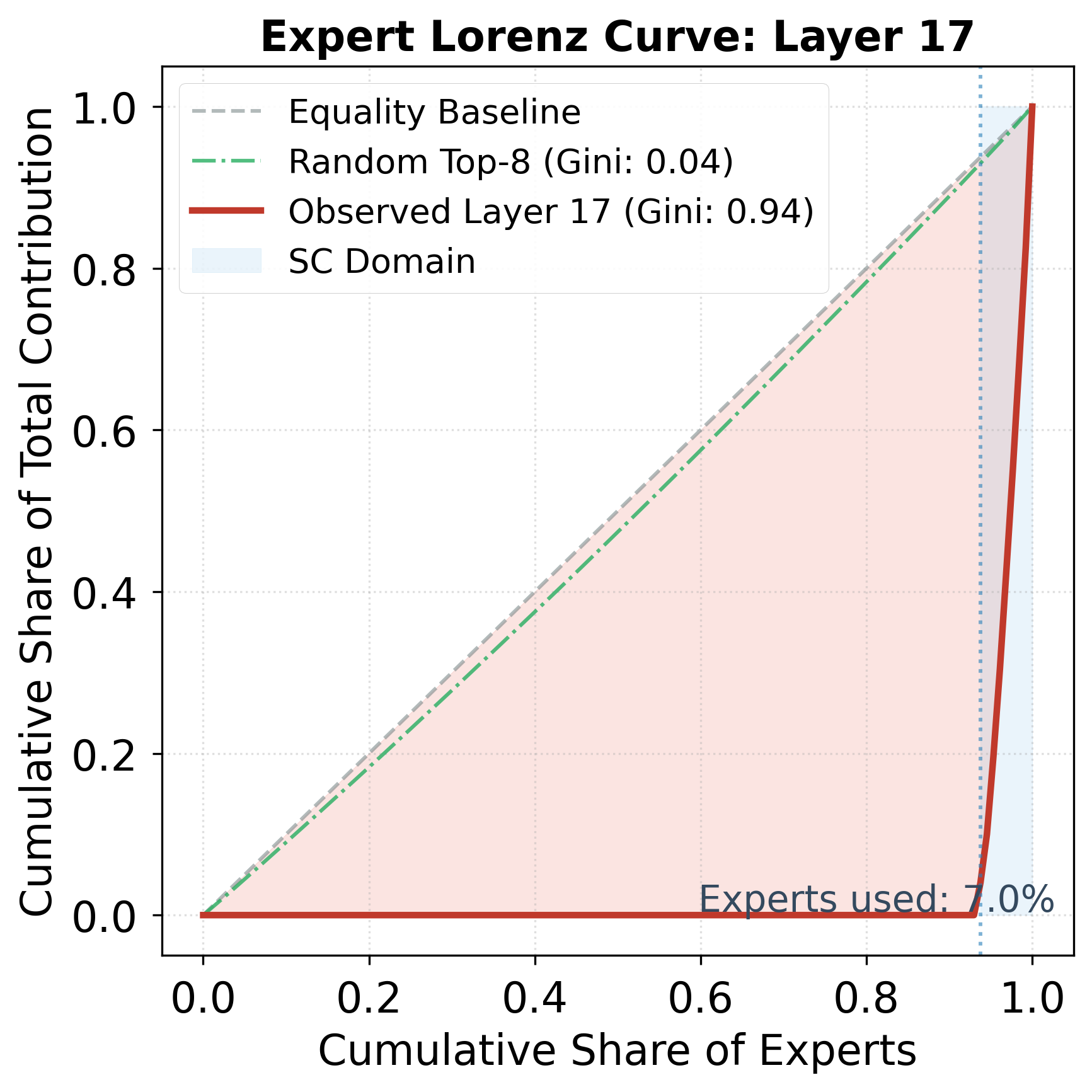}
\caption{Layer 17}
\end{subcaptionblock}

\medskip

\begin{subcaptionblock}{0.19\linewidth}
\includegraphics[width=\linewidth]{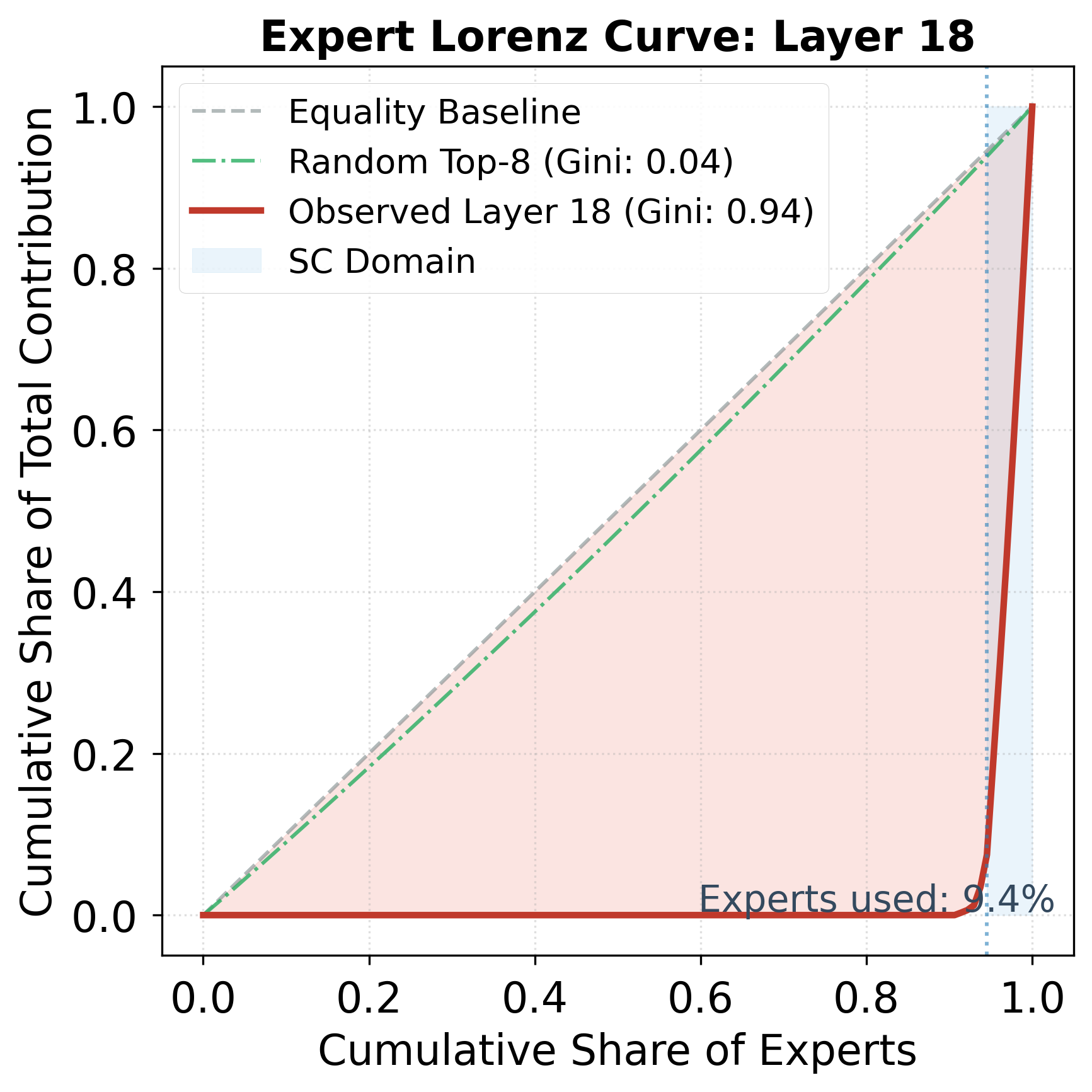}
\caption{Layer 18}
\end{subcaptionblock}
\begin{subcaptionblock}{0.19\linewidth}
\includegraphics[width=\linewidth]{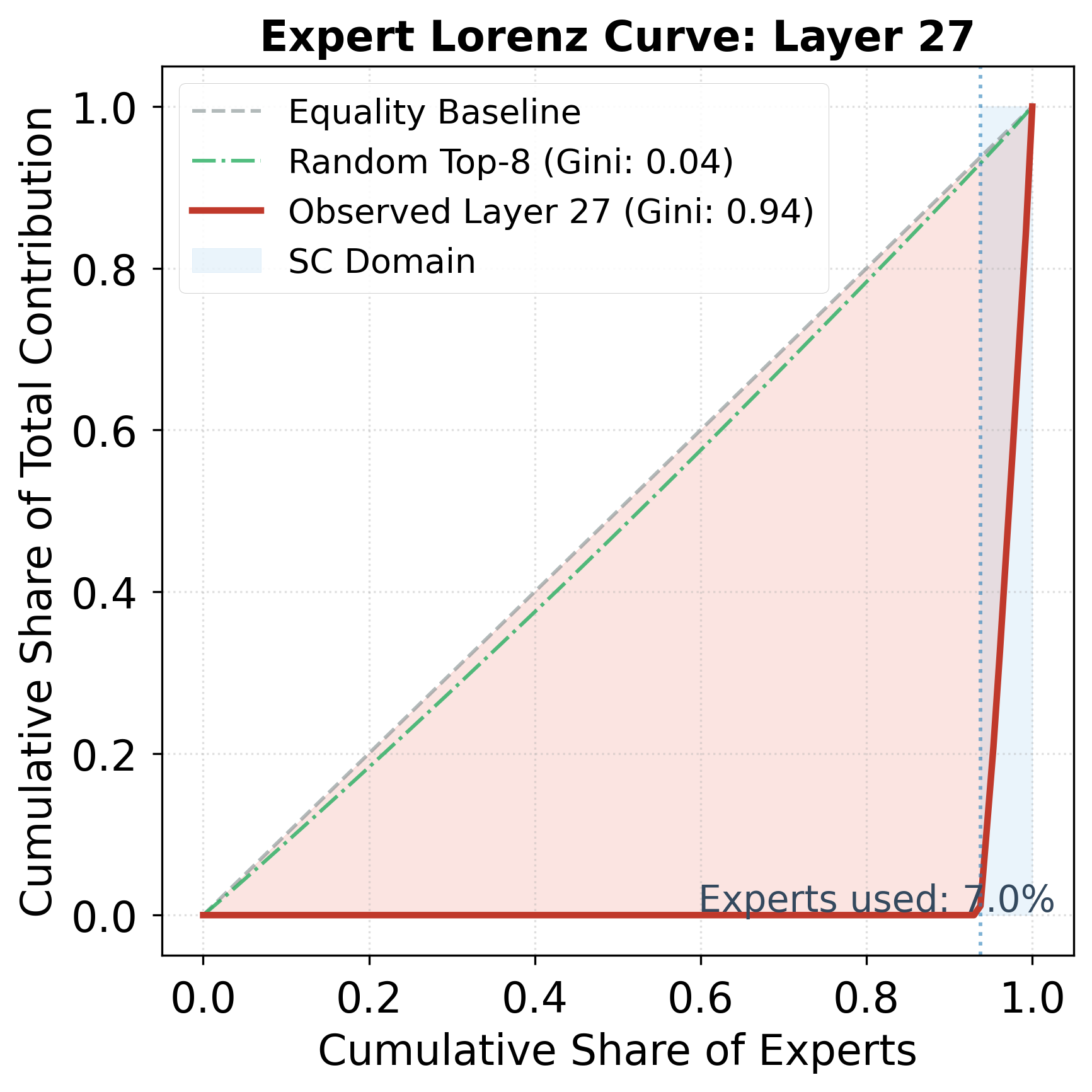}
\caption{Layer 27}
\end{subcaptionblock}
\begin{subcaptionblock}{0.19\linewidth}
\includegraphics[width=\linewidth]{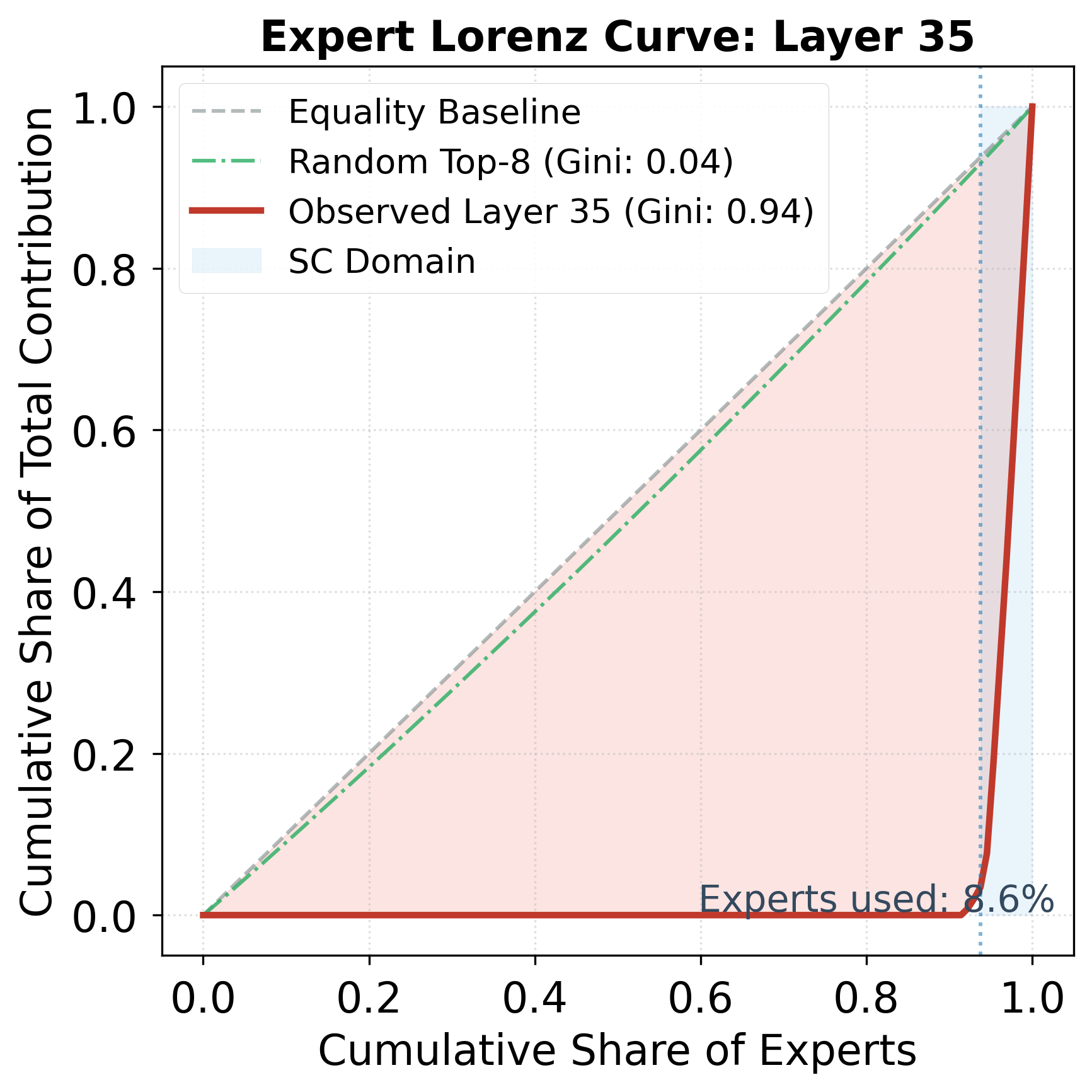}
\caption{Layer 35}
\end{subcaptionblock}

\medskip

\begin{subcaptionblock}{0.19\linewidth}
\includegraphics[width=\linewidth]{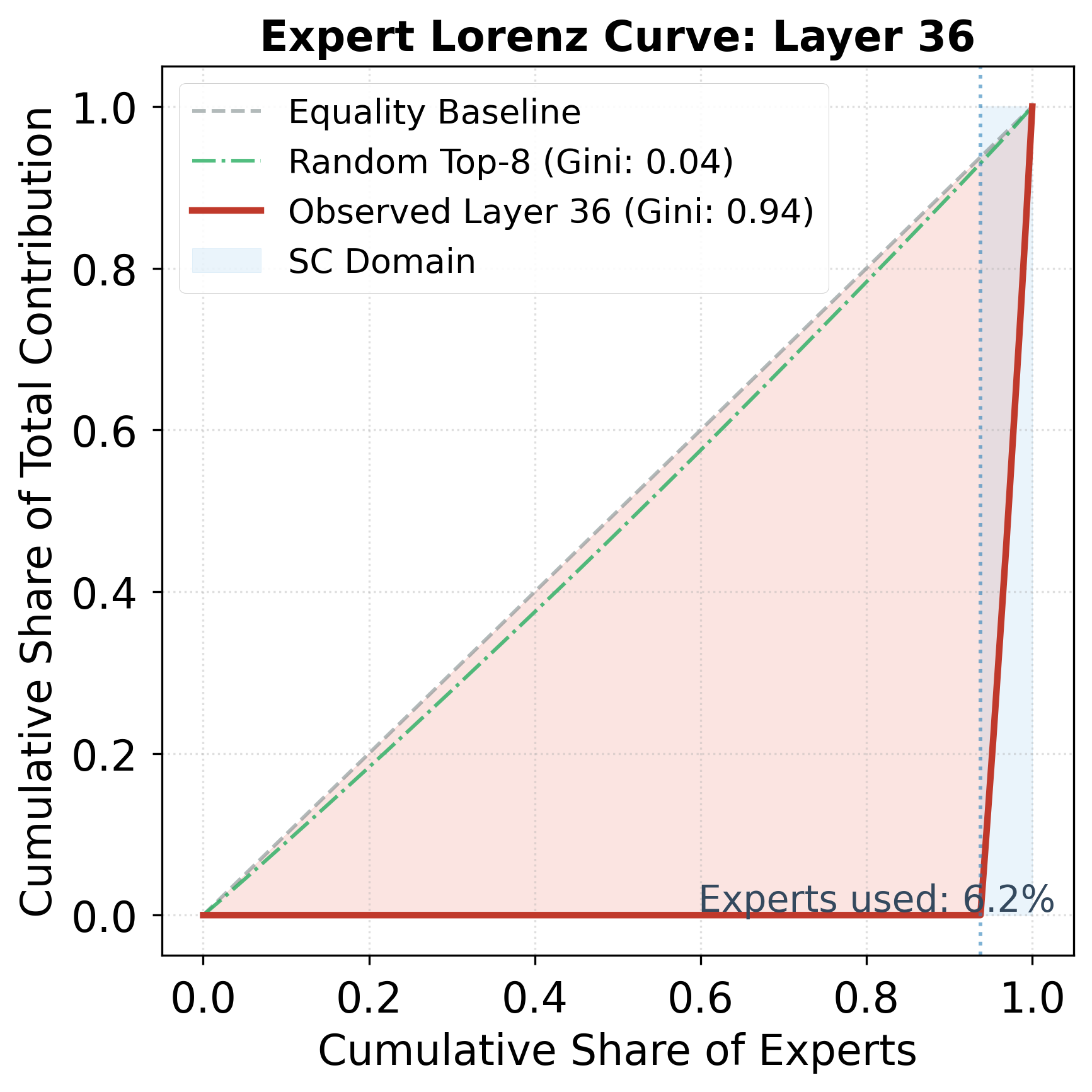}
\caption{Layer 36}
\end{subcaptionblock}
\begin{subcaptionblock}{0.19\linewidth}
\includegraphics[width=\linewidth]{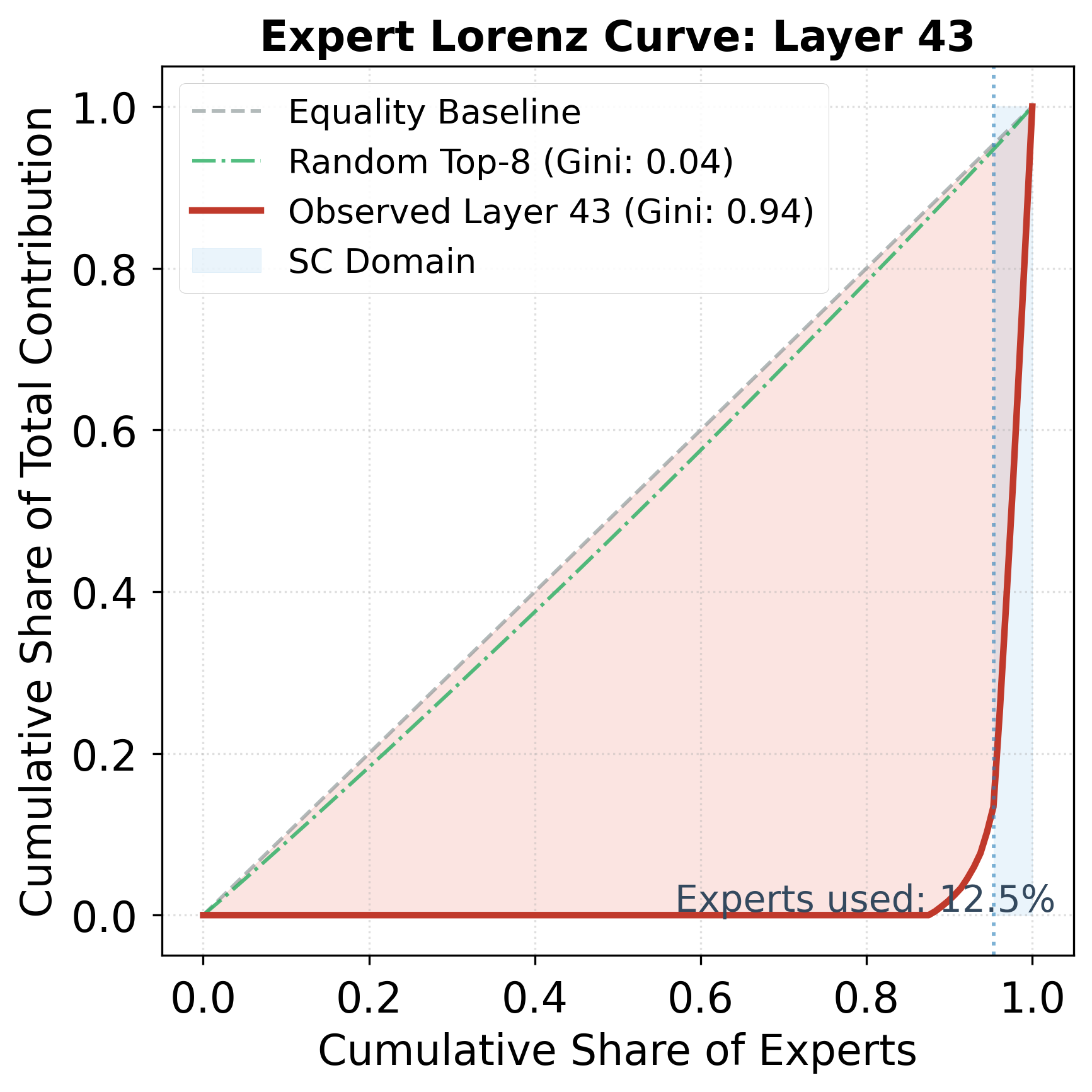}
\caption{Layer 43}
\end{subcaptionblock}
\begin{subcaptionblock}{0.19\linewidth}
\includegraphics[width=\linewidth]{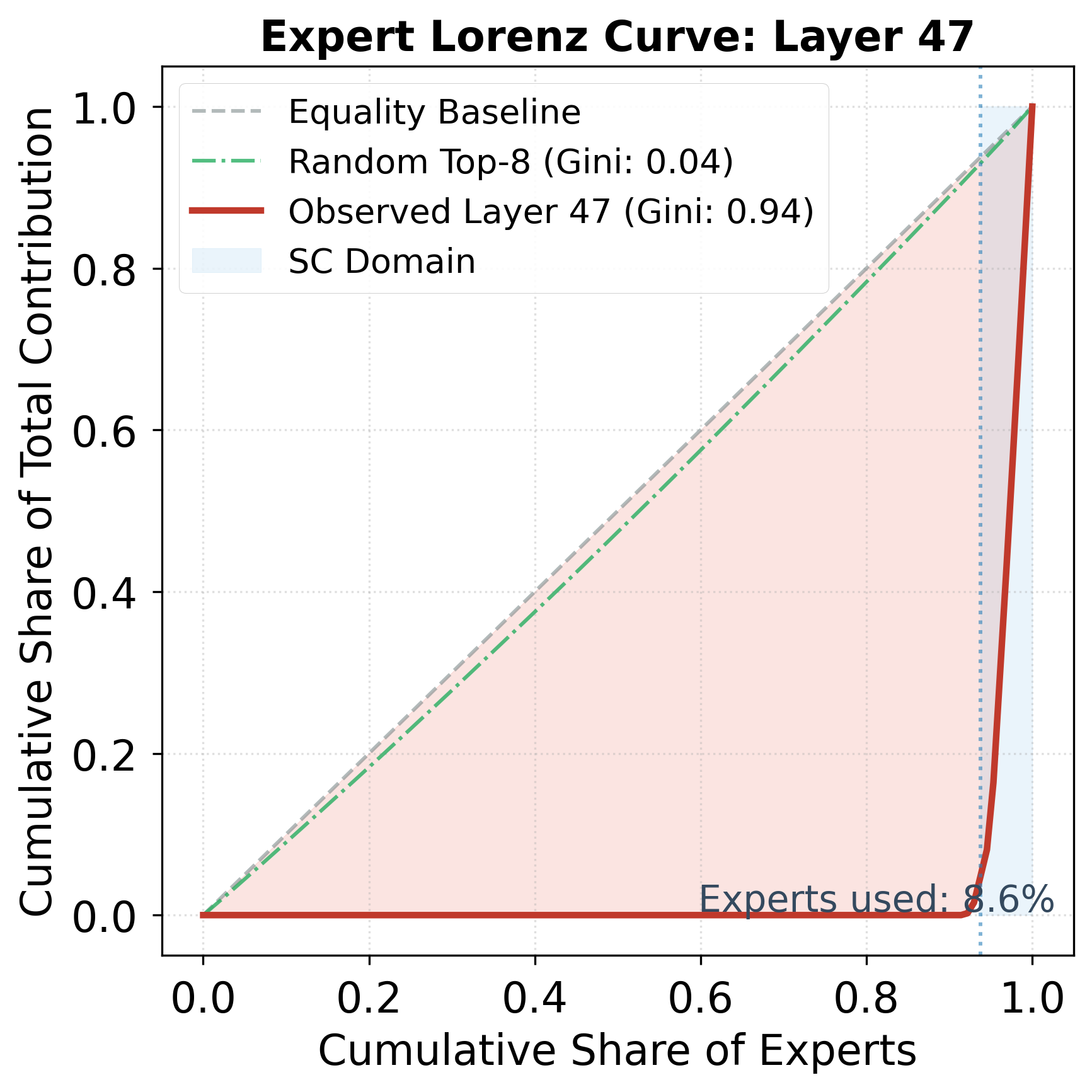}
\caption{Layer 47}
\end{subcaptionblock}

\caption{Expert Lorenz Curves across layers for QWen3-30B-A3B model.}
\label{fig:lorenz_all_qwen}
\end{figure*}

As shown in Figure~\ref{fig:lorenz_all_qwen}, Qwen-30B-A3B shows one of the strongest forms of contribution inequality across all models we study. The Lorenz curves are almost vertical near the right edge, yielding Gini coefficients around $0.94$ across layers. This indicates that routing mass is funneled into an extremely small subset of experts. In several layers, fewer than $10\%$ of experts account for nearly all effective contribution, leaving the majority effectively idle.

The pattern is also highly consistent across depth. Early layers, mid-depth layers, and late layers exhibit nearly identical Lorenz profiles, suggesting that the model does not gradually diversify expert usage as representations become more abstract. Instead, the router repeatedly returns to the same high-traffic subset, which acts as a default computational pathway for most inputs.

A striking observation is that the proportion of “active” experts fluctuates between $6\%$ and $12\%$, yet the inequality curve barely changes. Even when more experts are nominally activated, the cumulative contribution remains concentrated in a tiny elite group. Additional experts merely contribute marginal amounts, without altering the dominance structure.

These results indicate that contribution centralization intensifies as model capacity increases. In Qwen-30B-A3B, a large pool of experts does not translate into broader participation. Rather, the gating dynamics amplify the emergence of a persistent Standing Committee, while most experts remain structurally available but functionally peripheral.

\subsection{Cross-Model Synthesis.}
Across architectures of very different sizes and routing designs, we observe a consistent pattern of extreme contribution concentration. OLMoE, DeepSeek-V2-Lite, and Qwen-30B-A3B all display Lorenz curves that deviate sharply from the equality baseline, with Gini coefficients typically above $0.88$ and often exceeding $0.94$. In every case, only a small fraction of experts accounts for the vast majority of effective routing mass, while the remaining experts make negligible contributions.

Importantly, this phenomenon persists across depth. Early layers, middle layers, and late layers show nearly identical inequality profiles, indicating that expert participation does not broaden as representations become more abstract. Instead, the gating networks repeatedly allocate computation to a compact, stable subset of experts that serve as default processing routes, regardless of layer position or domain.

At the same time, fluctuations in the proportion of “used” experts do not materially change this distribution. Even when more experts are nominally activated, the cumulative contribution remains dominated by the same small core. Additional experts tend to act as low-impact auxiliaries rather than genuine participants in computation.

Taken together, these results suggest that contribution concentration is not merely an artifact of scale, architecture, or routing hyperparameters. Rather, it reflects a robust inductive tendency of sparse MoE optimization. The models converge toward a Standing Committee structure, in which a persistent core of experts monopolizes computation while most experts operate peripherally.

\section{Full Masking-Based Intervention Results}
\label{masking_exp}

To further examine the functional importance of Standing Committee experts, we conduct a masking-based intervention on DeepSeek-V2-Lite. For each selected layer, we suppress the routing weights of the identified Standing Committee experts, re-normalize the remaining routing distribution, and then evaluate answer outcomes on MMLU. While the main text reports representative shallow, middle, and deep layers, Table~\ref{tab:committee_masking_full} provides the complete set of evaluated layers.

The full results reveal a consistent degradation pattern across depth. Relative to the unmasked baseline, masking committee experts sharply reduces the correct answer rate and substantially increases the proportion of \textit{No Answer} responses. The effect becomes especially severe in deeper layers, where correct responses nearly collapse.

\begin{table}[t]
\centering
\small
\caption{Full masking-based intervention results on DeepSeek-V2-Lite. For each selected layer, we suppress the routing weights of the identified Standing Committee experts, re-normalize the remaining routing distribution, and evaluate answer outcomes on MMLU. The unmasked baseline is shown in the first row. Representative shallow, middle, and deep results are reported in the main text.}
\label{tab:committee_masking_full}

\resizebox{\linewidth}{!}{
\begin{tabular}{l l c c c}
\toprule
\textbf{Phase} & \textbf{Mask Layer} & \textbf{Correct} & \textbf{Wrong} & \textbf{No Answer} \\
\midrule
Baseline & None & 0.39 & 0.58 & 0.03 \\
\midrule
Shallow & 1  & 0.19 & 0.46 & 0.35 \\
Shallow & 2  & 0.12 & 0.52 & 0.36 \\
Shallow & 3  & 0.17 & 0.50 & 0.33 \\
\midrule
Middle  & 9  & 0.11 & 0.53 & 0.36 \\
Middle  & 10 & 0.09 & 0.55 & 0.36 \\
Middle  & 11 & 0.10 & 0.54 & 0.36 \\
\midrule
Deep    & 19 & 0.07 & 0.56 & 0.37 \\
Deep    & 20 & 0.06 & 0.57 & 0.37 \\
Deep    & 21 & 0.05 & 0.58 & 0.37 \\
Deep    & 25 & 0.04 & 0.58 & 0.38 \\
Deep    & 26 & 0.03 & 0.59 & 0.38 \\
\bottomrule
\end{tabular}
}
\end{table}
\end{document}